\documentclass[twoside]{article}

\usepackage[accepted]{aistats2019}
\usepackage[round]{natbib}

\usepackage[utf8]{inputenc} 
\usepackage[T1]{fontenc}    
\usepackage{hyperref} 
\usepackage{url}            
\usepackage{booktabs}       
\usepackage{amsfonts}       
\usepackage{nicefrac}       
\usepackage{microtype}      

\usepackage{dsfont, bbm, setspace, overpic}
\usepackage{amsmath, amsthm, amssymb}
\usepackage{mathabx} 
\usepackage{epsfig, psfrag, graphicx}
\usepackage{relsize, floatpag,wrapfig} 
\usepackage{listings}
\usepackage[usenames]{color}
\usepackage{hyperref} 
\usepackage{grffile} 
\usepackage{colortbl}
\usepackage[ruled,noend,linesnumbered, noline]{algorithm2e}
\usepackage[shortlabels]{enumitem} 
\usepackage[linewidth=1pt]{mdframed}
\usepackage[bottom]{footmisc}
\usepackage{tabularx}
\usepackage{placeins}
\usepackage{balance} 

\usepackage{subcaption}
\usepackage{multirow}  
\DeclareMathOperator*{\E}{\mathbb{E}}  

\usepackage[linewidth=1pt]{mdframed}

\allowdisplaybreaks 
\usepackage{sectsty, stmaryrd}

\usepackage{multibib} 
\newcites{si}{Additional References for the Supplementary Information}

\newcommand{\beginsupplement}{ 
        \setcounter{section}{0}
        \renewcommand{\thesection}{S\arabic{section}} %
         \renewcommand{\thesubsection}{\thesection.\arabic{subsection}}
        \setcounter{table}{0}
        \renewcommand{\thetable}{S\arabic{table}} %
        \setcounter{figure}{0}
        \renewcommand{\thefigure}{S\arabic{figure}} %
     }

\DeclareMathOperator*{\argmax}{argmax}

\newcommand\minidots{\hbox to 1em{.\hss.\hss.}}
 
\newtheorem{prop}{Proposition}
\newcounter{factnum}
\newcounter{claimnum}
 \newcounter{defnum}
 \newcounter{exnum}
 \newtheorem{ex}{Example}

\makeatletter 
\newcommand{\removelatexerror}{\let\@latex@error\@gobble}
\makeatother 

\definecolor{LightGray}{gray}{0.9}

\makeatletter 
\def\blfootnote{\xdef\@thefnmark{}\@footnotetext}
\makeatother

\newcommand{\papertitle}{What made you do this? \\ Understanding black-box decisions with sufficient input subsets}

\newcommand{\qhs}{\mathrm{QHS}}

\runningtitle{Understanding black-box decisions with sufficient input subsets}
\runningauthor{Carter, Mueller, Jain, Gifford}

\begin{document}

%

%

\twocolumn[
\aistatstitle{\papertitle}
\aistatsauthor{ Brandon Carter* \And Jonas Mueller* \And Siddhartha Jain \And  David Gifford }
\aistatsaddress{ MIT Computer Science and Artificial Intelligence Laboratory } 
]

\begin{abstract}
Local explanation frameworks aim to rationalize particular decisions made by a black-box prediction model. 
Existing techniques are often restricted to a specific type of predictor or based on input saliency, which may be undesirably sensitive to factors unrelated to the model's decision making process.
We instead propose \emph{sufficient input subsets} that identify minimal subsets of features whose observed values alone suffice for the same decision to be reached, even if all other input feature values are missing.  
General principles that globally govern a model's decision-making can also be revealed by searching for clusters of such input patterns across many data points. 
Our approach is conceptually straightforward, entirely model-agnostic, simply implemented using instance-wise backward selection, and able to produce more concise rationales than existing techniques.
We demonstrate the utility of our interpretation method on various neural network models trained on text, image, and genomic data. 
\end{abstract}

\section{Introduction}
\label{sec:intro}

The rise of neural networks and nonparametric methods in machine learning (ML) has driven significant improvements in prediction capabilities, while simultaneously earning the field a reputation of producing complex black-box models. 
Vital applications, which could  benefit most from improved prediction, are often deemed too sensitive for opaque learning systems.  
Consider the widespread use of ML for screening people, including models that deny defendants' bail \citep{Kleinberg18} or reject loan applicants \citep{Sirignano18}.  It is imperative that such decisions can be interpretably rationalized. 
Interpretability is also crucial in scientific applications, where it is hoped that general principles may be extracted from accurate predictive models~\citep{interpretability, Lipton16}. 


\blfootnote{*Equal contribution. Code for this paper is available at: \url{http://github.com/b-carter/SufficientInputSubsets}}
One simple explanation for \emph{why} a particular black-box decision is  reached may be obtained via a sparse subset of the input features whose values form the basis for the model's decision -- a \emph{rationale}. 
For text (or image) data, a rationale might consist of a subset of positions in the document (or image) together with the words (or pixel-values) occurring at these positions (see Figures \ref{fig:beer-all-aspects-example} and \ref{fig:95sises}).   
To ensure interpretations remain fully faithful to an arbitrary model, our rationales do not attempt to summarize the (potentially complex) operations carried out within the model, and instead merely point to the relevant information it uses to arrive at a decision \citep{Lei16}. For high-dimensional inputs, sparsity of the rationale is imperative for greater interpretability.

Here, we propose a local explanation framework to produce rationales for a learned model that has been trained to map inputs $\mathbf{x} \in \mathcal{X}$ via some arbitrary learned function $f: \mathcal{X} \rightarrow \mathbb{R}$.  Unlike many other interpretability techniques, our approach is not restricted to vector-valued data and does not require gradients of $f$.
Rather, each input example is solely  presumed to have a set of indexable features $ \mathbf{x}  = [x_1,\dots, x_p]$,  
where each $x_i \in \mathbb{R}^d$ for $i \in [p] = \{1,\dots,p\}$.  We allow for features that are unordered (set-valued input) and whose number $p$ may vary from input to input.  A rationale corresponds to a sparse subset of these indices $S \subseteq [p]$ together with the specific values of the features in this subset.   

To understand why a certain decision was made for a given input example  $\mathbf{x}$, we propose a  particular rationale called a \emph{sufficient input subset} (SIS).  Each SIS consists of a minimal input pattern present in $\mathbf{x}$ that alone suffices for $f$ to produce the same decision, even if provided no other information about the rest of $\mathbf{x}$.
Presuming the decision is based on $f(\mathbf{x})$ exceeding some prespecified threshold $\tau \in \mathbb{R}$, we specifically seek a minimal-cardinality subset $S$ of the input features such that $f(\mathbf{x}_S) \ge \tau$.
Throughout, we use $\mathbf{x}_S \in \mathcal{X}$ to denote a modified input example in which all information about the values of features outside subset $S$ has been masked with features in $S$ remaining at their original values.    
Thus, each SIS characterizes a particular standalone input pattern that drives the model toward this decision, providing sufficient justification for this choice from the model's perspective, even without any information on the values of the other features in $\mathbf{x}$.

In classification settings, $f$ might represent the predicted probability of class $C$ where we decide to assign the input to class $C$ if $f(\mathbf{x}) \ge \tau$,  chosen based on precision/recall considerations. Each SIS in such an application corresponds to a small input pattern that on its own is highly indicative of class $C$, according to our model.
Note that by suitably defining $f$ and $\tau$ with respect to the predictor outputs, any particular decision for input $\mathbf{x}$ can be precisely identified with the occurrence of $f(\mathbf{x}) \ge \tau$, where higher values of $f$ are associated with greater confidence in this decision.

For a given input $\mathbf{x}$ where $f(\mathbf{x}) \ge \tau$, this work presents a simple method to find a complete collection of sufficient input subsets, each satisfying $f(\mathbf{x}_S) \ge \tau$, such that there exists no additional SIS outside of this collection.  Each SIS may be understood as a disjoint piece of evidence that would lead the model to the same decision, and why this decision was reached for $\mathbf{x}$ can be unequivocally attributed to the SIS-collection.  
Furthermore, global insight on the general principles underlying the model's decision-making process may be gleaned by clustering the types of SIS extracted across different data points (see Figure \ref{fig:mnist-clustering-4-maintext} and \ref{fig:mnist-model-differences-clustering-digit4}). 
Such insights allow us to compare models based not only on their accuracy, but also on human-determined relevance of the concepts they target.
Our method's simplicity facilitates its utilization by non-experts who may know very little about the models they wish to interrogate.

\section{Related Work}
\label{sec:rw}

Certain neural network variants such as attention mechanisms \citep{Sha17} and the generator-encoder of \citet{Lei16} have been proposed as powerful yet human-interpretable learners.  
Other interpretability efforts have tailored decompositions to certain convolutional/recurrent networks \citep{James18,olah17,olah18}
, but these approaches are model-specific and only suited for ML experts. 
Many applications necessitate a model outside of these families, either to ensure supreme accuracy, or if training is done separately with access restricted to a  black-box API \citep{Caruana15,predictionapi}. 

An alternative model-agnostic approach to interpretability produces local explanations of $f$ for a particular input $\mathbf{x}$. 
Local explanation often relies on attribution methods that quantify how much each feature influences the output of $f$ at $\mathbf{x}$.  Examples include LIME, which locally approximates $f$ \citep{lime}, 
saliency maps based on gradients of $f$  
\citep{Baehrens10, Simonyan14}, 
Layer-wise Relevance Propagation  \citep{Bach15},
as well as the discrete DeepLIFT approach \citep{deeplift} and its continuous variant -- Integrated Gradients (IG) \citep{Sundararajan17}, developed to ensure attributions  reflect the cumulative difference in $f$ at $\mathbf{x}$ vs.\ a reference input.  
A separate class of input-signal-based explanation techniques such as DeConvNet \citep{Zeiler14}, Guided Backprop \citep{Springenberg15}, and PatternNet \citep{Kindermans18} employ gradients of $f$ in order to identify input patterns that cause $f$ to output large values. 
However, many gradient-based saliency methods have been deemed unreliable, depending not only on the learned function $f$, but also on its specific architectural implementation and how inputs are scaled \citep{Kindermans17,Kindermans18}.
More like our approach, recent techniques from 
\citet{Dabkowski17,Kim18,Chen18}
also aim to identify input patterns that best explain certain decisions, but additionally require either a predefined set of such patterns or an auxiliary neural network trained to identify them.

In comparison with the aforementioned methods, our SIS approach is: conceptually simple, entirely faithful to any type of model, and requires neither gradients of $f$ nor auxiliary training of the underlying model $f$ or a surrogate explanation model.  
Also related to our subset-selection methodology are the ideas of \citet{erasure} and \citet{Fong17}, which for a particular input  seek a small feature subset whose omission causes a substantial drop in $f$ such that a different decision would be reached.  However, this objective can produce adversarial artifacts that are hard to interpret.
In contrast, we focus on identifying small subsets of input features whose values suffice to ensure $f$ outputs significantly positive predictions, even in the absence of any other information about the rest of the input. 
While the techniques of \citet{erasure} and \citet{Fong17} produce rationales that remain highly dependent on the rest of the input outside of the selected feature subset, 
each rationale identified by our SIS approach is independently considered by $f$ as an entirely sufficient justification for a particular decision in the absence of other information.

\section{Methods}
\label{sec:methods}

Our approach to rationalizing why a particular black-box decision is reached only applies to input examples $\mathbf{x} \in \mathcal{X}$ that meet the decision criterion ${f(\mathbf{x}) \ge \tau}$.  
For such an input $\mathbf{x}$, we aim to identify a SIS-collection of disjoint feature subsets $S_1,\dots, S_K \subseteq [p]$ that satisfy the following criteria: 
\vspace*{-1.0em}  
\begin{enumerate}
\item[(1)] $f(\mathbf{x}_{S_k}) \ge \tau$ for each $k =1,\dots, K$
\\[-1.5em]
\item[(2)] There exists no feature subset $S' \subset S_k$ for some $k = 1,\dots,K$ such that $f(\mathbf{x}_{S'}) \ge \tau$
\\[-1.5em]
\item[(3)] ${f(\mathbf{x}_{R}) < \tau}$ for $R = [p] \hspace*{0.5mm} \setminus \hspace*{0.2mm}  \bigcup_{k=1}^K S_k$ (the remaining features outside of the SIS-collection)
\end{enumerate}
\vspace*{-1.0em}  

\begin{figure*}[t!]
\centering
    \fbox{\includegraphics[width=0.9\textwidth]{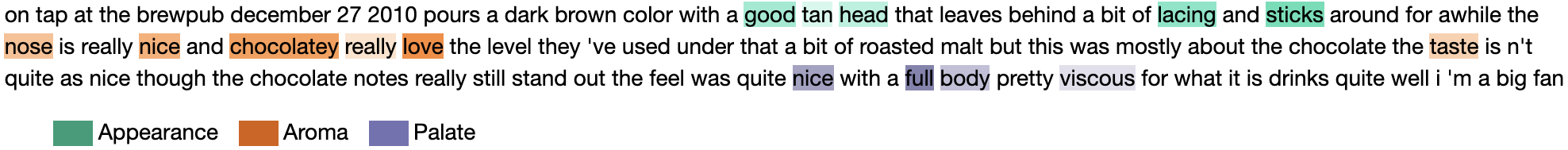}}
    \vspace*{-0.5em}
    \caption[Beer review with SIS for each aspect]{Beer review with one sufficient input subset identified for the prediction of each aspect.}
    \label{fig:beer-all-aspects-example}

\vspace*{0.8em}
    \centering
    \fbox{\includegraphics[width=0.9\textwidth]{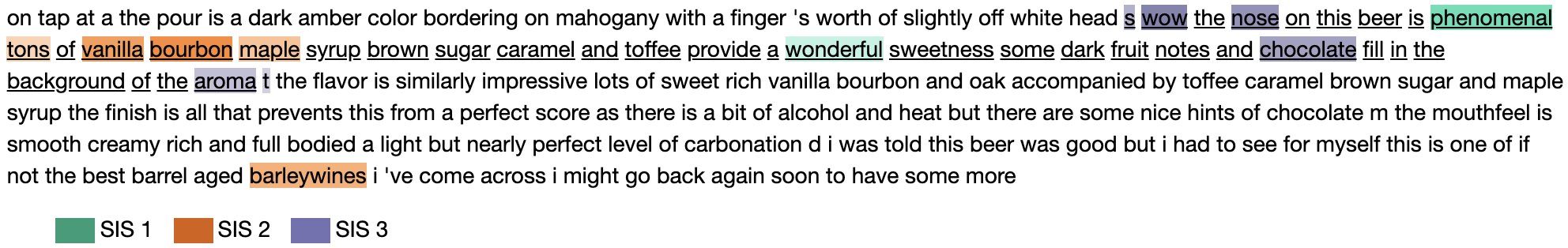}}
    \vspace*{-0.5em}
    \caption[Beer review with 3 SIS subsets]{Beer review with three disjoint SIS $S_1, S_2, S_3$ identified for a positive aroma prediction. Underlined are sentences that human labelers manually annotated as capturing the aroma sentiment.}
    \label{fig:beer-aroma-multisis}
\end{figure*}

Criterion (1) ensures that for any SIS $S_k$, the values of the features in this subset alone suffice to justify the decision in the absence of any information regarding the values of the other features.  To ensure information that is not vital to reach the decision is not included within the SIS, criterion (2) encourages each SIS to contain a minimal  number of features, which facilitates interpretability.  Finally, we require that our SIS-collection satisfies a notion of completeness via criterion (3), which states that the same decision is no longer reached for the input after the entire SIS-collection has been masked. This implies the remaining feature values of the input no longer contain sufficient evidence for the same decision.
Figures~\ref{fig:beer-aroma-multisis} and \ref{fig:95sises} show SIS-collections found in text/image inputs.

Recall that $\mathbf{x}_S \in \mathcal{X}$ denotes a modified input in which the information about the values of features outside subset $S$ is considered to be missing.    We construct $\mathbf{x}_S$ as new input whose values on features in $S$ are identical to those in the original  $\mathbf{x}$, and whose remaining features $x_i \in [p] \setminus S$ are each replaced by a special mask $z_i \in \mathbb{R}^{d_i}$ used to represent a missing  observation.  While certain models are specially adapted to handle inputs with missing observations \citep{Smola05}, this is generally not the case.  To ensure our approach is applicable to all models, we draw inspiration from data imputation techniques which are a common way to represent missing data \citep{rubin1976inference}.  

Two popular strategies include hot-deck imputation, in which unobserved values are sampled from their marginal feature distribution, and mean imputation, in which each $z_i$ simply fixed to the average value of feature $i$ in the data.  Note that for a linear model, these two strategies are expected to produce an identical change in prediction ${f(\mathbf{x}) - f(\mathbf{x}_S)}$. We find in practice that the change in predictions resulting from either masking strategy is roughly equivalent even for nonlinear models such as neural networks (Figure~\ref{fig:beer-aroma-mean-embedding}).
In this work, we favor the mean-imputation approach over sampling-based imputation, which would be computationally-expensive and nondeterministic (undesirable for facilitating interpretability).  
One may also view $\mathbf{z}$ as the \emph{baseline} input value used by feature attribution methods \citep{Sundararajan17,deeplift}, a value which should not lead to particularly noteworthy decisions. Since our  interests primarily lie in rationalizing atypical decisions, the average input arising from mean imputation serves as a suitable baseline.
Zeros have also been used to mask image/categorical data \citep{erasure}, but empirically, this mask appears undesirably more informative than the mean (predictions  more affected by zero-masking).

For an arbitrarily complex function $f$ over inputs with many features $p$, the combinatorial search to identify sets which satisfy  objectives (1)-(3) is computationally infeasible. 
To find a SIS-collection in practice, we employ a straightforward backward selection strategy, which is here applied separately on an example-by-example basis (unlike standard statistical tools which perform backward selection globally to find a fixed set of features for all inputs).  The \textbf{SIScollection} algorithm details our straightforward procedure to identify disjoint SIS subsets that satisfy (1)-(3) approximately (as detailed in \S\ref{sec:properties}) for an input $\mathbf{x} \in \mathcal{X}$ where ${f(\mathbf{x}) \ge \tau}$.

\begin{figure*}[tb]
\IncMargin{1em}
\begin{minipage}[t]{0.32 \textwidth}
  \vspace{0pt}
\begingroup
\removelatexerror
  \begin{algorithm}[H]
  \DontPrintSemicolon 
  \NoCaptionOfAlgo
    \caption{ \ \textbf{SIScollection}($f$, $\mathbf{x}$, $\tau$) }
        $S = [p]$ \; 
        \For{$k=1,2,\dots$}{
       \vspace*{-0.2em} ${R = \text{\textbf{BackSelect}}(f, \mathbf{x}, S)}$ \; \vspace*{-0.1em}
       ${S_k =  \text{\textbf{FindSIS}}(f, \mathbf{x}, \tau, R)}$  \; \vspace*{-0.1em}
       $S \leftarrow S \setminus S_k$ \;
      
\hspace*{-3.5mm} \mbox{ \textbf{if} \hspace*{-2mm} $f(\mathbf{x}_{S}) \hspace*{-0.7mm} < \hspace*{-0.7mm} \tau$: \hspace*{-1.5 mm} \textbf{return} \hspace*{-1.6mm} $S_1$,...,$S_{k-1}$ }
}
 \end{algorithm}
 \endgroup
\end{minipage}
\hspace*{0.01 \textwidth}
\begin{minipage}[t]{0.32 \textwidth}
  \vspace{0pt}
  \begingroup
\removelatexerror
  \begin{algorithm}[H]
  \DontPrintSemicolon 
  \NoCaptionOfAlgo
    \caption{ \ \textbf{BackSelect}($f$, $\mathbf{x}$, $S$) }
        $R = $ empty stack \;
    \vspace*{-0.1em} \While {$S \neq \varnothing$} { 
    \vspace*{-0.2em} $ i^* = \argmax_{i \in S} f(\mathbf{x}_{S \setminus \{i \} })$ \;
    \vspace*{-0.1em}
   Update $S \leftarrow S \setminus \{i^*\}$ \;
   Push $i^*$ onto top of $R$ \;
   }
\vspace*{-0.2em} \Return $R$ \;
 \end{algorithm}
\endgroup
\end{minipage}
\hspace*{0.01 \textwidth}
\begin{minipage}[t]{0.32 \textwidth}
  \vspace{0pt}
\begingroup
\removelatexerror
  \begin{algorithm}[H]
    \DontPrintSemicolon
    \NoCaptionOfAlgo
    \caption{ \ \textbf{FindSIS}($f$, $\mathbf{x}$, $\tau$, $R$) }
    $S = \varnothing$ \;
     \vspace*{-0.1em} \While {$f(\mathbf{x}_{S}) < \tau$ } {
    \vspace*{-0.2em} Pop $i$ from top of $R$ \;
    Update $S \leftarrow S \cup \{i\}$ \;
    }
   \vspace*{-0.2em} \textbf{if} \hspace*{-1mm} $f(\mathbf{x}_{S}) \ge \tau$: \ \textbf{return } $S$  \;
    \textbf{else}: \ \textbf{return} \emph{None} \;
  \end{algorithm}
\endgroup
\end{minipage}
\DecMargin{1em}
\end{figure*}

Our overall strategy is to find a SIS subset $S_k$ (via \textbf{BackSelect} and \textbf{FindSIS}), mask it out, and then repeat these two steps restricting each search for the next SIS solely to features disjoint from the currently  found SIS-collection $S_1,\dots, S_k$, until the decision of interest is no longer supported by the remaining feature values.
In the \textbf{BackSelect} procedure, $S \subset [p]$ denotes the set of remaining unmasked features that are to be considered during backward selection.  
For the current subset $S$, step 3 in \textbf{BackSelect} identifies which remaining feature $i \in S$ produces the \emph{minimal} reduction in $f(\mathbf{x}_S) - f(\mathbf{x}_{S \setminus \{i\}})$ (meaning it least reduces the output of $f$ if additionally masked), a question trivially answered by running each of the remaining possibilities through the model.  This strategy aims to gradually mask out the least important features in order to reveal  the core input pattern that is perceived by the model as sufficient evidence for its decision.  
Finally, we build our SIS up from the last $\ell$ features omitted during the backward selection, selecting a $\ell$ value just large enough to meet our sufficiency criterion (1).  Because this approach always queries a prediction over the joint set of remaining features $S$, it is better suited to account for interactions between these features and ensure their sufficiency (i.e.\ that $f(\mathbf{x}_S) \ge \tau$) compared to a forward selection in the opposite direction which builds the SIS upwards one feature at a time by greedily maximizing marginal gains.  Throughout its execution,  \textbf{BackSelect} attempts to maintain the sufficiency of $\mathbf{x}_{S}$ as the set $S$ shrinks.

\subsection{Properties of the SIS-collection}
\label{sec:properties}

Given $p$ input features, our algorithm requires $\mathcal{O}(p^2k)$ evaluations of $f$ to identify $k$ SIS, but we can achieve $\mathcal{O}(pk)$ by parallelizing each argmax in \textbf{BackSelect} (e.g.\ batching on GPU).
Throughout, let ${S_1,\minidots, S_K}$ denote the output of  \textbf{{SIScollection}} when applied to a given input $\mathbf{x}$ for which $f(\mathbf{x}) \ge \tau$.  Disjointness of these sets is crucial to ensure computational tractability and that the number of SIS per example does not grow huge and hard to interpret.
Proposition \ref{prop:min} below proves that each SIS produced by our procedure will satisfy an approximate notion of minimality.  
Because we desire minimality of the SIS as specified by (2), it is not appropriate to terminate the backward elimination in \textbf{BackSelect} as soon as the sufficiency condition $f(\mathbf{x}_S) \ge \tau$ is violated, due to the possible presence of local minima in $f$  along the path of subsets encountered during backward selection 
(as shown in Figure \ref{fig:mnist-local-minimum}). 

Proposition \ref{prop:complete} additionally guarantees that masking out the entirety of the feature values in the SIS-collection will ensure the model makes a different decision.  Given $f{(\mathbf{x}) \ge \tau}$, it is thus necessarily the case that the observed values responsible for this decision lie within the SIS-collection ${S_1,\dots, S_K}$.  
We point out that for an easily reached decision, where $f(\mathbf{z}) \ge \tau$ (i.e.\ this decision is reached even for the average input), our approach will not output any SIS.  Because this same decision would likely be anyway reached for a vast number of inputs in the training data (as a sort of default decision), it is conceptually difficult to grasp what particular aspect of the given $\mathbf{x}$ is responsible.

\vspace*{0.1em}
\begin{prop} There exists no feature $i$ in any set $S_1, \dots, S_K$ that can be  additionally masked while retaining sufficiency of the resulting subset (i.e.\ ${f(\mathbf{x}_{S_k \setminus \{i\}}) < \tau}$ for any $k = 1,\minidots,K, i \in S_k$).  Also, among all subsets $S$ considered during the backward selection phase used to produce $S_k$, this set has the smallest cardinality of those which satisfy ${f(\mathbf{x}_{S}) \ge \tau}$. 
\label{prop:min}
\end{prop}

\vspace*{0.1em}
\begin{prop}  For $\mathbf{x}_{[p] \setminus S^*}$, modified by masking all features in the entire SIS-collection  ${S^* = \bigcup_{k=1}^K S_k}$, we must have:  ${f(\mathbf{x}_{[p] \setminus S^*}) < \tau}$ when $S^* \neq [p]$.
\label{prop:complete}
\end{prop}


Unfortunately, nice assumptions like convexity/submodularity are inappropriate for estimated functions in ML.  We present various simple forms of practical decision functions for which our algorithms are guaranteed to produce desirable explanations.  
Example 1 considers interpreting functions of a generalized linear form, Examples 2 \& 3 describe functions whose operations resemble generalized logical \emph{OR} \& \emph{AND} gates, and Example 4 considers functions that seek out a particular input pattern.  
Note that 
features ignored by $f$ are always masked in our backward selection and thus never appear in the resulting SIS-collection.

\vspace*{0.1em}
\begin{ex}
Suppose the input data are vectors and  $\displaystyle {f(\mathbf{x}) = g(\beta^T \mathbf{x} + \beta_0})$, where $g$ is monotonically increasing.  We also presume $\tau > g(\beta_0)$ and the data were  centered such that each feature has mean zero (for ease of notation).
In this case, ${S_1,\minidots, S_K}$ must satisfy criteria (1)-(3). $S_1$ will consist of the features whose indices correspond to the largest $\ell$ entries of ${\{\beta_1 x_1 , \minidots,  \beta_p x_p \}}$
for some suitable $\ell$ that depends on the value of $\tau$.  
It is also guaranteed that $f(\mathbf{x}_{S_1}) \ge f(\mathbf{x}_{S})$ for any subset $S \subseteq [p]$ of the same cardinality  $|S| = \ell$.  
For each individual feature $i$ where $g(\beta_i x_i + \beta_0) \ge \tau$, there will be exist a corresponding SIS $S_k$ consisting only of $\{ i \}$.  
No SIS will include features whose coefficient $\beta_i = 0$, 
or those whose difference between the observed and average value $z_i$ ($=0$ here) is of an opposite sign than the corresponding model coefficient (i.e.\ $\beta_i (x_i - z_i) \le 0$).  
\end{ex}
  
\vspace*{0.1em}
\begin{ex} Let $\displaystyle f(\mathbf{x}) = \max\{g_1(\mathbf{x}_{S'_1}), \dots, g_{L}(\mathbf{x}_{S'_{L}})\}$ for some disjoint $S'_1,\minidots, S'_{L} \subset [p]$ and functions $g_1,\minidots, g_{L}$, such that for the given  $\mathbf{x}$ and threshold $\tau$: 
${g_1(\mathbf{x}_{S'_1}) >  \dots > g_{L}(\mathbf{x}_{S'_{L}})} \ge \tau$ and  ${g_k(\mathbf{x}_{S'_k \setminus \{i\}}) < \tau}$ for each ${1 \le k \le {L}, i \in S'_k}$.  Such $f$ might be functions that model strong interactions between the features in each $S_k$ or look for highly specific value patterns to occur these subsets.
In this case, \textbf{\emph{SIScollection}} will return $L$ sets such that  ${S_1 = S'_1, S_2 = S'_2, \dots, S_{L} = S'_{L}}$.  
\end{ex}

\vspace*{0.1em}
\begin{ex} If $\displaystyle f(\mathbf{x}) = \min\{g_1(\mathbf{x}_{S'_1}), \dots, g_{L}(\mathbf{x}_{S'_{L}})\}$  and the same conditions from Example 2 are met, then \textbf{\emph{SIScollection}} will return a single set $S_1  = \bigcup_{k=1}^{L} S'_k$.
\end{ex}

\vspace*{0.1em}
\begin{ex} Suppose $\displaystyle \mathbf{x} \in \mathbb{R}^p$ with $\displaystyle f(\mathbf{x}) = h(|| \mathbf{x}_S - \mathbf{c}_S ||)$ where $h$ is monotonically decreasing and $\mathbf{c}_S$ specifies a fixed pattern of input values for features in a certain subset $S$.  
For input $\mathbf{x}$ and threshold choice $\tau = f(\mathbf{x})$, \textbf{\emph{SIScollection}} will return a single set ${S_1 = \{i \in S : |x_i - c_i| < | z_i - c_i |\} }$.
\end{ex} 

\begin{figure*}[h!]
\vspace*{-0.7em}
    \centering
    \begin{minipage}[t]{0.48\textwidth}
        \centering
        \includegraphics[width=1.0\textwidth]{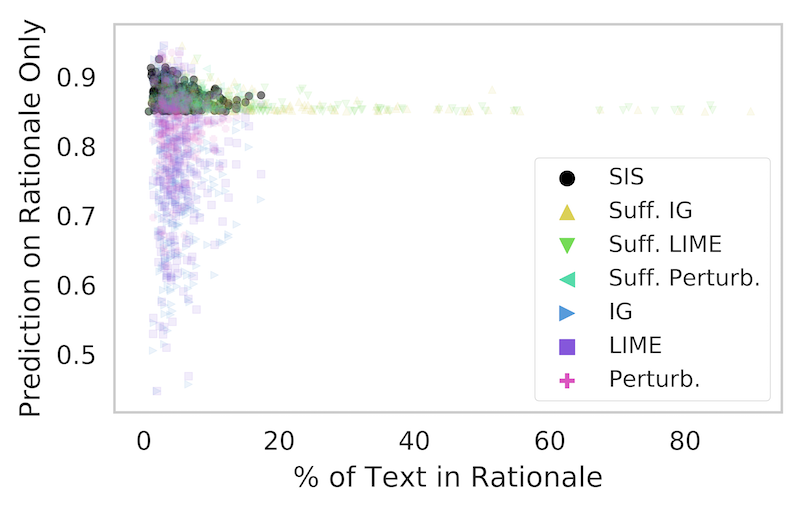}
        \vspace*{-2.0em}
        \caption[Prediction on rationales only vs.\ rationale length (aroma prediction)]{Prediction on rationales only vs.\ rationale length for various methods in reviews  with positive aroma prediction ($\tau = 0.85$).}
        \label{fig:beer-aroma-prediction-vs-length}
    \end{minipage}
    \hfill
    \begin{minipage}[t]{0.48\textwidth}
        \centering
        \includegraphics[width=1.0\linewidth]{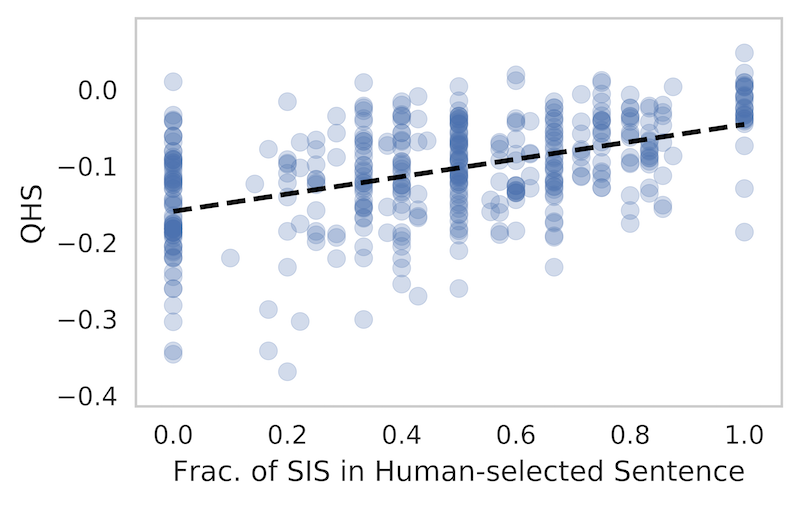}
        \vspace*{-2.0em}
        \caption[QHS vs.\ SIS-annotation similarity]{QHS vs.\ similarity between SIS \& annotation in the reviews with positive aroma sentiment (Pearson $\rho = 0.491$, $p$-value $= 1.5\mathrm{e}{-25}$).}
        \label{fig:beer-aroma-quality-annotations}
    \end{minipage}
\end{figure*}

\section{Results}
\label{sec:results}



We apply our methods to analyze neural networks for text, DNA, and image data. 
\textbf{SIScollection} is compared with alternative subset-selection methods for producing rationales (see descriptions in Supplement \S\ref{sec:alternativedets}).  
Note that our \textbf{BackSelect} procedure determines an ordering of elements, $R$,  subsequently used to construct the SIS.
Depictions of each SIS are shaded based on the feature order in $R$ (darker = later), which can indicate relative feature importance within the SIS.

In the ``Suff. IG,'' ``Suff. LIME,'' and ``Suff. Perturb.'' (\textit{sufficiency constrained}) methods, we instead compute the ordering of elements $R$ according to the feature attribution values output by integrated gradients \citep{Sundararajan17}, LIME \citep{lime}, or a perturbative approach that measures the change in prediction when  individually masking each feature (see \S\ref{sec:alternativedets}).
The rationale subset $S$ produced under each method is subsequently assembled using \textbf{FindSIS} exactly as in our approach and thus is guaranteed to satisfy $f(\mathbf{x}_S) \ge \tau$.
In the ``IG,'' ``LIME,'' and ``Perturb.'' (\textit{length constrained}) methods, we use the same previously described ordering $R$, but always select the same number of features in the rationale as in the SIS produced by our method (per example).
We also compare against the additional ``Top IG'' method, in which top features from $R$ are added into the rationale until sum of integrated gradients attributions suggests that the rationale has met our sufficiency criterion (see \S\ref{sec:alternativedets}).

\begin{figure*}[h!]
\centering
    \fbox{\includegraphics[width=0.98\textwidth]{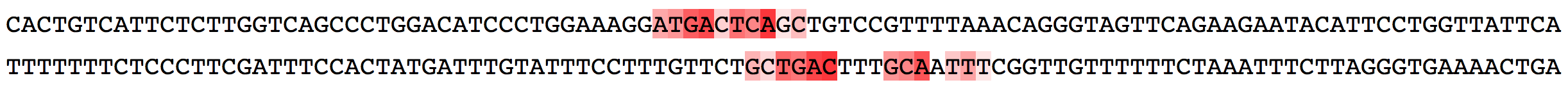}}
    \vspace*{-0.2em}
    \caption[Example DNA sequences]{Two DNA sequences that receive positive TF binding predictions for the MAFF factor (SIS is shaded).
    }
    \label{fig:tf-binding-seq-examples}

 \vspace*{-0.8em}
\centering
\begin{subfigure}[t]{.325\textwidth}
  \centering
      \subcaption{}
  \includegraphics[width=\textwidth]{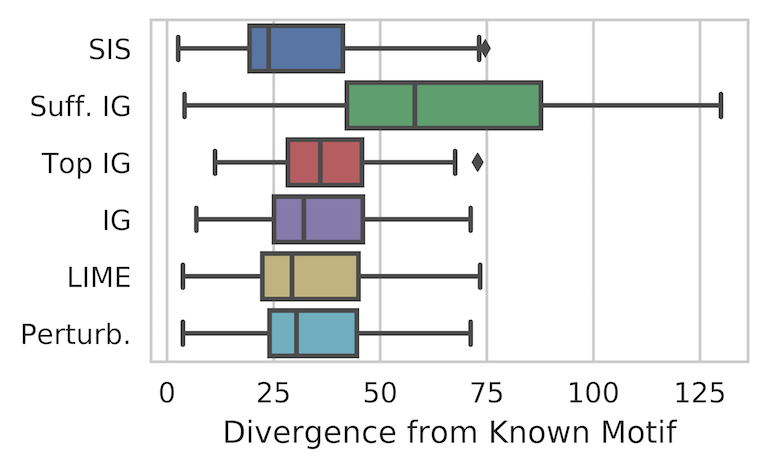}
    \label{fig:tf-motif-divergence}
\end{subfigure}%
\hspace*{0.05\textwidth}
  \begin{subfigure}[t]{.18\textwidth}
    \centering
      \subcaption{}
    \begingroup  
    \tiny
    \begin{tabular}{cc}
      SIS & Freq.  \\
      \midrule
      \texttt{GCTGAGTCAT} & 197  \\
      \texttt{ATGACTCAGC} & 185  \\
      \texttt{GCTGAGTCA-C} & 83  \\
      \texttt{GCTGAGTCAC} & 53  \\
      \texttt{GCTGACTCAGCA} & 42  \\
      &  \\[-0.8em]  
      SIS & Freq.  \\
      \midrule
      \texttt{TGCTGA----GCA-TTT} & 12  \\
      \texttt{GCTGAC---GCA-TTT} & 8  \\
      \texttt{TGCTGAC---GCA-TT} & 6  \\
      \texttt{TGCTGAC---GCA-AA} & 5  \\
      \texttt{TGCTGAC---GCA-AT} & 4  \\
    \end{tabular}
    \endgroup
  \label{fig:tf-clustering-clusters}
\end{subfigure}
\hspace*{0.05\textwidth}
\begin{subfigure}[t]{.31\textwidth}
  \centering
    \caption{}
  \includegraphics[width=\textwidth]{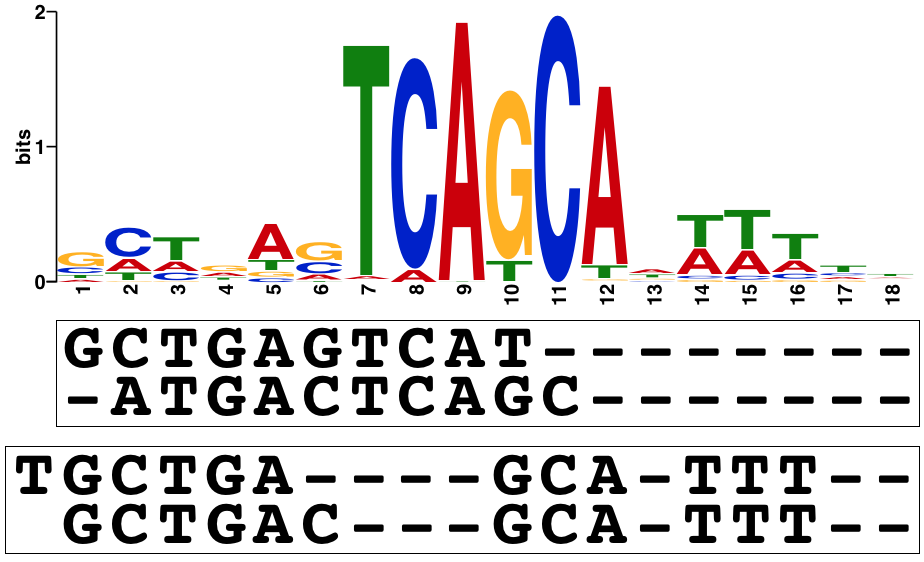}
  \label{fig:tf-clustering-logo}
\end{subfigure}%
\vspace*{-1.6em}
\caption[KL divergence for TF motifs and rationales]{\textbf{(a)} KL divergence between JASPAR motifs (known ground truth) and rationales found via various  methods. Shown are results for 422 TF datasets (each one summarized by median divergence).
\textbf{(b)}  In the SIS found in data from one TF, DBSCAN identified two clusters (most frequently-occurring SIS in each  shown). \\
\textbf{(c)} Known JASPAR motif (top) and alignment with cluster modes (bottom).
}
\label{fig:tf-clustering}
\end{figure*}

\subsection{Sentiment Analysis of Reviews}
\label{sec:sentanalysis}
\vspace*{-0.5em}

We first consider a dataset of beer reviews from \citet{mcauley2012learning}. 
Taking the text of a review as input, different LSTM networks \citep{lstm} are trained 
to predict user-provided numerical ratings of aspects like aroma, appearance, and palate (details in \S\ref{sec:beerdets}).
Figure~\ref{fig:beer-all-aspects-example} shows a sample beer review where we highlight the SIS identified for the LSTM that predicts each aspect. Each SIS only captures sentiment toward the relevant aspect.
Figure~\ref{fig:beer-aroma-multisis} depicts the SIS-collection identified from a review the LSTM decided to flag for positive aroma.

Figure~\ref{fig:beer-aroma-prediction-vs-length} shows that when the alternative methods described in \S\ref{sec:results} are length constrained, the rationales they produce often badly fail to meet our sufficiency criterion.  Thus, even though the same number of feature values are preserved in the rationale and these alternative methods select the features to which they have assigned the largest attribution values, their rationales lead to  significantly reduced $f$ outputs compared to our SIS subsets.  
If the sufficiency constraint is instead enforced for these alternative methods, the rationales they identify become significantly larger than those produced by  \textbf{SIScollection}, and also contain many more unimportant features (Table~\ref{table:beer-stats}, Figure~\ref{fig:beer-aroma-perturbation}).

\begin{figure}
    \centering
    \includegraphics[width=1.0\linewidth]{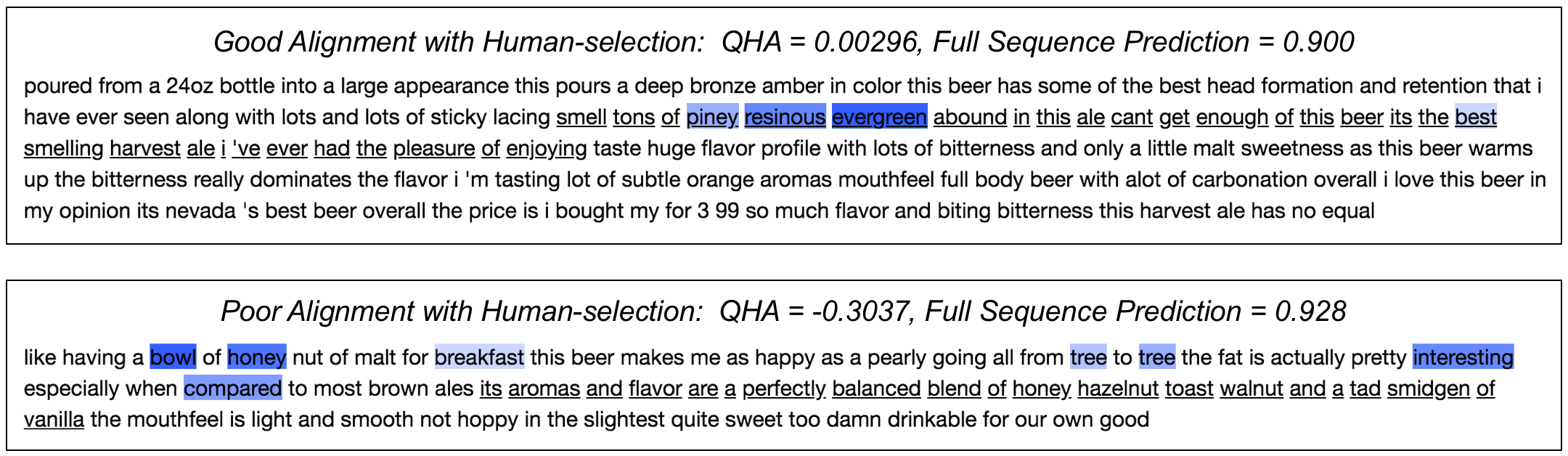}
    \caption[Alignment of human rationales for beer reviews with predictive model]{Beer reviews (aroma) in which human-selected sentences (underlined) are aligned well (top) and poorly (bottom) with predictive model. Fraction of SIS in the human sentences corresponds accordingly. In the bottom example (poor alignment between human-selection and predictive model), our procedure has surfaced a case where the LSTM has learned features that diverge from what a human would expect (and may suggest overfitting).}
    \label{fig:beer-aroma-annots-examples}
\end{figure}

Benchmarking interpretability methods is difficult because a learned $f$ may behave counterintuitively such that seemingly unreasonable model explanations are in fact faithful descriptions of a model's decision-making process.  For some reviews, a human annotator has manually selected which sentences carry the relevant sentiment for the aspect of interest, so we treat these annotations as an alternative rationale for the LSTM prediction. For a review $\mathbf{x}$ whose true and predicted aroma exceed our decision threshold, we define the \emph{quality of human-selected sentences for model explanation} $\qhs = f(\mathbf{x}_{S}) - f(\mathbf{x})$ where $S$ is the human-selected-subset of words in the review (see examples in Figure~\ref{fig:beer-aroma-annots-examples}).
High variability of $\qhs$ in the annotated reviews (Figure~\ref{fig:beer-aroma-quality-annotations}) indicates the human rationales often do not contain sufficient information to preserve the LSTM's decision.  Figure~\ref{fig:beer-aroma-quality-annotations} shows the LSTM makes many decisions based on different subsets of the text than the parts that humans find appropriate for this task. 
Reassuringly, our SIS more often lie within the selected annotation for reviews with high $\qhs$ scores. 





\subsection{Transcription Factor Binding}
\label{sec:tf-binding}
\vspace*{-0.5em}

We next analyze convolutional neural networks (CNN) used to classify whether a given  transcription factor (TF) will bind to a specific DNA sequence \citep{zeng2016convolutional}.
From 422 different datasets of DNA sequences bound-or-not by different TFs (and 422 different CNN models), we extract SIS-collections from sequences with high (top 10\%) predicted binding affinity for the TF profiled in each dataset (details in \S\ref{sec:tfdets}).
Figure~\ref{fig:tf-binding-seq-examples} depicts two input examples and the corresponding identified SIS.
Again, rationales produced via our SIS approach are shorter and better at preserving large $f$-values than rationales from other methods (Figures \ref{fig:tf-binding-rationale-lengths} and \ref{fig:tf-binding-prediction-vs-length}).

\begin{figure*}[htbp!]
    \begin{minipage}[t]{0.67\textwidth}
    \vspace{0pt} \centering
    \includegraphics[width=1.0\textwidth]{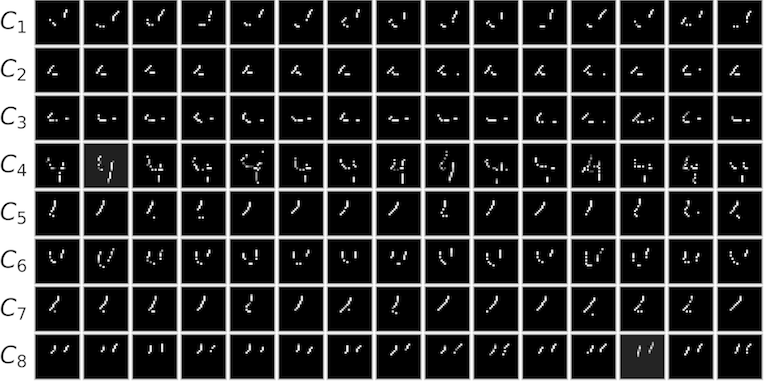}
    \vspace*{-1.6em}
    \caption[SIS clusters for digit 4]{Eight clusters of SIS identified from examples of digit 4. \\ Each row contains fifteen random SIS from a single cluster.}
    \label{fig:mnist-clustering-4-maintext}
    \end{minipage}
   \hfill
    \begin{minipage}[t]{0.3\textwidth}
     \begin{subfigure}[t]{\textwidth}
     \vspace{0pt}
    \centering
    \includegraphics[width=0.87\textwidth]{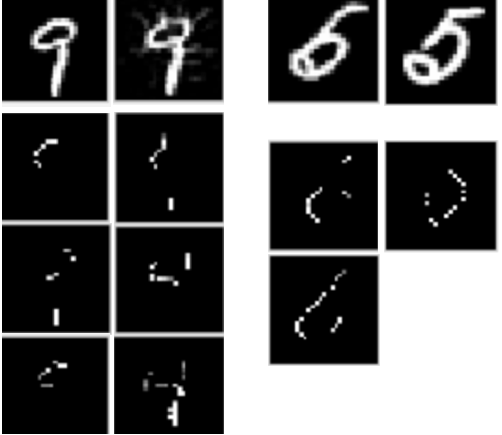} \\
     \vspace*{-2.1em} 
      (a) \hspace*{0.38\textwidth} (b) \\
      \vspace*{0.58em} 
      \hspace*{0.12\textwidth} (a) \hspace*{0.6\textwidth} \\
    \label{fig:mnist-ad-misclass}
    \end{subfigure} 
    \vspace*{-0.7em}
    \caption[SIS-collections for adversarial and misclassified MNIST digits]{\textbf{(a)} SIS for correctly classified 9 (1st column) and when adversarially perturbed toward class 4 (2nd column).  \textbf{(b)} SIS for digits 5 that are misclassified as 6 (1st column) and as 0 (2nd column).}
    \label{fig:95sises}
    \end{minipage} 
\end{figure*}

To predict binding so accurately, the CNN must faithfully reflect the biological mechanisms that relate the DNA sequence to the probability of TF occupancy.
We evaluate the rationales found by our methods against known TF binding motifs from JASPAR \citep{jaspar}, adopting KL divergence between the known motif and each proposed rationale as a quality measure (see \S\ref{sec:tf-rationale-analysis}).
Figure \ref{fig:tf-motif-divergence} shows the divergence of rationales produced by  \textbf{SIScollection} is significantly lower than that of rationales identified using other methods (Wilcoxon $p \leq 1\mathrm{e}{-5}$ in all cases).  
SIS is thus more effective at uncovering the underlying biological principles than the alternative methods we applied.



\subsection{MNIST Digit Classification}
\label{sec:mnist}
\vspace*{-0.5em}

Finally, we study a 10-way CNN classifier trained on the 
MNIST handwritten digits data \citep{mnist}.  
Here, we only consider predicted probabilities for one class of interest at a time and always set $\tau = 0.7$ as the probability threshold for deciding that an image belongs to the class. 
We extract the SIS-collection from all corresponding test set examples (details in \S\ref{sec:mnistdets}). 
Example images and corresponding SIS-collections are shown in Figures \ref{fig:95sises} and \ref{fig:mnist-sis-examples-all-digits}. Figure \ref{fig:95sises}a illustrates how the SIS-collection drastically changes for an example of a correctly-classified 9 that has been adversarially manipulated \citep{carlini2017towards} to become confidently classified as the digit 4. 
Furthermore, these SIS-collections immediately enable us to understand why certain misclassifications occur (Figure \ref{fig:95sises}b).

\subsection{Clustering SIS for General Insights}
\label{sec:clustering-insights}
\vspace*{-0.5em}
Identifying the different input patterns that justify a decision can help us better grasp the general operating principles of a model.  To this end, we cluster all of the SIS produced by \textbf{SIScollection} applied across a large number of data examples that received the same decision. Clustering is done via DBSCAN, a widely applicable algorithm that merely requires specifying pairwise distances between points \citep{dbscan}.

We first apply this procedure to the SIS found across all test-set DNA sequences which our CNN model predicted would be bound by some TF.
Here, the pairwise distance between two sufficient input subsets is taken to be the Levenshtein (edit) distance. 
Figure~\ref{fig:tf-clustering} shows the clusters for a particular TF where two SIS clusters were found.  Despite no contiguity being enforced in our algorithm, each cluster is comprised of short sequences that clearly capture different aspects of the underlying DNA motif known to bind this TF.

\begin{figure}[tb!]
\vspace*{-1em}
    \centering
    \includegraphics[width=1.0\linewidth]{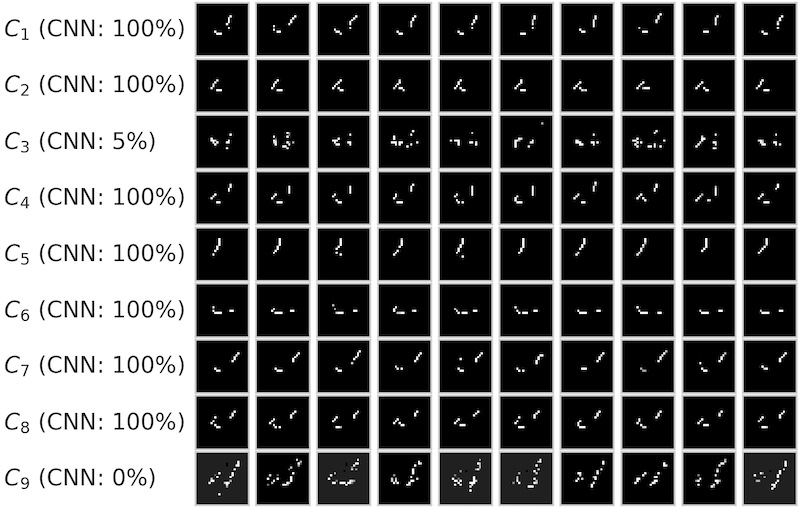}
    \vspace*{-1.6em}
    \caption[Joint clustering MNIST digit 4 SIS from CNN \& MLP]{Jointly clustering the MNIST digit 4 SIS from CNN and MLP. We list the percentage of SIS in each cluster stemming from the CNN (rest from MLP).}
    \label{fig:mnist-model-differences-clustering-digit4}
    \vspace*{-1em}
\end{figure}

\begin{figure*}[t!]
    \centering
    \begin{subfigure}[t]{0.5\textwidth}
        \centering
        \subcaption{}
        \vspace*{-0.6em}
        \includegraphics[width=\textwidth]{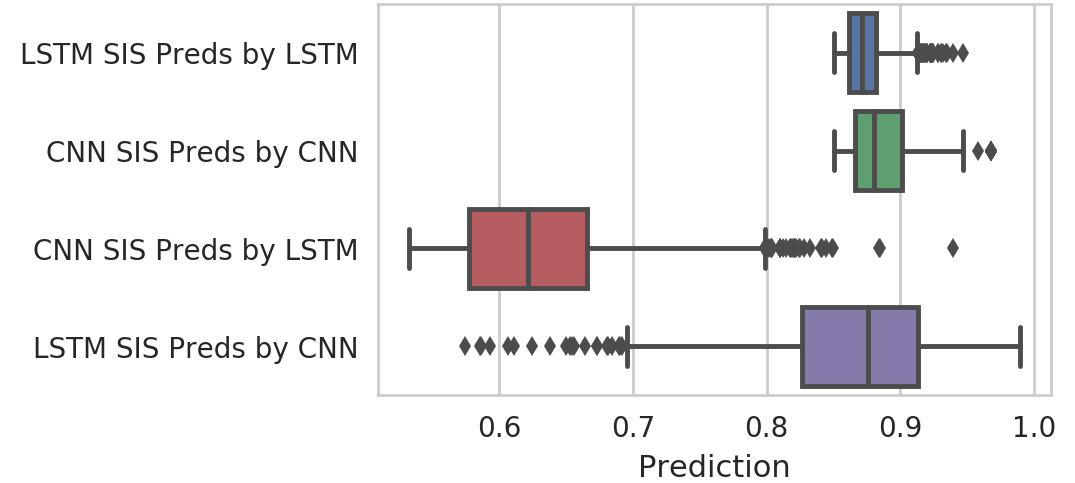}
        \label{fig:beer-aroma-lstm-cnn-sis-preds}
    \end{subfigure}%
    \begin{subfigure}[t]{0.5\textwidth}
        \centering
        \subcaption{}
        \vspace*{-0.6em}
        \includegraphics[width=\textwidth]{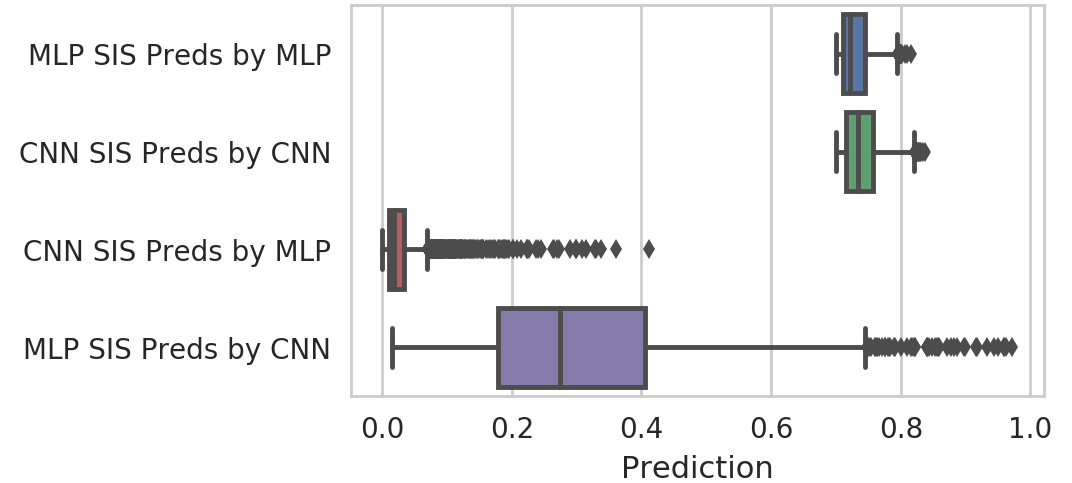}
        \label{fig:mnist-cnn-mlp-sis-preds}
    \end{subfigure}%
    \vspace*{-1.6em}
    \caption[Predictions on the SIS from alternative models on beer reviews and MNIST digits]{Predictions by one model on the SIS extracted from the other model in: \textbf{(a)} beer reviews with positive LSTM/CNN aroma predictions, and \textbf{(b)} MNIST digits confidently classified as 4 by CNN/MLP.} 
    \label{fig:sis-preds-by-other-models}
\end{figure*}

We also apply DBSCAN clustering to the SIS found across all MNIST test-examples confidently identified by the CNN as a particular class. Pairwise distances are here defined as the \emph{energy distance} \citep{energydistance}
over pixel locations between two SIS subsets (see \S\ref{sec:energy}). 
Figure~\ref{fig:mnist-clustering-4-maintext} depicts the SIS clusters identified for digit 4 (others in Figure~\ref{fig:mnist-all-clusters}).
These reveal distinct feature patterns learned by the CNN to distinguish 4 from other digits, which are clearly present in the vast majority of test set images confidently classified as a 4. For example, cluster $C_8$ depicts parallel slanted lines, a pattern that never occurs in  other digits.

Subsequently, we cluster the SIS found across held-out beer reviews (Test-Fold in Table \ref{table:beerdets-dataset-stats}) that received positive aroma predictions from our LSTM network. The distance between two SIS is taken as the Jaccard distance  between their bag of words representations. 
Three clusters depicted in Table~\ref{tab:beer-aroma-clustering} (rest in Tables~\ref{tab:beer-aroma-positives-clustering-supplement},~\ref{tab:beer-aroma-negatives-clustering-supplement}) reveal isolated phrases that the LSTM associates with positive aromas in the absence of other context.


\begin{table*}[tbhp!]
\captionsetup{width=.48\linewidth}
    \centering
    \begin{minipage}[t]{0.495\textwidth} \vspace{0pt}
    \caption[Three SIS clusters from beer reviews]{3 clusters of SIS extracted from beer reviews with positive CNN aroma predictions.  Each row shows 4 most frequent unique SIS in a cluster (each SIS shown as ordered word list with text-positions omitted).  Each unique SIS can be present many times in one cluster.}
    \vspace*{-0.0em}
    \begingroup  
    \footnotesize
    \begin{tabularx}{\linewidth}[t]{|@{\hspace{-0.8em}}X@{\hspace{-0.8em}}|X|X|X|X|}
\hline
\textbf{Clu.\ } & \textbf{SIS \#1} & \textbf{SIS \#2} &   \textbf{SIS \#3} & \textbf{ SIS \#4 }
 \\
\hline
$C_{1}$ &     smell amazing wonderful &                             nice wonderful nose &                  wonderful amazing &                   amazing amazing  \\
\hline
$C_{2}$ &  grapefruit mango pineapple &           pineapple grapefruit pineapple grapefruit &   hops grapefruit pineapple floyds &  mango pineapple incredible  \\
\hline
$C_{3}$ &         creme brulee brulee  &                      creme brulee decadent &          incredible creme brulee &    creme brulee exceptional  \\
\hline
\end{tabularx}
    \endgroup
    \label{tab:beer-aroma-clustering}
    \end{minipage}
    \hfill
    \begin{minipage}[t]{0.495\textwidth} \vspace{0pt}
    \caption[Joint clustering of SIS from LSTM and CNN on beer reviews, aroma aspect]{Joint clustering of the SIS from beer reviews predicted to have positive aroma by LSTM or CNN. 
    Dashes are used in clusters with under 4 unique SIS.
    Percentages quantify SIS per cluster from the LSTM.
    }
    \vspace*{-0.3em}
    \begingroup  
    \footnotesize
    \begin{tabularx}{\linewidth}[t]{|@{\hspace{-0.9em}}X@{\hspace{-0.9em}}|@{\hspace{-0.5em}}X@{\hspace{-0.5em}}|@{\hspace{-0.1em}}X@{\hspace{-0.1em}}|@{\hspace{-0.1em}}X@{\hspace{-0.1em}}|@{\hspace{-0.1em}}X@{\hspace{-0.1em}}|@{\hspace{-0.1em}}X@{\hspace{-0.1em}}|}
    \hline
\textbf{Clu.} &  \textbf{LSTM} &  \textbf{SIS \#1}  &                                  \textbf{SIS \#2} &                               \textbf{SIS \#3} & \textbf{SIS \#4} \\
\hline
$C_{1}$ & 0\% &     delicious &                                             - &          - & - \\
\hline
$C_{2}$  & 0\% &      very nice &                                      - &          - &         - \\
\hline
$C_{3}$  & 20\% &  rich chocolate &                          very rich &   chocolate complex &  smells rich \\
\hline
$C_{4}$  & 33\%  &   oak chocolate &  chocolate raisins raisins oak bourbon  & chocolate oak &   raisins chocolate  \\
\hline
$C_{5}$ & 70\% &   complex aroma  &          aroma complex peaches complex &  aroma complex interesting cherries &   aroma complex  \\
\hline
\end{tabularx}
    \endgroup
  \label{tab:beer-aroma-model-differences-clustering}
  \end{minipage}
\end{table*}

The general insights revealed by our SIS-clustering can also be used to compare the operating-behavior of different models.
For the beer reviews, we also train a CNN to compare with our existing LSTM (see~\S\ref{sec:beer-understanding-differences-dets}). For MNIST, we train a multilayer perceptron (MLP) and compare to our existing CNN (see~\S\ref{sec:mnist-understanding-differences-dets}).
Both networks exhibit similar performance in each task, so it is not immediately clear which model would be preferable in practice.
Figure~\ref{fig:sis-preds-by-other-models} shows the SIS extracted under one model are typically insufficient to receive the same decision from the other model, indicating these models base their positive predictions on different evidence.

Figure \ref{fig:mnist-model-differences-clustering-digit4} depicts results from a joint  clustering of all SIS extracted from held-out MNIST images confidently classified as a 4  by either the MLP or CNN.  Evidently, our MNIST-CNN bases its confidence primarily on spatially-contiguous strokes comprising only a small portion of each digit.  MLP-decisions are in contrast based on pixels located throughout the digit, demonstrating this model relies more on the global shape of the handwriting. 
Thus, the CNN is more susceptible to mistaking other (non-digit) handwritten characters for 4s if they happen to share some of the same strokes.
Table~\ref{tab:beer-aroma-model-differences-clustering} contains results of jointly clustering the SIS extracted from beer reviews with positive aroma predictions under our LSTM or text-CNN.
This CNN tends to learn localized (unigram/bigram) word patterns, while the LSTM identifies more complex multi-word interactions that truly seem more relevant to the target aroma value. 
Many CNN-SIS are simply phrases with universally-positive sentiment, indicating this model is less capable at distinguishing between positive sentiment toward aroma vs.\ other aspects such as taste/look.

\section{Discussion}
\label{sec:dis}

This work introduced the idea of interpreting black-box decisions on the basis of sufficient input subsets -- minimal input patterns that alone provide sufficient evidence to justify a particular decision. 
Our methodology is easy to understand for non-experts, applicable to all ML models without any additional training steps, and remains fully faithful to the underlying model without making approximations.
While we focus on local explanations of a single decision, clustering the SIS-patterns extracted from many data points reveals insights about a model's general decision-making process.
Given multiple models of comparable accuracy, SIS-clustering can uncover critical operating differences, such as which model is more susceptible to spurious training data correlations or will generalize worse to counterfactual inputs that lie outside the data distribution.

 \clearpage 
 
\subsubsection*{Acknowledgements}
We thank Haoyang Zeng and Ge Liu for help with the TF data/models. 
This work was supported by NIH Grants R01CA218094, R01HG008363, and
R01HG008754.

\bibliographystyle{apa}
{ 
\balance
\bibliography{interpretability}

\begin{thebibliography}{}

\bibitem[\protect\astroncite{Bach et~al.}{2015}]{Bach15}
Bach, S., Binder, A., Montavon, G., Klauschen, F., M{\"u}ller, K.-R., and
  Samek, W. (2015).
\newblock On pixel-wise explanations for non-linear classifier decisions by
  layer-wise relevance propagation.
\newblock {\em PloS One}, 10(7):e0130140.

\bibitem[\protect\astroncite{Baehrens et~al.}{2010}]{Baehrens10}
Baehrens, D., Schroeter, T., Harmeling, S., Kawanabe, M., Hansen, K., and
  M{\"u}ller, K.-R. (2010).
\newblock How to explain individual classification decisions.
\newblock {\em Journal of Machine Learning Research}, 11:1803--1831.

\bibitem[\protect\astroncite{Carlini and Wagner}{2017}]{carlini2017towards}
Carlini, N. and Wagner, D. (2017).
\newblock Towards evaluating the robustness of neural networks.
\newblock In {\em IEEE Symposium on Security and Privacy}.

\bibitem[\protect\astroncite{Caruana et~al.}{2015}]{Caruana15}
Caruana, R., Lou, Y., Gehrke, J., Koch, P., Sturm, M., and Elhadad, N. (2015).
\newblock Intelligible models for healthcare: Predicting pneumonia risk and
  hospital 30-day readmission.
\newblock In {\em ACM SIGKDD International Conference on Knowledge Discovery
  and Data Mining}.

\bibitem[\protect\astroncite{Chen et~al.}{2018}]{Chen18}
Chen, J., Song, L., Wainwright, M.~J., and Jordan, M.~I. (2018).
\newblock Learning to explain: An information-theoretic perspective on model
  interpretation.
\newblock In {\em International Conference on Machine Learning}.

\bibitem[\protect\astroncite{Dabkowski and Gal}{2017}]{Dabkowski17}
Dabkowski, P. and Gal, Y. (2017).
\newblock Real time image saliency for black box classifiers.
\newblock In {\em Advances in Neural Information Processing Systems}.

\bibitem[\protect\astroncite{Doshi-Velez and Kim}{2017}]{interpretability}
Doshi-Velez, F. and Kim, B. (2017).
\newblock Towards a rigorous science of interpretable machine learning.
\newblock {\em arXiv:1702.08608}.

\bibitem[\protect\astroncite{Ester et~al.}{1996}]{dbscan}
Ester, M., Kriegel, H.-P., Sander, J., and Xu, X. (1996).
\newblock A density-based algorithm for discovering clusters a density-based
  algorithm for discovering clusters in large spatial databases with noise.
\newblock In {\em {ACM} {SIGKDD} International Conference on Knowledge
  Discovery and Data Mining}.

\bibitem[\protect\astroncite{Fong and Vedaldi}{2017}]{Fong17}
Fong, R.~C. and Vedaldi, A. (2017).
\newblock Interpretable explanations of black boxes by meaningful perturbation.
\newblock In {\em IEEE Conference on Computer Vision and Pattern Recognition}.

\bibitem[\protect\astroncite{Hochreiter and Schmidhuber}{1997}]{lstm}
Hochreiter, S. and Schmidhuber, J. (1997).
\newblock Long short-term memory.
\newblock {\em Neural computation}, 9(8):1735--1780.

\bibitem[\protect\astroncite{Kim et~al.}{2018}]{Kim18}
Kim, B., Wattenberg, M., Gilmer, J., Cai, C., Wexler, J., Viegas, F., and
  Sayres, R. (2018).
\newblock Interpretability beyond feature attribution: Quantitative testing
  with concept activation vectors {(TCAV)}.
\newblock In {\em International Conference on Machine Learning}.

\bibitem[\protect\astroncite{Kindermans et~al.}{2017}]{Kindermans17}
Kindermans, P.-J., Hooker, S., Adebayo, J., Alber, M., Sch{\"u}tt, K.~T.,
  D{\"a}hne, S., Erhan, D., and Kim, B. (2017).
\newblock The (un) reliability of saliency methods.
\newblock In {\em {NIPS W}orkshop: Interpreting, Explaining and Visualizing
  Deep Learning {-} Now what?}

\bibitem[\protect\astroncite{Kindermans et~al.}{2018}]{Kindermans18}
Kindermans, P.-J., Sch{\"u}tt, K.~T., Alber, M., M{\"u}ller, K.-R., Erhan, D.,
  Kim, B., and D{\"a}hne, S. (2018).
\newblock Learning how to explain neural networks: {PatternNet and
  PatternAttribution}.
\newblock In {\em International Conference on Learning Representations}.

\bibitem[\protect\astroncite{Kleinberg et~al.}{2018}]{Kleinberg18}
Kleinberg, J., Lakkaraju, H., Leskovec, J., Ludwig, J., and Mullainathan, S.
  (2018).
\newblock Human decisions and machine predictions.
\newblock {\em The Quarterly Journal of Economics}, 133(1):237--293.

\bibitem[\protect\astroncite{LeCun et~al.}{1998}]{mnist}
LeCun, Y., Bottou, L., Bengio, Y., and Haffner, P. (1998).
\newblock Gradient-based learning applied to document recognition.
\newblock {\em Proceedings of the IEEE}, 86(11):2278--2324.

\bibitem[\protect\astroncite{Lei et~al.}{2016}]{Lei16}
Lei, T., Barzilay, R., and Jaakkola, T. (2016).
\newblock Rationalizing neural predictions.
\newblock In {\em Empirical Methods in Natural Language Processing}.

\bibitem[\protect\astroncite{Li et~al.}{2017}]{erasure}
Li, J., Monroe, W., and Jurafsky, D. (2017).
\newblock Understanding neural networks through representation erasure.
\newblock {\em arXiv:1612.08220}.

\bibitem[\protect\astroncite{Lipton}{2016}]{Lipton16}
Lipton, Z.~C. (2016).
\newblock The mythos of model interpretability.
\newblock In {\em ICML Workshop on Human Interpretability of Machine Learning}.

\bibitem[\protect\astroncite{Mathelier et~al.}{2015}]{jaspar}
Mathelier, A., Fornes, O., Arenillas, D.~J., Chen, C.-y., Denay, G., Lee, J.,
  Shi, W., Shyr, C., Tan, G., Worsley-Hunt, R., et~al. (2015).
\newblock Jaspar 2016: a major expansion and update of the open-access database
  of transcription factor binding profiles.
\newblock {\em Nucleic acids research}, 44(D1):D110--D115.

\bibitem[\protect\astroncite{McAuley et~al.}{2012}]{mcauley2012learning}
McAuley, J., Leskovec, J., and Jurafsky, D. (2012).
\newblock Learning attitudes and attributes from multi-aspect reviews.
\newblock In {\em IEEE International Conference on Data Mining}, pages
  1020--1025.

\bibitem[\protect\astroncite{Murdoch et~al.}{2018}]{James18}
Murdoch, W.~J., Liu, P.~J., and Yu, B. (2018).
\newblock Beyond word importance: Contextual decomposition to extract
  interactions from {LSTM}s.
\newblock In {\em International Conference on Learning Representations}.

\bibitem[\protect\astroncite{Olah et~al.}{2017}]{olah17}
Olah, C., Mordvintsev, A., and Schubert, L. (2017).
\newblock Feature visualization.
\newblock {\em Distill}.

\bibitem[\protect\astroncite{Olah et~al.}{2018}]{olah18}
Olah, C., Satyanarayan, A., Johnson, I., Carter, S., Schubert, L., Ye, K., and
  Mordvintsev, A. (2018).
\newblock The building blocks of interpretability.
\newblock {\em Distill}.

\bibitem[\protect\astroncite{Ribeiro et~al.}{2016}]{lime}
Ribeiro, M.~T., Singh, S., and Guestrin, C. (2016).
\newblock "{W}hy should {I} trust you?": Explaining the predictions of any
  classifier.
\newblock In {\em {ACM} {SIGKDD} International Conference on Knowledge
  Discovery and Data Mining}, pages 1135--1144.

\bibitem[\protect\astroncite{Rizzo and Sz{\'e}kely}{2016}]{energydistance}
Rizzo, M.~L. and Sz{\'e}kely, G.~J. (2016).
\newblock Energy distance.
\newblock {\em Wiley Interdisciplinary Reviews: Computational Statistics},
  8(1):27--38.

\bibitem[\protect\astroncite{Rubin}{1976}]{rubin1976inference}
Rubin, D.~B. (1976).
\newblock Inference and missing data.
\newblock {\em Biometrika}, 63(3):581--592.

\bibitem[\protect\astroncite{Sha and Wang}{2017}]{Sha17}
Sha, Y. and Wang, M.~D. (2017).
\newblock Interpretable predictions of clinical outcomes with an
  attention-based recurrent neural network.
\newblock In {\em ACM International Conference on Bioinformatics, Computational
  Biology,and Health Informatics}.

\bibitem[\protect\astroncite{Shrikumar et~al.}{2017}]{deeplift}
Shrikumar, A., Greenside, P., and Kundaje, A. (2017).
\newblock Learning important features through propagating activation
  differences.
\newblock In {\em International Conference on Machine Learning}.

\bibitem[\protect\astroncite{Simonyan et~al.}{2014}]{Simonyan14}
Simonyan, K., Vedaldi, A., and Zisserman, A. (2014).
\newblock Deep inside convolutional networks: Visualising image classification
  models and saliency maps.
\newblock In {\em International Conference on Learning Representations}.

\bibitem[\protect\astroncite{Sirignano et~al.}{2018}]{Sirignano18}
Sirignano, J.~A., Sadhwani, A., and Giesecke, K. (2018).
\newblock Deep learning for mortgage risk.
\newblock {\em arXiv:1607.02470}.

\bibitem[\protect\astroncite{Smola et~al.}{2005}]{Smola05}
Smola, A.~J., Vishwanathan, S., and Hofmann, T. (2005).
\newblock Kernel methods for missing variables.
\newblock In {\em Artificial Intelligence and Statistics}.

\bibitem[\protect\astroncite{Springenberg et~al.}{2015}]{Springenberg15}
Springenberg, J.~T., Dosovitskiy, A., Brox, T., and Riedmiller, M. (2015).
\newblock Striving for simplicity: The all convolutional net.
\newblock In {\em International Conference on Learning Representations}.

\bibitem[\protect\astroncite{Sundararajan et~al.}{2017}]{Sundararajan17}
Sundararajan, M., Taly, A., and Yan, Q. (2017).
\newblock Axiomatic attribution for deep networks.
\newblock In {\em International Conference on Machine Learning}.

\bibitem[\protect\astroncite{Tramer et~al.}{2016}]{predictionapi}
Tramer, F., Zhang, F., Juels, A., Reiter, M.~K., and Ristenpart, T. (2016).
\newblock Stealing machine learning models via prediction {API}s.
\newblock In {\em USENIX Security Symposium}.

\bibitem[\protect\astroncite{Zeiler and Fergus}{2014}]{Zeiler14}
Zeiler, M.~D. and Fergus, R. (2014).
\newblock Visualizing and understanding convolutional networks.
\newblock In {\em European Conference on Computer Vision}.

\bibitem[\protect\astroncite{Zeng et~al.}{2016}]{zeng2016convolutional}
Zeng, H., Edwards, M.~D., Liu, G., and Gifford, D.~K. (2016).
\newblock Convolutional neural network architectures for predicting
  dna--protein binding.
\newblock {\em Bioinformatics}, 32(12):i121.

\end{thebibliography}


\begin{thebibliography}{}

\bibitem[\protect\astroncite{Carlini and
  Wagner}{2017a}]{carlini2017adversarial}
Carlini, N. and Wagner, D. (2017a).
\newblock Adversarial examples are not easily detected: Bypassing ten detection
  methods.
\newblock In {\em Proceedings of the 10th ACM Workshop on Artificial
  Intelligence and Security}.

\bibitem[\protect\astroncite{Carlini and Wagner}{2017b}]{carlini2017towardssi}
Carlini, N. and Wagner, D. (2017b).
\newblock Towards evaluating the robustness of neural networks.
\newblock In {\em IEEE Symposium on Security and Privacy}.

\bibitem[\protect\astroncite{Chollet et~al.}{2015}]{kerassi}
Chollet, F. et~al. (2015).
\newblock Keras.
\newblock \url{https://keras.io}.

\bibitem[\protect\astroncite{Consortium et~al.}{2012}]{encode2012integratedsi}
Consortium, E.~P. et~al. (2012).
\newblock An integrated encyclopedia of dna elements in the human genome.
\newblock {\em Nature}, 489(7414):57.

\bibitem[\protect\astroncite{Kingma and Ba}{2015}]{adamsi}
Kingma, D.~P. and Ba, J. (2015).
\newblock Adam: A method for stochastic optimization.
\newblock In {\em International Conference on Learning Representations}.

\bibitem[\protect\astroncite{LeCun et~al.}{1998}]{mnistsi}
LeCun, Y., Bottou, L., Bengio, Y., and Haffner, P. (1998).
\newblock Gradient-based learning applied to document recognition.
\newblock {\em Proceedings of the IEEE}, 86(11):2278--2324.

\bibitem[\protect\astroncite{Lei et~al.}{2016}]{Lei16si}
Lei, T., Barzilay, R., and Jaakkola, T. (2016).
\newblock Rationalizing neural predictions.
\newblock In {\em Empirical Methods in Natural Language Processing}.

\bibitem[\protect\astroncite{McAuley et~al.}{2012}]{mcauley2012learningsi}
McAuley, J., Leskovec, J., and Jurafsky, D. (2012).
\newblock Learning attitudes and attributes from multi-aspect reviews.
\newblock In {\em IEEE International Conference on Data Mining}, pages
  1020--1025.

\bibitem[\protect\astroncite{Papernot et~al.}{2017}]{papernot2017cleverhans}
Papernot, N., Carlini, N., Goodfellow, I., Feinman, R., Faghri, F., Matyasko,
  A., Hambardzumyan, K., Juang, Y.-L., Kurakin, A., Sheatsley, R., Garg, A.,
  and Lin, Y.-C. (2017).
\newblock cleverhans v2.0.0: an adversarial machine learning library.
\newblock {\em arXiv:1610.00768}.

\bibitem[\protect\astroncite{Radford et~al.}{2017}]{radford2017learning}
Radford, A., Jozefowicz, R., and Sutskever, I. (2017).
\newblock Learning to generate reviews and discovering sentiment.
\newblock {\em arXiv:1704.01444}.

\bibitem[\protect\astroncite{Ribeiro et~al.}{2016}]{limesi}
Ribeiro, M.~T., Singh, S., and Guestrin, C. (2016).
\newblock "{W}hy should {I} trust you?": Explaining the predictions of any
  classifier.
\newblock In {\em {ACM} {SIGKDD} International Conference on Knowledge
  Discovery and Data Mining}.

\bibitem[\protect\astroncite{Rizzo and Sz{\'e}kely}{2016}]{energydistancesi}
Rizzo, M.~L. and Sz{\'e}kely, G.~J. (2016).
\newblock Energy distance.
\newblock {\em Wiley Interdisciplinary Reviews: Computational Statistics},
  8(1):27--38.

\bibitem[\protect\astroncite{Sundararajan et~al.}{2017}]{Sundararajan17si}
Sundararajan, M., Taly, A., and Yan, Q. (2017).
\newblock Axiomatic attribution for deep networks.
\newblock In {\em International Conference on Machine Learning}.

\bibitem[\protect\astroncite{Wang et~al.}{2016}]{wang2016attentionsi}
Wang, Y., Huang, M., Zhao, L., et~al. (2016).
\newblock Attention-based lstm for aspect-level sentiment classification.
\newblock In {\em Proceedings of the 2016 Conference on Empirical Methods in
  Natural Language Processing}.

\bibitem[\protect\astroncite{Zeiler}{2012}]{adadeltasi}
Zeiler, M.~D. (2012).
\newblock Adadelta: An adaptive learning rate method.
\newblock {\em arXiv:1212.5701}.

\bibitem[\protect\astroncite{Zeng et~al.}{2016}]{zeng2016convolutionalsi}
Zeng, H., Edwards, M.~D., Liu, G., and Gifford, D.~K. (2016).
\newblock Convolutional neural network architectures for predicting
  dna--protein binding.
\newblock {\em Bioinformatics}, 32(12):i121.

\end{thebibliography}
}

\clearpage
\newpage \beginsupplement
\setcounter{page}{1}
\thispagestyle{empty}
\onecolumn
\begin{center}
{\Large \textbf{Supplementary Information for: \ 
\papertitle 
}}
\end{center}
\vspace*{0mm}

\begingroup 
\let\orignumberline\numberline
\def\numberline#1{\orignumberline{#1}\kern1ex}
\renewcommand{\baselinestretch}{0.75}\normalsize
\setcounter{tocdepth}{0}
\begin{spacing}{0.85}
{\small \tableofcontents}
\end{spacing}
\addtocontents{toc}{\setcounter{tocdepth}{2}}
\endgroup
\begin{spacing}{0.85}
{\small  \listoffigures }
{\small \listoftables }
\end{spacing}
\renewcommand{\baselinestretch}{1.0}\normalsize

\clearpage 
\rhead{\thepage}
\pagestyle{fancy}  
\pagenumbering{arabic}
\setcounter{page}{3}
\clearpage
\section{Detailed Description of Alternative Methods}
\label{sec:alternativedets}
In Section~\ref{sec:methods}, we describe a number of alternative methods for identifying rationales for comparison with our method.
We use methods based on integrated gradients \citepsi{Sundararajan17si}, LIME \citepsi{limesi}, and feature perturbation.
Note that integrated gradients is an attribution method which assigns a numerical score to each input feature.
LIME likewise assigns a weight to each feature using a local linear regression model for $f$ around $\mathbf{x}$.
In the perturbative approach, we compute the change in prediction when each feature is individually masked, as in Equation~\ref{eq:feature-importance} (of  Section~\ref{sec:beerdets-feature-importance}).
Each of these feature orderings $R$ is used to construct a rationale using the \textbf{FindSIS} procedure (Section~\ref{sec:methods}) for the ``Suff. IG,'' ``Suff. LIME,'' and ``Suff. Perturb.'' (\textit{sufficiency constrained}) methods.

Note that our text classification architecture (described in Section~\ref{sec:beerdets-model-training}) encodes discrete words as 100-dimensional continuous word embeddings.
The integrated gradients method returns attribution scores for each coordinate of each word embedding.
For each word embedding $x_i \in \mathbf{x}$ (where each $x_i \in \mathbb{R}^{100}$), we summarize the attributions along the corresponding embedding into a single score $y_i$ using the $L_1$ norm: $y_i = \sum_d |x_{id}|$ and compute the ordering $R$ by sorting the $y_i$ values.

We use an implementation of integrated gradients for Keras-based models from \url{https://github.com/hiranumn/IntegratedGradients}.
In the case of the beer review dataset (Section~\ref{sec:sentanalysis}), we use the mean embedding vector as a baseline for computing integrated gradients.
In the case of TF binding (Section~\ref{sec:tf-binding}), we use the $[0.25, 0.25, 0.25, 0.25]$ uniform mean vector as the baseline reference value.
As suggested in \citetsi{Sundararajan17si}, we verified that the prediction at the baseline and the integrated gradients sum to approximately the prediction of the input.

For LIME and our beer reviews dataset, we use the approach described in \citetsi{limesi} for textual data, where individual words are removed entirely from the input sequence.
In our TF binding dataset, LIME replaces bases with the unknown \verb|N| base (represented as the uniform-distribution $[0.25, 0.25, 0.25, 0.25]$).
We use the implementation of LIME at: \url{https://github.com/marcotcr/lime}.
The \verb|LimeTextExplainer| module is used with default parameters, except we set the maximal number of features used in the regression to be the full input length so we can order all input features.

Additionally, we explore methods in which we use the same ordering $R$ by these alternative methods but select the same number of input features in the rationale to be the median SIS length in the SIS-collection computed by our method on each example: the ``IG,'' ``LIME,'' and ``Perturb.'' (\textit{length constrained}) methods.
In the TF binding models, we use a baseline of zero vectors such that the integrated gradients result along the encoded sequence is also one-hot.
We compute the feature ordering based on the absolute value of the non-zero integrated gradient attributions.

In TF binding data (Section~\ref{sec:tf-binding}), we add an additional method, ``Top IG,'' in which we compute integrated gradients using an all-zeros baseline and order features by attribution magnitude (as in the length constrained IG method).
But, we select elements for the rationale by finding the minimum number of elements necessary such that the sum of integrated gradients of those features equals $\tau - f(\boldsymbol{0})$, where $\boldsymbol{0}$ is the all-zeros baseline for integrated gradients.
Note that for the length constrained and Top IG methods, there is no guarantee of sufficiency $f(\mathbf{x}_S) \ge \tau$ for any input subset $S$.

\section{Details of the Transcription Factor Binding Analysis}
\label{sec:tfdets}

\subsection{Dataset and Model}
\label{sec:tfdets-dataset-model}

We use the \textit{motif occupancy} datasets\footnote{available at \url{http://cnn.csail.mit.edu}} from \citetsi{zeng2016convolutionalsi}, where each dataset originates from a ChIP-seq experiment from the ENCODE project \citepsi{encode2012integratedsi}.
Each of the 422 datasets studies a particular transcription factor, containing between 600 and 700,000 (median 50,000) 101 base-pair DNA sequences (inputs)  each associated with a  binary label  based on whether the sequence is bound by the TF or not.
Each dataset also contains a test set ranging between 150 and 170,000 sequences (median 12,000).
Here, the positive and negative classes in each dataset are balanced, and we filter out all sequences containing the unknown base (\verb|N|).
The nucleotide occurring at base position (\verb|A|, \verb|C|, \verb|G|, \verb|T|) is encoded as a one-hot representation which is fed into the CNN.  \citetsi{zeng2016convolutionalsi} showed that convolutional neural network architectures outperform other models for this TF binding prediction task.

For each of the 422 prediction tasks, we employ the best-performing ``1layer\_128motif'' architecture from \citetsi{zeng2016convolutionalsi}, defined as follows:
\begin{enumerate}
	\item \textbf{Input}: (101 x 4) sequence encoding
    \item \textbf{Convolutional Layer 1}: Applies 128 kernels of window size 24, with ReLU activation
    \item \textbf{Global Max Pooling Layer 1}: Performs global max pooling
    \item \textbf{Dense Layer 1}: 32 neurons, with ReLU activation and dropout probability 0.5
    \item \textbf{Dense Layer 2}: 1 neuron (output probability), with sigmoid activation
\end{enumerate}
We hold out 1/8 of each train set for validation and minimize binary cross-entropy using the Adadelta optimizer \citepsi{adadeltasi} with default parameter settings in Keras \citepsi{kerassi}.
We train each model on each of the 422 datasets for 10 epochs (using batch size 128) with early-stopping based on validation loss.
Figure~\ref{fig:tf-binding-auc} shows the area under the receiver operating curve (AUC) over the 422 datasets, and we note that the performance of our models closely resembles that in~\citetsi{zeng2016convolutionalsi}.

\subsection{Rationale length comparison between SIS and other methods}
\label{sec:tfdets-sis}

\begin{figure}
    \centering
    \begin{minipage}[t]{0.49\textwidth}
    \centering
    \includegraphics[width=1.0\textwidth]{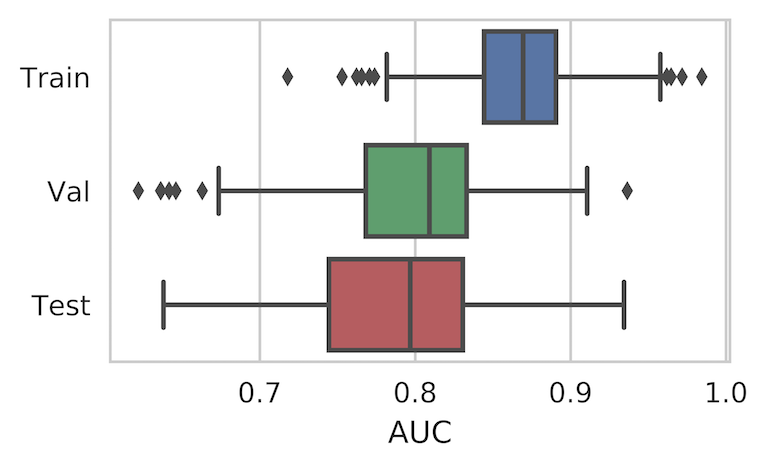}
    \caption[AUC performance of CNN models on TF binding prediction]{Median area under the receiver operating curve (AUC) for all 422 transcription factor binding motif occupancy datasets. The validation set is held-out at training but used to choose model parameters; the test set is not seen until after training.}
    \label{fig:tf-binding-auc}
    \end{minipage}\hfill
    \begin{minipage}[t]{0.49\textwidth}
        \centering
        \includegraphics[width=1.0\textwidth]{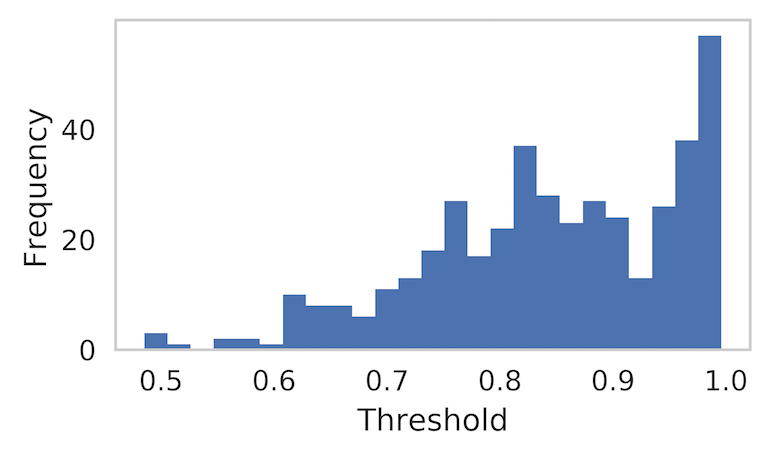}
        \caption[TF dataset sufficiency thresholds]{Thresholds $\tau$ used for identifying sufficient input subsets in TF binding datasets. In each dataset, the threshold is defined as the 90th percentile of the predictive test distribution.}
        \label{fig:tf-binding-thresholds} 
    \end{minipage}
\end{figure}

For each dataset, we define the sufficiency threshold $\tau$ as the 90th percentile of the predictive distribution on all test sequences.
The distribution of thresholds is shown in Figure~\ref{fig:tf-binding-thresholds}.
We compute the complete set of sufficient input subsets for each corresponding test sequence.
Since A,C,G,T nucleotides all occur with similar frequency in this data, our SIS analysis simply masks each base using a uniform embedding ($[0.25, 0.25, 0.25, 0.25]$).  This is also the standard strategy to represent unknown ``N'' nucleotides in DNA sequences that typically arise from issues in read quality.
We generally find that there is only a single SIS per example for the sequences in these datasets.

\begin{figure}
    \centering
    \begin{minipage}[t]{0.49\textwidth}
        \centering
        \includegraphics[width=1.0\textwidth]{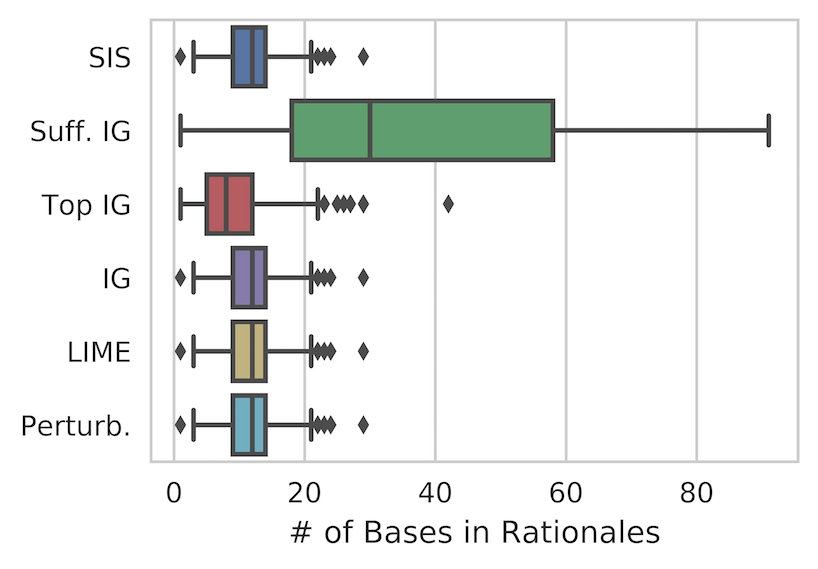}
        \caption[Length of  rationales in TF binding prediction]{Length (number of bases) of rationales identified by various methods. Note that the sufficiency constraint ($f(\mathbf{x}_S) \ge \tau$) is only enforced for SIS and Suff. IG. The lengths of IG, LIME, and Perturb. rationales are constrained to the length of SIS rationales.}
        \label{fig:tf-binding-rationale-lengths}
    \end{minipage}\hfill
    \begin{minipage}[t]{0.49\textwidth}
        \centering
        \includegraphics[width=1.0\textwidth]{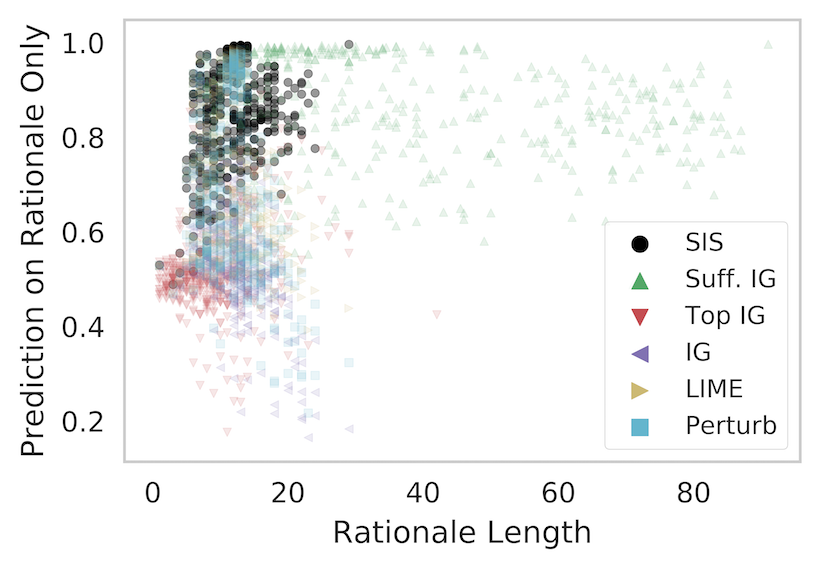}
        \caption[TF task rationale only prediction comparison]{Prediction on rationale only (all other bases masked) vs. rationale length (number of bases) for various methods in the TF binding task.}
        \label{fig:tf-binding-prediction-vs-length}
    \end{minipage}
\end{figure}

On each dataset, we compute the median rationale length (as number of bases in the rationale).
The distribution of median rationale length over all datasets by various methods is shown in Figure~\ref{fig:tf-binding-rationale-lengths}.
Note that for the IG, LIME, and Perturb. methods, rationale length was constrained to the length of the rationales produced by our method.
For the Top IG method, neither sufficiency or length constraints are enforced.
We see that when the sufficiency constraint is enforced in alternative methods (Suff. IG), the rationales are significantly longer than those identified by SIS.
Moreover, as shown in Figure~\ref{fig:tf-binding-prediction-vs-length}, when the sufficiency constraint is not enforced (or the rationale lengths are constrained to the length of SIS rationales) in alternative methods, the rationales have significantly less predictive power, often not satisfying $f(\mathbf{x}_S) \ge \tau$.

\subsection{Evaluation of the quality of TF Rationales}
\label{sec:tf-rationale-analysis}

Each rationale is padded with ``\verb|N|'' (unknown) bases to the length of a full input sequence (101 bases) and optimally aligned with the known motif\footnote{A JASPAR motif is a $n\times 4$ right stochastic matrix $M$. The columns represent the ACGT DNA bases and the rows a DNA sequence. It represents the marginal probability of the base $j$ at position $i$ being present with probability $M_{ij}$. The unknown base ``N'' receives uniform $1/4$ probability for each of ACGT.} 
according to the likelihood criterion.
The aligned motif is then also padded to the same length, and we compute the divergence between between the rationale $R$ and known motif $M$ as:
\begin{equation*}
    \mathrm{Div}(R, M) = \sum_{i} D_{\mathrm{KL}}(R_i || M_i)
\end{equation*}
where $D_{\mathrm{KL}}(R_i || M_i) = \sum_{j} R_i(j) \log \frac{R_i(j)}{M_i(j)}$ is the Kullback-Leibler divergence from $M_i$ to $R_i$, and $M_i$ and $R_i$ are distributions over bases (\verb|A|, \verb|C|, \verb|G|, \verb|T|) at position $i$.
Note that as $R$ and $M$ become more dissimilar, $\mathrm{Div}(R, M)$ increases.
We ensure $M_{ij} > 0 \,\, \forall \,\, i, j$ so $D_{\mathrm{KL}}$ is always finite.

\clearpage
\section{Details of the MNIST Analysis}
\label{sec:mnistdets}

\subsection{Dataset and Model} 
\label{sec:mnist-model-dets}
The MNIST database of handwritten digits contains 60k training images and 10k test images \citepsi{mnistsi}.
All images are 28x28 grayscale, and we normalize them such that all pixel values are between 0 and 1.
We use the convolutional architecture provided in the Keras MNIST CNN example.\footnote{\url{http://github.com/keras-team/keras/blob/master/examples/mnist_cnn.py}}
The architecture is as follows:
\begin{enumerate}
	\item \textbf{Input}: (28 x 28 x 1) image, all values $\in [0, 1]$
	\item \textbf{Convolutional Layer 1}: Applies 32 3x3 filters with ReLU activation
    \item \textbf{Convolutional Layer 2}: Applies 64 3x3 filters, with ReLU activation
    \item \textbf{Pooling Layer 1}: Performs max pooling with a 2x2 filter and dropout probability 0.25
    \item \textbf{Dense Layer 1}: 128 neurons, with ReLU activation and dropout probability 0.5
    \item \textbf{Dense Layer 2}: 10 neurons (one per digit class), with softmax activation
\end{enumerate}

The Adadelta optimizer ~\citepsi{adadeltasi} is used to minimize cross-entropy loss on the training set.
The final model achieves 99.7\% accuracy on the train set and 99.1\% accuracy on the held-out test set.

\subsection{Local Minima in Backward Selection}

\begin{figure}[h!]
  \centering
  \begin{subfigure}[t]{.5\textwidth}
    \centering
    \subcaption{}
    \vspace*{-0.5mm}
    \includegraphics[width=\textwidth]{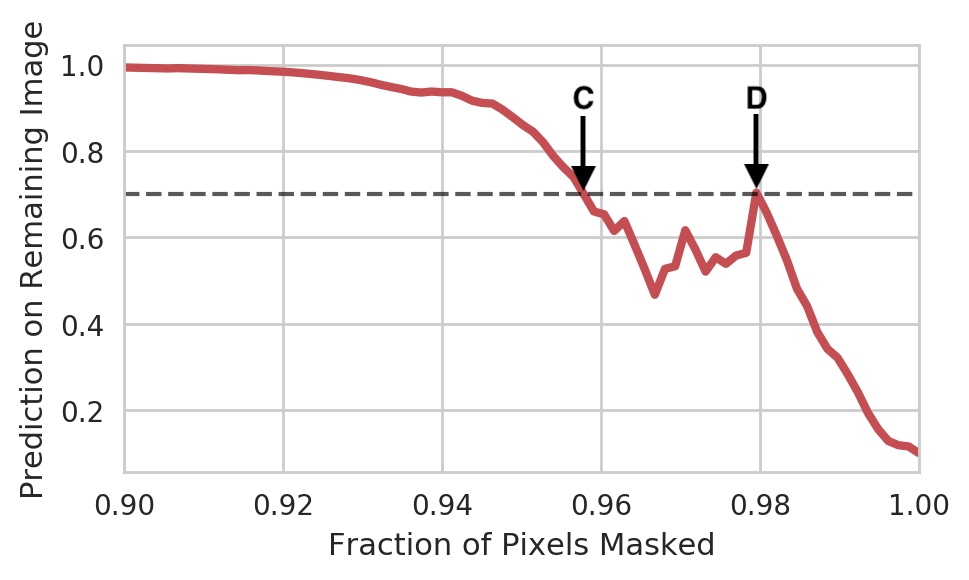}
      \label{fig:mnist-local-minimum-history}
  \end{subfigure}%
  \begin{subfigure}[t]{.16\textwidth}
    \centering
    \subcaption{}
    \vspace*{3mm}
    \includegraphics[width=0.98\textwidth]{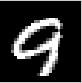}
      \label{fig:mnist-local-minimum-original}
  \end{subfigure}%
  \begin{subfigure}[t]{.16\textwidth}
    \centering
    \subcaption{}
    \vspace*{3mm}
    \includegraphics[width=0.98\textwidth]{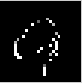}
      \label{fig:mnist-local-minimum-early-sis}
  \end{subfigure}%
  \begin{subfigure}[t]{.16\textwidth}
    \centering
    \subcaption{}
    \vspace*{3mm}
    \includegraphics[width=0.98\textwidth]{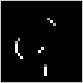}
      \label{fig:mnist-local-minimum-final-sis}
  \end{subfigure}%
  \vspace*{-1.5em}
  \caption[Local Minimum in Backward Selection]{\textbf{(a)} Prediction on remaining image as pixels are masked during backward selection, when our CNN classifier is fed the MNIST digit in (b). The dashed line depicts the threshold $\tau = 0.7$.
  \textbf{(b)} Original image (class 9).
  \textbf{(c)} SIS if backward selection were to terminate the first time prediction on remaining image drops below 0.7, corresponding to point \textbf{C} in (a) (CNN predicts class 9 with probability 0.700 on this SIS).
  \textbf{(d)} Actual SIS produced by our \textbf{FindSIS} algorithm, corresponding to point \textbf{D} in (a) (CNN predicts class 9 with probability 0.704 on this SIS).
  }
  \label{fig:mnist-local-minimum}
\end{figure}

Figure~\ref{fig:mnist-local-minimum} demonstrates an example MNIST digit for which there exists a local minimum in the backward selection phase of our algorithm to identify the initial SIS.  Note that if we were to terminate the backward selection as soon as predictions drop below the decision threshold, the resulting SIS would be overly large, violating our minimality criterion. It is also evident from Figure~\ref{fig:mnist-local-minimum} that the smaller-cardinality SIS in (d), found after the initial local optimum in (c), presents a more interpretable input pattern that enables better understanding of the core motifs influencing our classifier's decisions. 
To avoid suboptimal results, it is important to run a complete backward selection sweep until the entire input is masked before building the SIS upward, as done in our \textbf{SIScollection} procedure.

\subsection{Energy Distance Between Image SIS} 
\label{sec:energy}

To cluster SIS from the image data, we compute the pairwise distance between two SIS subsets $S_1$ and $S_2$ as the energy distance \citepsi{energydistancesi} between two distributions over the image pixel coordinates that comprise the SIS, $X_1$ and $X_2$ $\in \mathbb{R}^2$:  
\begin{equation*} 
    D(X_1, X_2) = 2 \cdot \E ||X_1 - X_2|| - \E ||X_1 - X_1'|| - \E ||X_2 - X_2'|| \geq 0
\end{equation*}

Here, $X_i$ is uniformly distributed over the pixels that are selected as part of the SIS subset $S_i$, $X_i'$ is an i.i.d. copy of $X_i$, and $|| \cdot ||$ represents the Euclidean norm.  
Unlike a Euclidean distance between images, our usage of the energy distance takes into account distances between the similar pixel coordinates that comprise each SIS.  The energy distance offers a more efficiently computable integral probability metric than the optimal transport distance, which has been widely adopted as an appropriate measure of distance between images.

\subsection{SIS Clustering and Adversarial Analysis}
\begin{figure}
    \centering
    \begin{minipage}[t]{0.48\textwidth}
        \centering
        \includegraphics[width=1.0\textwidth]{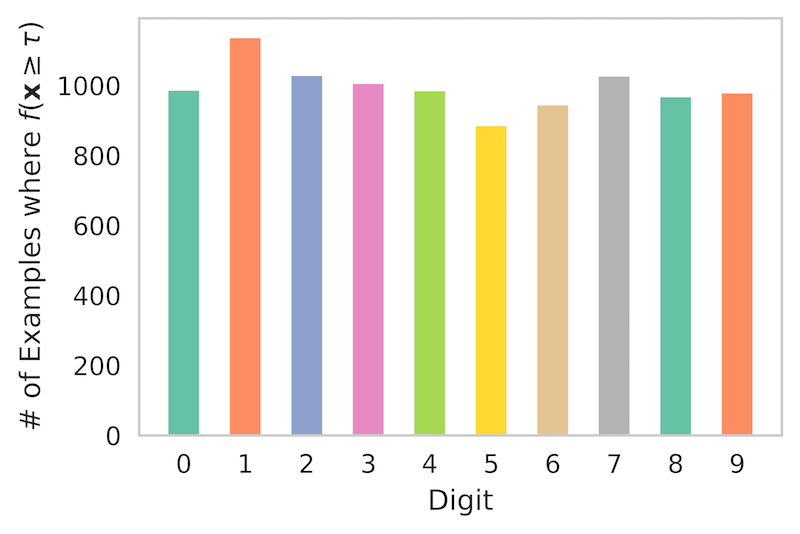}
        \caption[Number of examples per MNIST digit]{Number of examples per digit in the test set for which $f(\mathbf{x}) \ge \tau$ for the top class. The complete set of sufficient input subsets is computed for all of these examples.}
        \label{fig:mnist-num-past-threshold}
    \end{minipage}
    \hfill
    \begin{minipage}[t]{0.48\textwidth}
        \centering
        \includegraphics[width=1.0\textwidth]{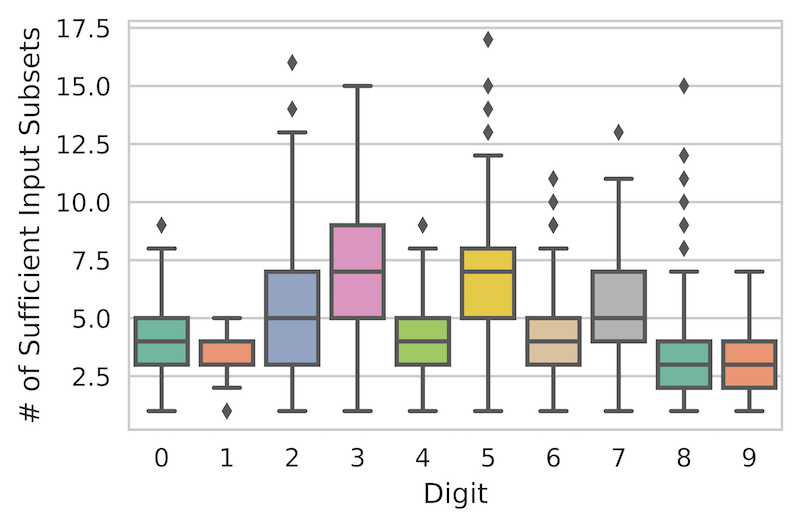}
        \caption[Number of SIS per MNIST digit]{Distributions of number of sufficient input subsets identified per image, by digit.}
        \label{fig:mnist-num-sis}
    \end{minipage}   
\end{figure}

We set the threshold $\tau = 0.7$ for SIS to ensure that the model is confident in its class prediction (probability of the predicted class is $\geq$ 0.7).
Almost all test examples initially have $f(\mathbf{x}) \ge \tau$ for the top class (Figure~\ref{fig:mnist-num-past-threshold}).
We identify all test examples that satisfy this condition and use SIS to identify all sufficient input subsets.
The number of sufficient input subsets per digit is shown in Figure~\ref{fig:mnist-num-sis}.

\begin{figure}
  \centering
  \begin{subfigure}[b]{.49\linewidth}
    \includegraphics[width=\linewidth, height=3.62cm,keepaspectratio]{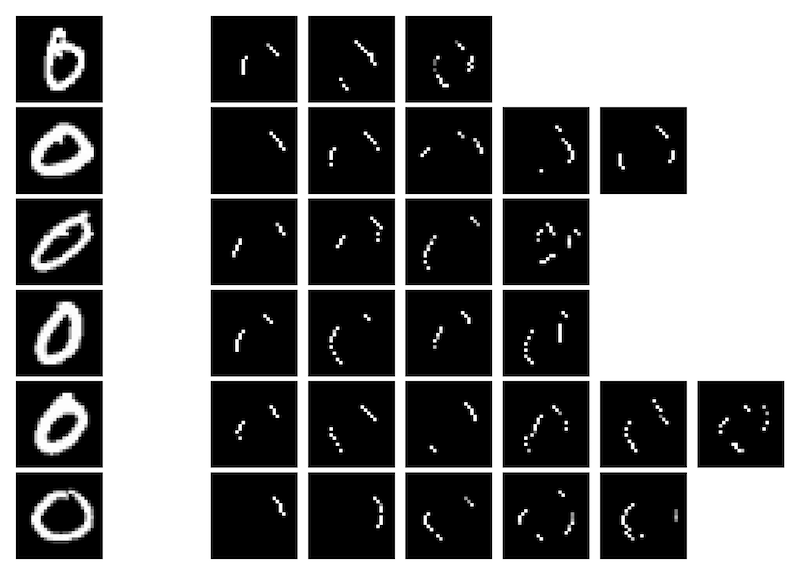}
    \caption{Digit 0}\label{fig:mnist-sis-examples-0}
  \end{subfigure}
  \begin{subfigure}[b]{.49\linewidth}
     \includegraphics[width=\linewidth,height=3.62cm,keepaspectratio]{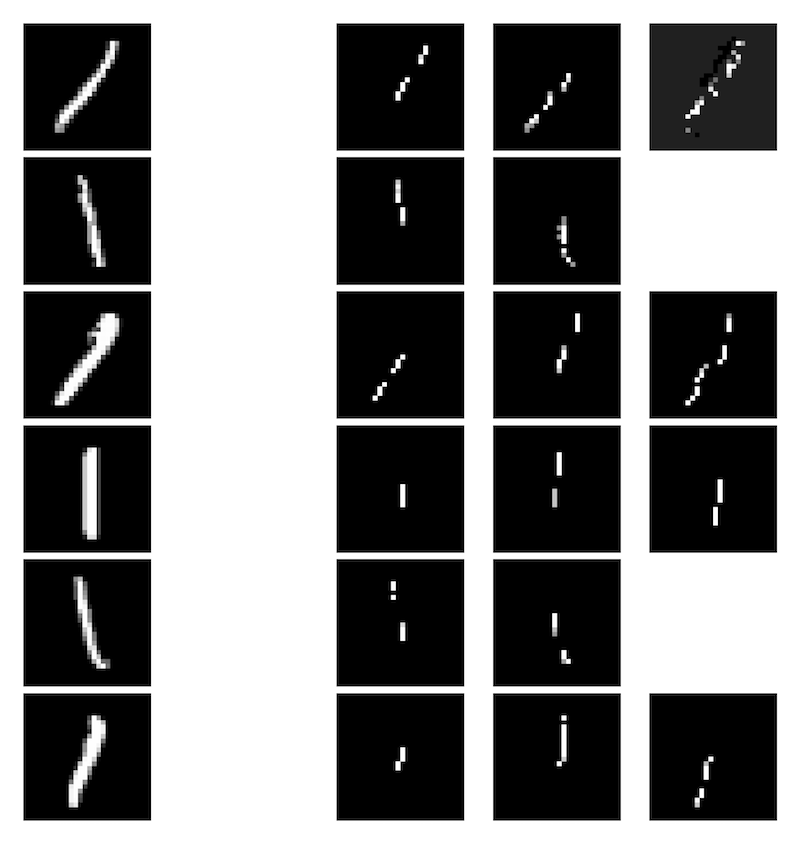}
    \caption{Digit 1}\label{fig:mnist-sis-examples-1}
  \end{subfigure}
  
  \begin{subfigure}[b]{.49\linewidth}
    \includegraphics[width=\linewidth,height=3.62cm,keepaspectratio]{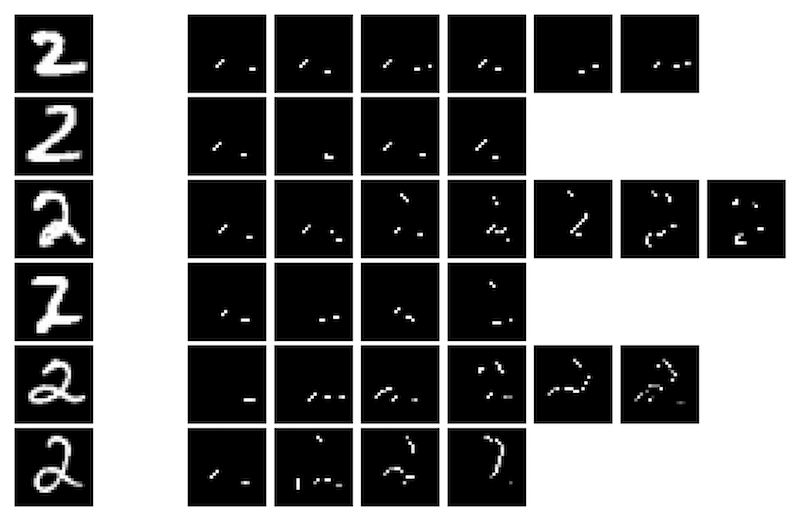}
    \caption{Digit 2}\label{fig:mnist-sis-examples-2}
  \end{subfigure}
  \begin{subfigure}[b]{.49\linewidth}
     \includegraphics[width=\linewidth,height=3.62cm,keepaspectratio]{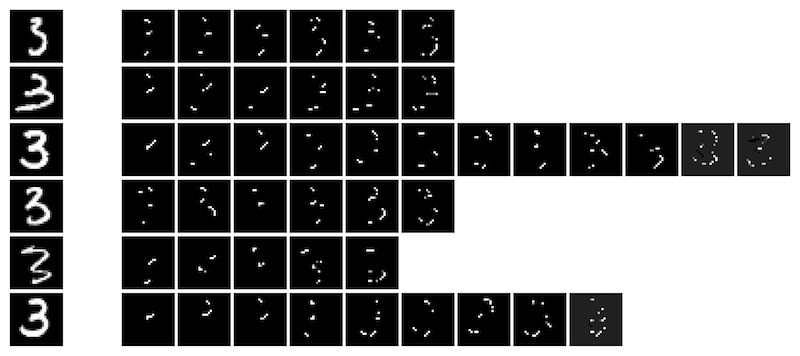}
    \caption{Digit 3}\label{fig:mnist-sis-examples-3}
  \end{subfigure}
  
  \begin{subfigure}[b]{.49\linewidth}
    \includegraphics[width=\linewidth,height=3.62cm,keepaspectratio]{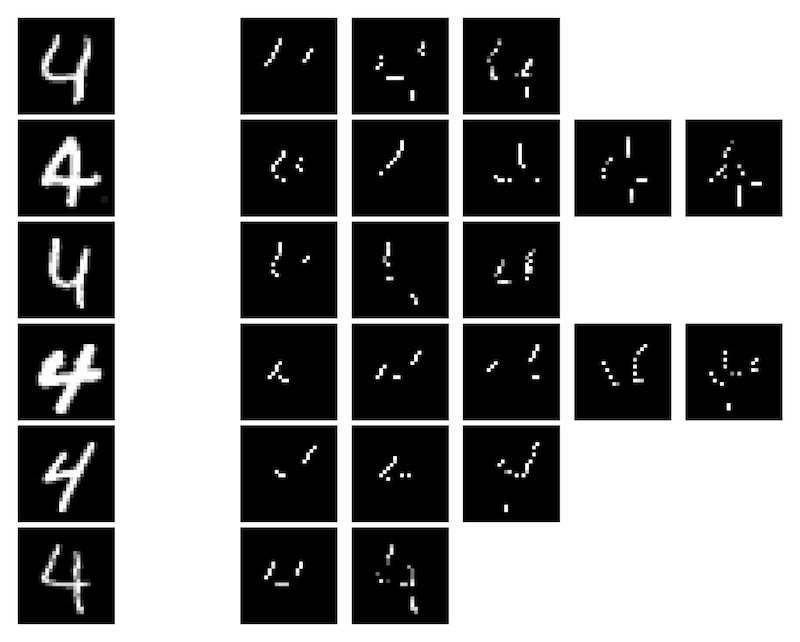}
    \caption{Digit 4}\label{fig:mnist-sis-examples-4}
  \end{subfigure}
  \begin{subfigure}[b]{.49\linewidth}
     \includegraphics[width=\linewidth,height=3.62cm,keepaspectratio]{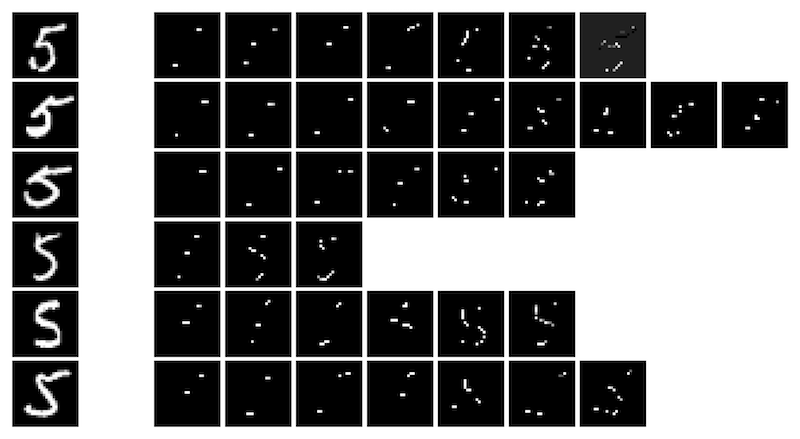}
    \caption{Digit 5}\label{fig:mnist-sis-examples-5}
  \end{subfigure}
  
  \begin{subfigure}[b]{.49\linewidth}
    \includegraphics[width=\linewidth,height=3.62cm,keepaspectratio]{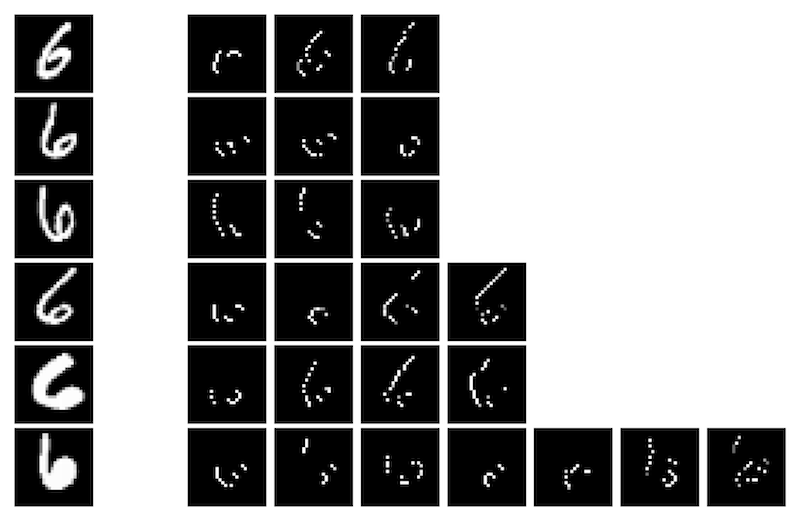}
    \caption{Digit 6}\label{fig:mnist-sis-examples-6}
  \end{subfigure}
  \begin{subfigure}[b]{.49\linewidth}
      \includegraphics[width=\linewidth,height=3.62cm,keepaspectratio]{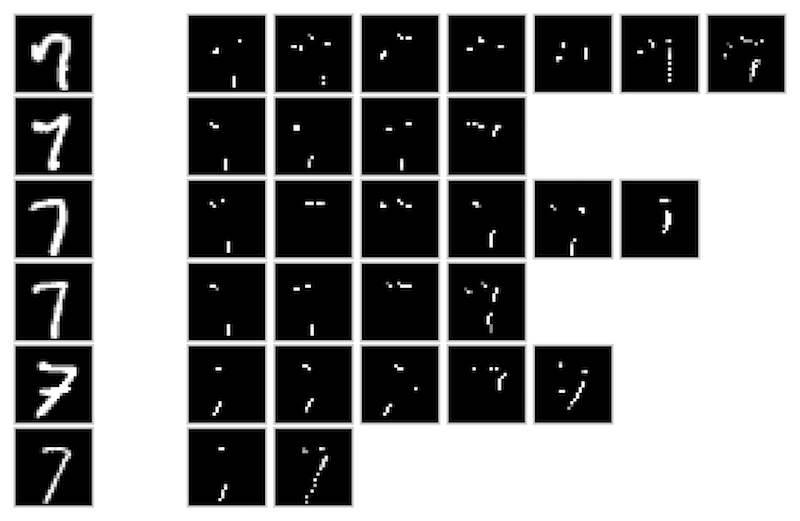}
    \caption{Digit 7}\label{fig:mnist-sis-examples-7}
  \end{subfigure}
  
  \begin{subfigure}[b]{.49\linewidth}
    \includegraphics[width=\linewidth,height=3.62cm,keepaspectratio]{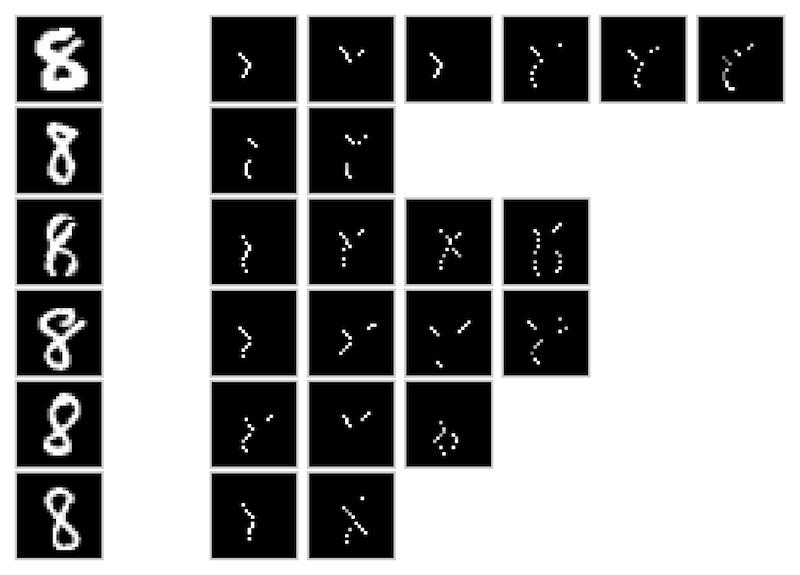}
    \caption{Digit 8}\label{fig:mnist-sis-examples-8}
  \end{subfigure}
  \begin{subfigure}[b]{.49\linewidth}
      \includegraphics[width=\linewidth,height=3.62cm,keepaspectratio]{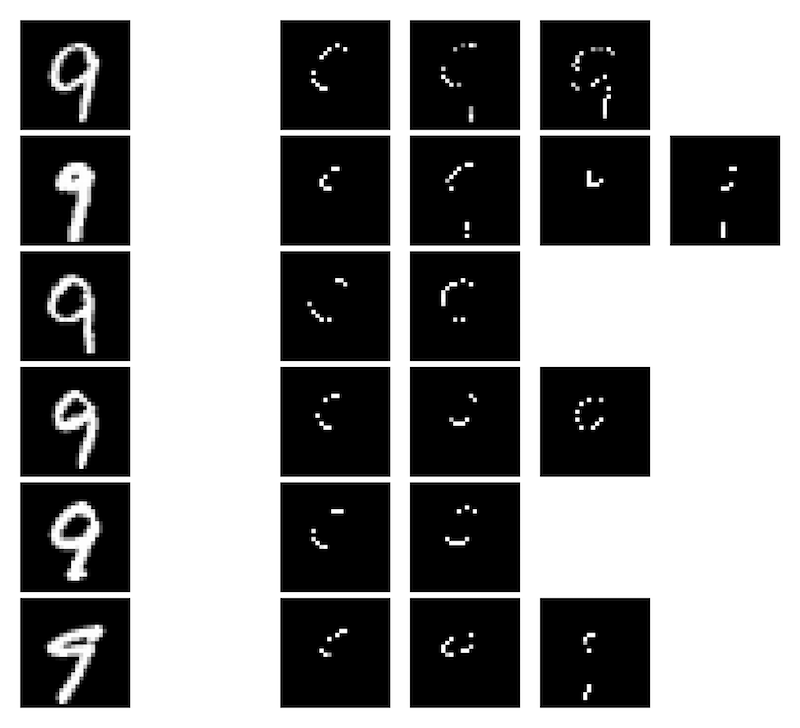}
    \caption{Digit 9}\label{fig:mnist-sis-examples-9}
   \end{subfigure}

  \caption[SIS examples on MNIST (CNN)]{Visualization of SIS-collections identified from MNIST digits that are confidently classified by the CNN. For each class, six examples were chosen randomly. For each example, we show the original image (left) and the complete set of sufficient input subsets identified for that example (remaining images in each row). Each individual SIS satisfies $f(\mathbf{x}_S) \ge \tau$ for that class.}
  \label{fig:mnist-sis-examples-all-digits}
\end{figure}

\begin{figure}
  \centering
  \begin{subfigure}[b]{.40\linewidth}
  \includegraphics[width=\linewidth]{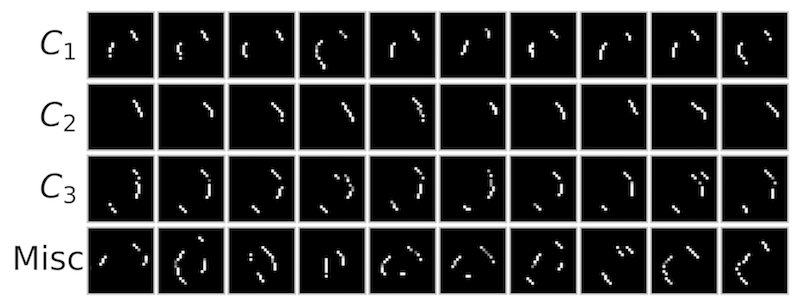}
  \caption{Digit 0}\label{fig:mnist-cluster-0}
  \end{subfigure}
  \begin{subfigure}[b]{.40\linewidth}
  \includegraphics[width=\linewidth]{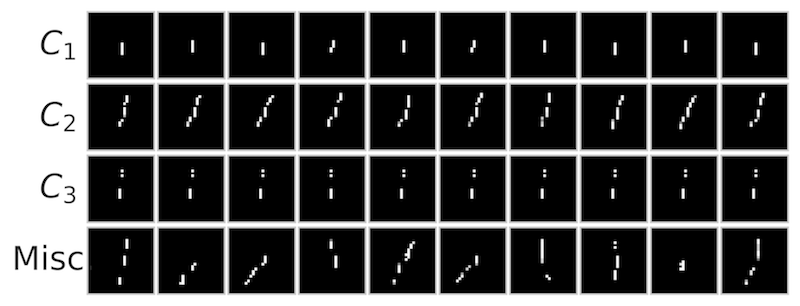}
  \caption{Digit 1}\label{fig:mnist-cluster-1}
  \end{subfigure}

  \begin{subfigure}[b]{.40\linewidth}
  \includegraphics[width=\linewidth]{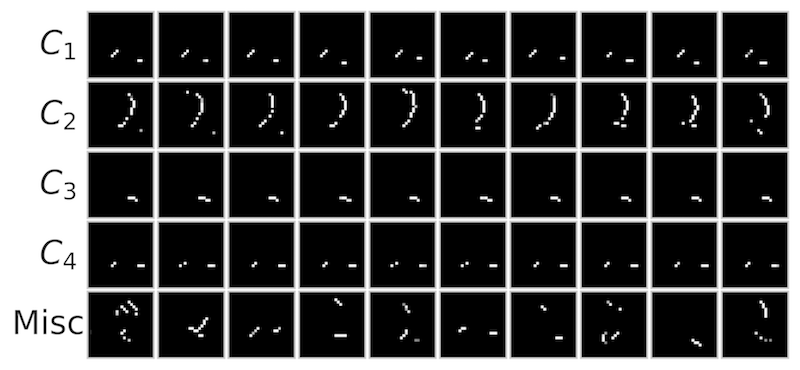}
  \caption{Digit 2}\label{fig:mnist-cluster-2}
  \end{subfigure}
  \begin{subfigure}[b]{.40\linewidth}
  \includegraphics[width=\linewidth]{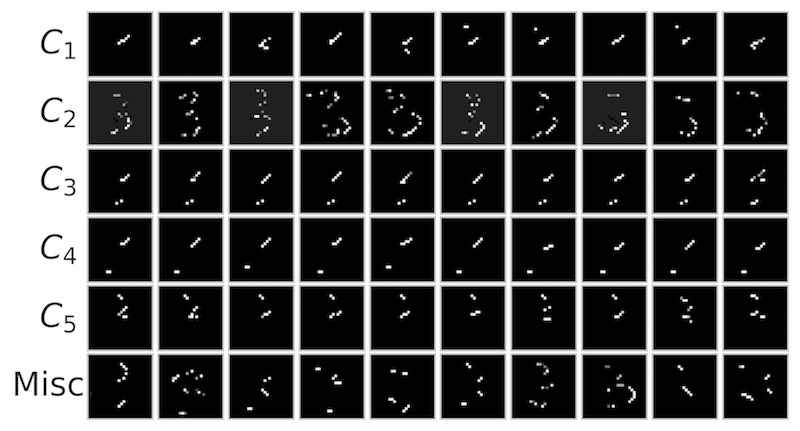}
  \caption{Digit 3}\label{fig:mnist-cluster-3}
  \end{subfigure}

  \begin{subfigure}[b]{.40\linewidth}
  \includegraphics[width=\linewidth]{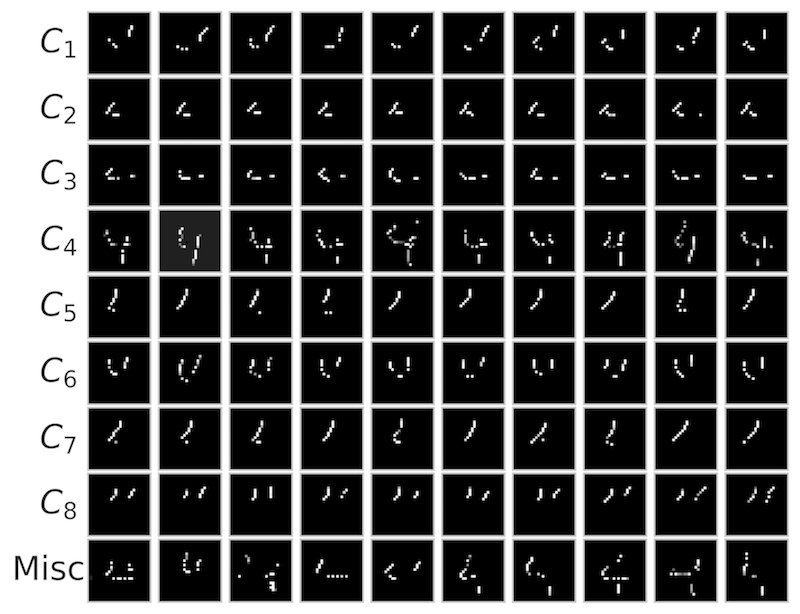}
  \caption{Digit 4}\label{fig:mnist-cluster-4}
  \end{subfigure}
  \begin{subfigure}[b]{.40\linewidth}
  \includegraphics[width=\linewidth]{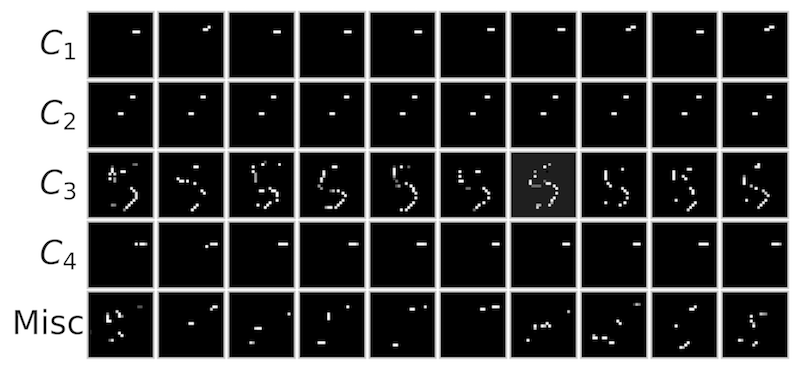}
  \caption{Digit 5}\label{fig:mnist-cluster-5}
  \end{subfigure}

  \begin{subfigure}[b]{.40\linewidth}
  \includegraphics[width=\linewidth]{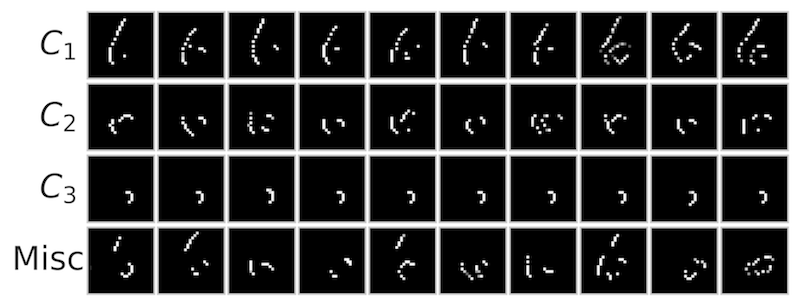}
  \caption{Digit 6}\label{fig:mnist-cluster-6}
  \end{subfigure}
  \begin{subfigure}[b]{.40\linewidth}
  \includegraphics[width=\linewidth]{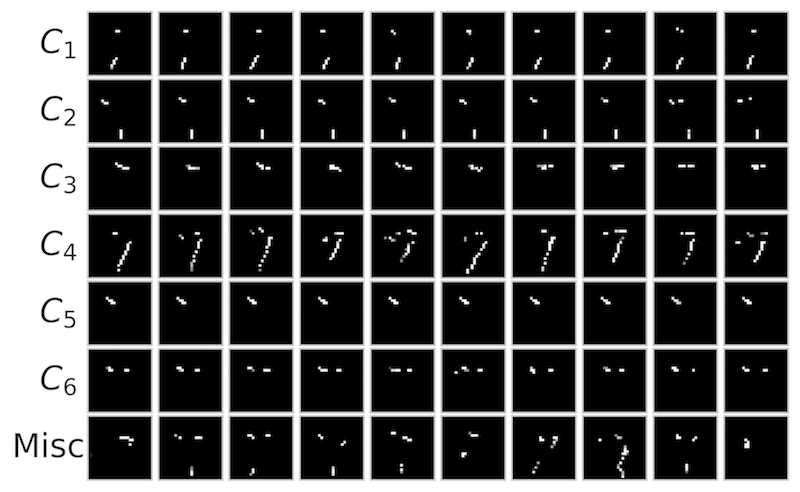}
  \caption{Digit 7}\label{fig:mnist-cluster-7}
  \end{subfigure}

  \begin{subfigure}[b]{.40\linewidth}
  \includegraphics[width=\linewidth]{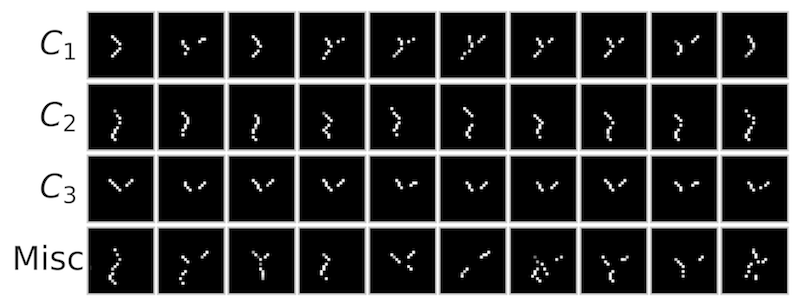}
  \caption{Digit 8}\label{fig:mnist-cluster-8}
  \end{subfigure}
  \begin{subfigure}[b]{.40\linewidth}
  \includegraphics[width=\linewidth]{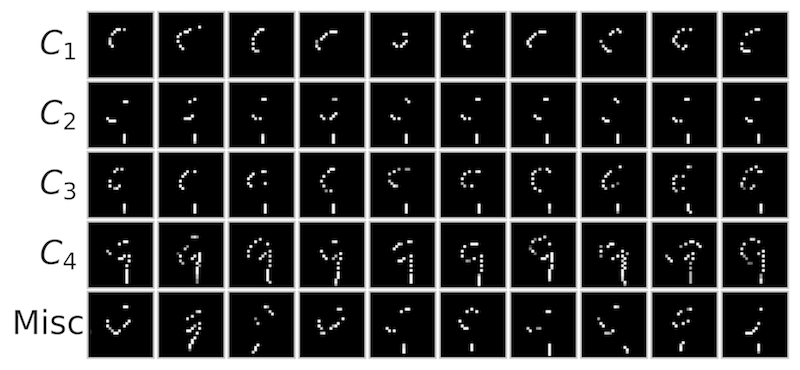}
  \caption{Digit 9}\label{fig:mnist-cluster-9}
  \end{subfigure}

  \caption[SIS clusters identified on MNIST (CNN)]{Clustering all the SIS  found for each digit under the CNN model (see Section~\ref{sec:clustering-insights}). Each row contains images drawn from one cluster. The bottom row (``Misc'') contains a sample of miscellaneous SIS not assigned to any cluster by DBSCAN.}
  \label{fig:mnist-all-clusters}
\end{figure}

We apply our \textbf{SIScollection} algorithm to identify sufficient input subsets on MNIST test digits (Section~\ref{sec:mnist}).
Examples of the complete SIS-collection corresponding to randomly chosen digits are shown in Figure~\ref{fig:mnist-sis-examples-all-digits}.
We also cluster all the sufficient input subsets identified for each class (Section~\ref{sec:clustering-insights}), depicting the results in Figure~\ref{fig:mnist-all-clusters}.

In Figure~\ref{fig:95sises}, we show an MNIST image of the digit 9, adversarially perturbed to 4, and the sufficient subsets corresponding to the adversarial prediction. Although a visual inspection of the perturbed image does not really reveal exactly how it has been manipulated, it becomes immediately clear from the SIS-collection for the adversarial image.  These sets shows that the perturbation modifies pixels in such a way that input patterns similar to the typical SIS-collection for a 4 (Figure \ref{fig:mnist-clustering-4-maintext}) become embedded in the image. 
The adversarial manipulation was done using the Carlini-Wagner $L_2$ (CW2) attack\footnote{Implemented in the cleverhans library of \citetsi{papernot2017cleverhans}} \citepsi{carlini2017towardssi} with a confidence parameter of 10. The CW2 attack tries to find the minimal change to the image, with respect to the $L_2$ norm, that will lead the image to be misclassified. \citetsi{carlini2017adversarial} demonstrate it to be one of the strongest extant adversarial attacks.  

\FloatBarrier
\subsection{Understanding Differences Between MNIST Classifiers}
\label{sec:mnist-understanding-differences-dets}
We use SIS and our clustering procedure to understand and visualize differences in features learned by two different models trained on the same MNIST digit classification task.
In addition to the previously-described CNN model (see Section~\ref{sec:mnist-model-dets}), we also trained a simple multilayer perceptron (MLP) on the same task.
The MLP architecture is as follows:
\begin{enumerate}
	\item \textbf{Input}: 784-dimensional (flattened) image, all values $\in [0, 1]$
    \item \textbf{Dense Layer 1}: 250 neurons, ReLU activation, and dropout probability 0.2
    \item \textbf{Dense Layer 2}: 250 neurons, ReLU activation, and dropout probability 0.2
    \item \textbf{Dense Layer 3}: 10 neurons (one per digit class), with softmax activation
\end{enumerate}
As with the CNN, Adadelta \citepsi{adadeltasi} is used to minimize cross-entropy loss on the training set.
The final MLP model achieves 99.7\% accuracy on the train set and 98.3\% accuracy on the test set, which is close to the performance of the CNN (see Section~\ref{sec:mnist-model-dets}).

We apply the same procedure as in Section~\ref{sec:mnist} to extract the SIS-collection from all applicable test images using the MLP.  
To understand differences between the feature patterns that each model has learned to associate with predicting each digit, we combine all SIS (from both models for a particular class) and run our clustering procedure (see Section~\ref{sec:clustering-insights} and Figure~\ref{fig:mnist-model-differences-clustering-digit4}).
In the resulting clustering, we list what percentage of the SIS in each cluster stem from the CNN vs.\ the MLP.  Most clusters contain examples purely from a single model, indicating the two models have learned to associate different feature patterns with the target class (Figure~\ref{fig:mnist-model-differences-clustering-digit4}), which was chosen to be the digit 4 in this case.

\begin{figure}
  \centering
  \begin{subfigure}[t]{.5\textwidth}
    \centering
    \includegraphics[width=1.0\textwidth]{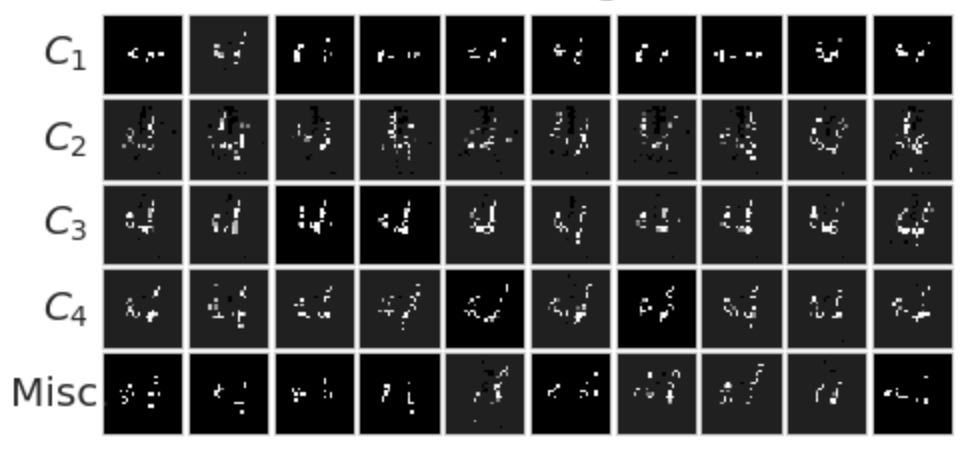}
    \subcaption{Digit 4}
    \label{fig:mnist-mlp-only-clustering-digit4}
  \end{subfigure}%
  \begin{subfigure}[t]{.5\textwidth}
    \centering
    \includegraphics[width=1.0\textwidth]{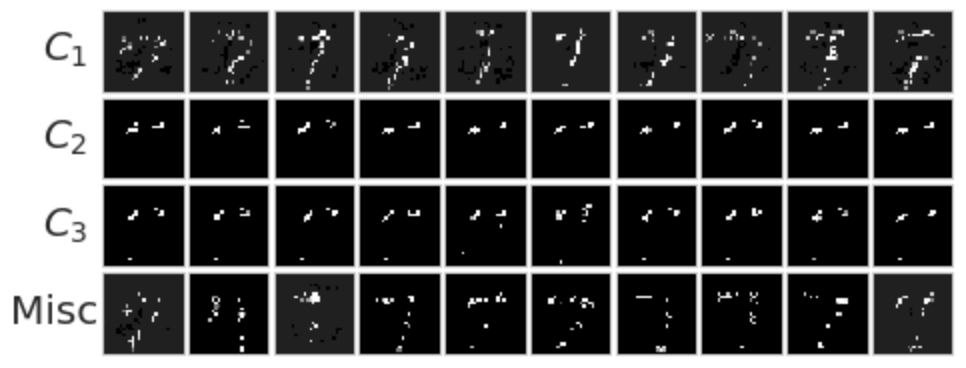}
    \subcaption{Digit 7}
    \label{fig:mnist-mlp-only-clustering-digit7}
  \end{subfigure}%
  \caption[SIS clusters identified on MNIST (MLP)]{Clustering all the SIS identified by our method on digits 4 and 7 under the MLP model (see Section~\ref{sec:clustering-insights}). Each row contains images drawn from one cluster.  The bottom row (``Misc'') contains a sample of miscellaneous SIS not assigned to any cluster by DBSCAN. Compare to the SIS-clustering from our CNN model (Figure~\ref{fig:mnist-all-clusters}).}
  \label{fig:mnist-mlp-only-clustering}
\end{figure}

\begin{figure}
  \centering
  \begin{subfigure}[t]{.5\textwidth}
    \centering
    \includegraphics[width=0.65\textwidth]{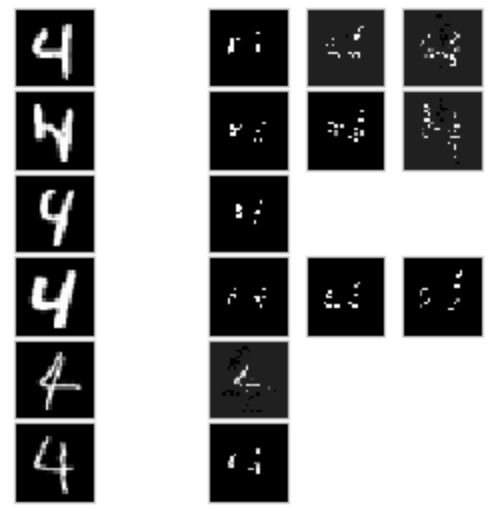}
    \subcaption{Digit 4}
    \label{fig:mnist-mlp-only-all-sis-digit4}
  \end{subfigure}%
  \begin{subfigure}[t]{.5\textwidth}
    \centering
    \includegraphics[width=0.65\textwidth]{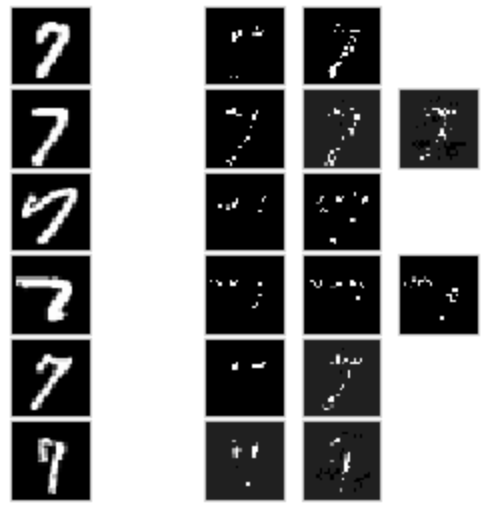}
    \subcaption{Digit 7}
    \label{fig:mnist-mlp-only-all-sis-digit7}
  \end{subfigure}%
  \caption[SIS examples on MNIST (MLP)]{Visualization of SIS-collections identified for MNIST digits 4 and 7 under the MLP model. For each class, six examples were chosen randomly. For each example, we show the original image (left) and the complete set of sufficient input subsets identified for that example (remaining images in each row). Note that each individual SIS satisfies $f(\mathbf{x}_S) \ge \tau$ for that class. Compare to the SIS extracted from our CNN (Figure~\ref{fig:mnist-sis-examples-all-digits}).}
  \label{fig:mnist-mlp-only-all-sis}
\end{figure}

For further comparison, we include clustering results for the SIS extracted from the MLP as evidence for digits 4 and 7 (Figure~\ref{fig:mnist-mlp-only-clustering}).  Additionally, Figure~\ref{fig:mnist-mlp-only-all-sis} shows all of the SIS extracted from example digits from these classes applying our procedure on the MLP.

\clearpage
\section{Details of the Beer Reviews Sentiment Analysis}
\label{sec:beerdets}

\subsection{Beer Reviews Data Description}
\label{sec:beerdets-dataset}
Following \citetsi{Lei16si}, we use a preprocessed version of the BeerAdvocate\footnote{\url{https://www.beeradvocate.com/}}
dataset\footnote{\url{http://snap.stanford.edu/data/web-BeerAdvocate.html}}  
which contains decorrelated numerical ratings toward three aspects: \textit{aroma}, \textit{appearance}, and \textit{palate} (each normalized to $[0, 1]$).
Dataset statistics can be found in Table~\ref{table:beerdets-dataset-stats}.
Reviews were tokenized by converting to lowercase and  filtering punctuation, and we used a vocabulary containing the top 10,000 most common words. 
\citetsi{mcauley2012learningsi} also provide a subset of human-annotated reviews, in which humans manually selected full sentences in each review that describe the relevant aspects.
This annotated set was never seen during training and used solely as part of our evaluation.

\subsection{Model Architecture and Training}
\label{sec:beerdets-model-training}
Long short-term memory (LSTM) networks are commonly employed for natural language tasks such as sentiment analysis \citepsi{wang2016attentionsi, radford2017learning}.
We use a recurrent neural network (RNN) architecture with two stacked LSTMs as follows:
\begin{enumerate}
	\item \textbf{Input/Embeddings Layer}: Sequence with 500 timesteps, the word at each timestep is represented by a (learned) 100-dimensional embedding
    \item \textbf{LSTM Layer 1}: 200-unit recurrent layer with LSTM (forward direction only)
    \item \textbf{LSTM Layer 2}: 200-unit recurrent layer with LSTM (forward direction only)
    \item \textbf{Dense}: 1 neuron (sentiment output), sigmoid activation
\end{enumerate}

With this architecture, we use the Adam optimizer \citepsi{adamsi} to minimize mean squared error (MSE) on the training set.
We use a held-out set of 3,000 examples for validation (sampled at random from the pre-defined test set from \citetsi{Lei16si}).
Our test set consists of the remaining 7,000 test examples.
Training results are shown in Table~\ref{table:beerdets-dataset-stats}.

\begin{table}[h!]
  \caption[Summary and performance statistics for beer reviews data]{Summary and performance statistics (mean squared error (MSE) and Pearson correlation coefficient $\rho$) for beer reviews data and LSTM models.}
  \label{table:beerdets-dataset-stats}
  \centering
  \begin{tabular}{lllll}
    \toprule
    Aspect & Fold & Size & MSE & Pearson $\rho$  \\
    \midrule
    \multirow{4}{*}{Appearance} & Train & 80,000 & 0.016 & 0.864  \\ 
    & Validation & 3,000 & 0.024 & 0.783  \\ 
    & Test & 7,000 & 0.023 & 0.801  \\ 
    & Annotation & 994 & 0.020 & 0.563  \\ 
    \midrule
    \multirow{4}{*}{Aroma} & Train & 70,000 & 0.014 & 0.873  \\ 
    & Validation & 3,000 & 0.024 & 0.767  \\ 
    & Test & 7,000 & 0.025 & 0.756  \\ 
    & Annotation & 994 & 0.021 & 0.598  \\ 
    \midrule
    \multirow{4}{*}{Palate} & Train & 70,000 & 0.016 & 0.835  \\ 
    & Validation & 3,000 & 0.029 & 0.680  \\ 
    & Test & 7,000 & 0.028 & 0.694  \\ 
    & Annotation & 994 & 0.016 & 0.592  \\ 
    \bottomrule
  \end{tabular}
\end{table}

\subsection{Imputation Strategies: Mean vs. Hot-deck}
\label{sec:beerdets-mean-embedding}
\begin{figure}[h!]
    \centering
    \includegraphics[width=0.65\textwidth]{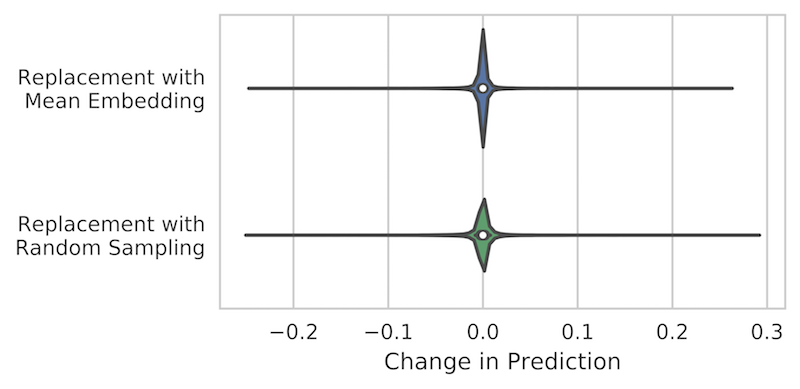}
    \caption[Mean vs. hot-deck imputation for aroma prediction]{Change in prediction ($f(\mathbf{x} \setminus \{i\}) - f(\mathbf{x})$) after masking a randomly chosen word with mean imputation or hot-deck imputation. 10,000 replacements were sampled from the aroma beer reviews training set.}
    \label{fig:beer-aroma-mean-embedding}
\end{figure}

In Section~\ref{sec:methods}, we discuss the problem of masking input features.
Here, we show that the mean-imputation approach (in which missing inputs are masked with a mean embedding, taken over the entire vocabulary) produces a nearly identical change in prediction to a nondeterministic hot-deck approach (in which missing inputs are replaced by randomly sampling feature-values from the data).
Figure~\ref{fig:beer-aroma-mean-embedding} shows the change in prediction $f(\mathbf{x} \setminus \{i\}) - f(\mathbf{x})$ by both imputation techniques after drawing a training example $\mathbf{x}$ and word $x_i \in \mathbf{x}$ (both uniformly at random) and replacing $x_i$ with either the mean embedding or a randomly selected word (drawn from the vocabulary, based on counts in the training corpus).
This procedure is repeated 10,000 times.
Both resulting distributions have mean near zero ($\mu_{\text{mean-embedding}} = -7.0\mathrm{e}{-4}$, $\mu_{\text{hot-deck}} = -7.4\mathrm{e}{-4}$), and the distribution for mean embedding is slightly narrower ($\sigma_{\text{mean-embedding}} = 0.013$, $\sigma_{\text{hot-deck}} = 0.018$).
We conclude that mean-imputation is a suitable method for masking information about particular feature values in our SIS analysis.

We also explored other options for masking word information, e.g. replacement with a zero embedding, replacement with the learned <PAD> embedding, and simply removing the word entirely from the input sequence, but each of these alternative  options led to undesirably larger changes in predicted values as a result of masking, indicating they appear more informative to $f$ than replacement via the feature-mean.

\subsection{Feature Importance Scores}
\label{sec:beerdets-feature-importance}

\begin{figure}
    \centering
    \begin{minipage}{0.48\textwidth}
        \centering
        \includegraphics[width=0.88\textwidth]{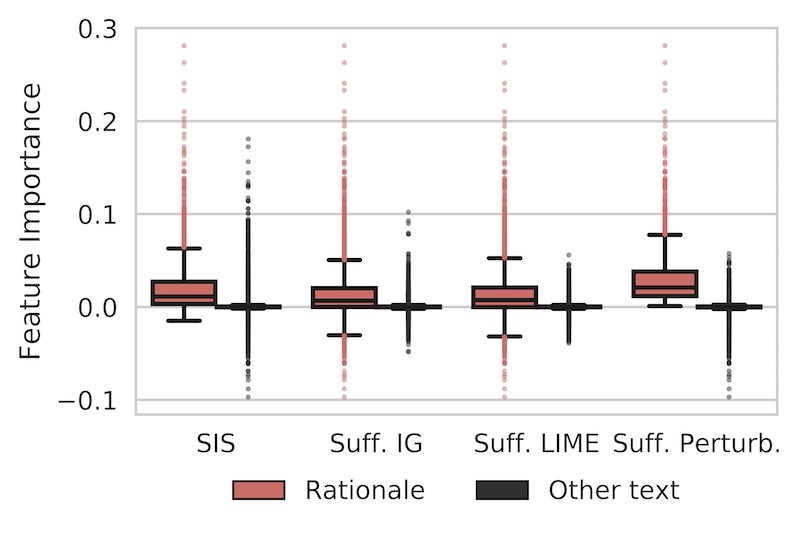}
        \caption[Feature importance comparison for aroma prediction]{Importance of individual features in the rationales for aroma prediction in beer reviews}
        \label{fig:beer-aroma-perturbation}
    \end{minipage}
    \hfill
    \begin{minipage}{0.48\textwidth}
        \centering
        \includegraphics[width=1.0\textwidth]{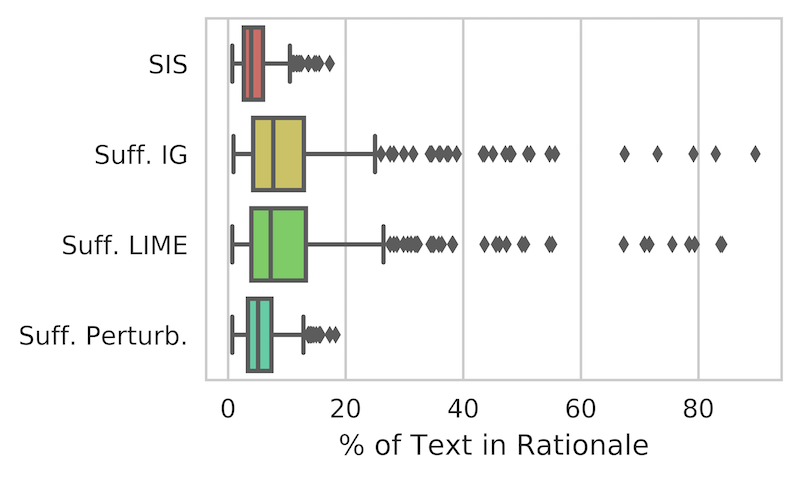}
        \caption[Length of rationales for aroma prediction]{Length of rationales for aroma prediction}
        \label{fig:beer-aroma-rationale-lengths}
    \end{minipage}
\end{figure}

For each feature $i$ in the input sequence, we quantify its marginal importance by individually perturbing only this feature:
\begin{equation}
    \label{eq:feature-importance}
    \text{Feature Importance}(i) = \text{prediction on original input} - \text{prediction with  feature } i \text{ masked}
\end{equation}
Note that these marginal Feature Importance scores are identical to those of the Perturb.\ method described in Section~\ref{sec:alternativedets}.
The marginal Feature Importance scores are summarized in Table~\ref{table:beer-stats} and Figure~\ref{fig:beer-aroma-perturbation}. 
Compared to the Suff. IG and Suff. LIME methods, our \textbf{SIScollection} technique produces rationales that are much shorter and contain fewer irrelevant (i.e.\ not marginally important) features (Table~\ref{table:beer-stats}, Figures~\ref{fig:beer-aroma-perturbation} and~\ref{fig:beer-aroma-rationale-lengths}).
Note that by construction, the rationales of the  Suff. Perturb. method contain features with the greatest Feature Importance, since this precisely how the ranking in Suff.\ Perturb. is defined.  


\begin{table}
  \caption[Statistics for rationale length and feature importance in aroma prediction]{Statistics for rationale length and feature importance in aroma prediction. For rationale length, median and max indicate percentage of input text in the rationale. For marginal perturbed feature importance, we indicate the median importance of features in rationales and features from the other (non-rationale) text. $p$-values are computed using a Wilcoxon rank-sum test.}
  \label{table:beer-stats}
  \centering
  \begin{tabular}{lcccccc}
    \toprule
    \multirow{2}{*}{Method} & \multicolumn{3}{c}{Rationale Length (\% of text)} & \multicolumn{3}{c}{Marginal Perturbed Feature Importance}  \\
    	       & Med. & Max & $p$ (vs. SIS) & Med. (Rationale) & Med. (Other) & $p$ (vs. SIS)  \\
    \midrule
    SIS & \textbf{3.9\%} & \textbf{17.3\%} & -- & 0.0112 & 1.50e-05 & --  \\
	Suff. IG & 7.7\% & 89.7\% & 5e-26 & 0.0068 & 1.85e-05 & 3e-42  \\
	Suff. LIME & 7.2\% & 84.0\% & 4e-23 & 0.0075 & 1.87e-05 & 1e-35  \\
	Suff. Perturb. & 5.1\% & 18.3\% & 1e-06 & 0.0209 & 1.90e-05 & 1e-72  \\
    \bottomrule
  \end{tabular}
\end{table}


\subsection{Additional Results for Aroma aspect}
\label{sec:beerdets-sis-analysis}
\begin{figure}
    \centering
    \begin{minipage}[t]{0.48\textwidth}
        \centering
        \includegraphics[width=1.0\textwidth]{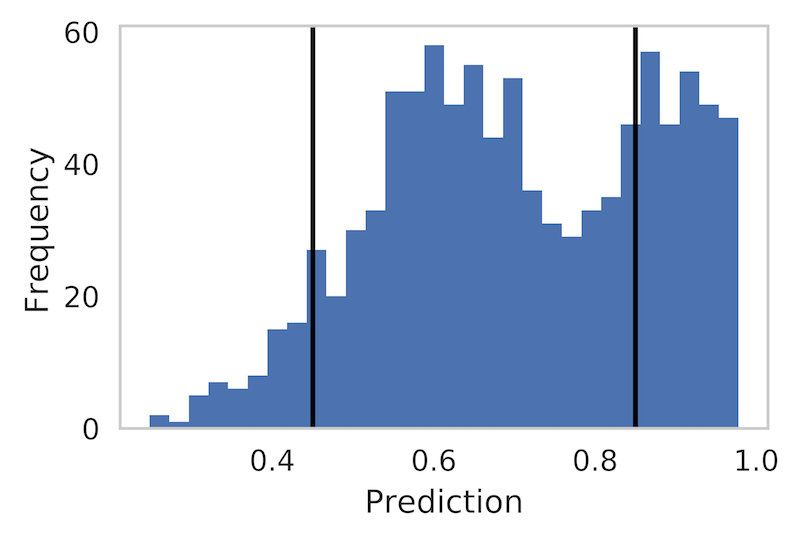}
        \caption[Prediction of aroma sentiment on the annotation set]{Predictive distribution on the annotation set (held-out) using the LSTM model for aroma. Vertical lines indicate decision thresholds ($\tau_{+} = 0.85$, $\tau_{-} = 0.45$) selected for \textbf{SIScollection}.}
        \label{fig:beer-aroma-predictive-dist}
    \end{minipage}\hfill
    \begin{minipage}[t]{0.48\textwidth}
        \centering
        \includegraphics[width=1.0\textwidth]{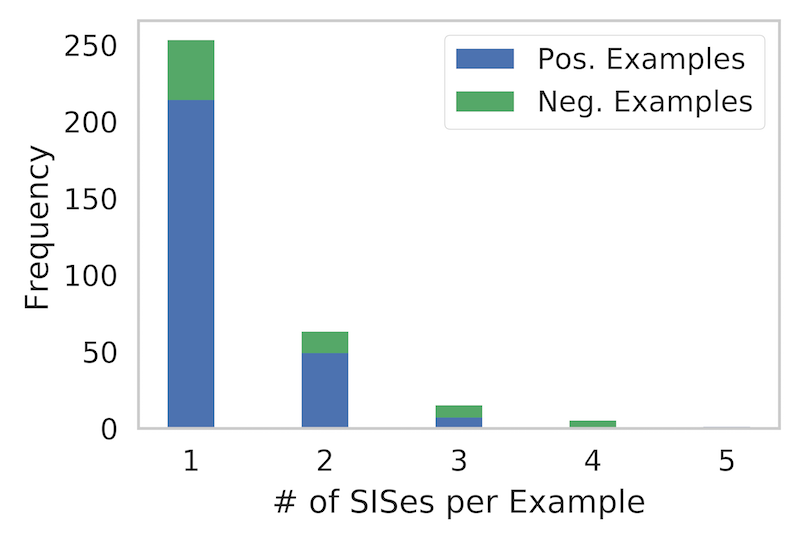}
        \caption[Number of SIS per for aroma beer reviews]{Number of sufficient input subsets for aroma identified by \textbf{SIScollection} per example.}
        \label{fig:beer-aroma-num-sis}
    \end{minipage}
\end{figure}

We apply our method to the set of reviews containing sentence-level annotations.
Note that these reviews (and the human annotations) were not seen during training.
We choose thresholds $\tau_{+} = 0.85$, $\tau_{-} = 0.45$ for strong positive and strong negative sentiment, respectively, and extract the complete set of sufficient input subsets using our method.  Note that in our formulation above, we apply our method to inputs $\mathbf{x}$ where $f(\mathbf{x}) \ge \tau$. For the sentiment analysis task, we analogously apply our method for both $f(\mathbf{x}) \ge \tau_{+}$ and $-f(\mathbf{x}) \ge -\tau_{-}$, where the model predicts either strong positive or strong negative sentiment, respectively.
These thresholds were set empirically such that they were sufficiently apart, based on the distribution of predictions (Figure~\ref{fig:beer-aroma-predictive-dist}).
For most reviews, \textbf{SIScollection} outputs just one or two SIS sets (Figure~\ref{fig:beer-aroma-num-sis}).

\begin{figure}
    \centering
    \includegraphics[width=0.5\textwidth]{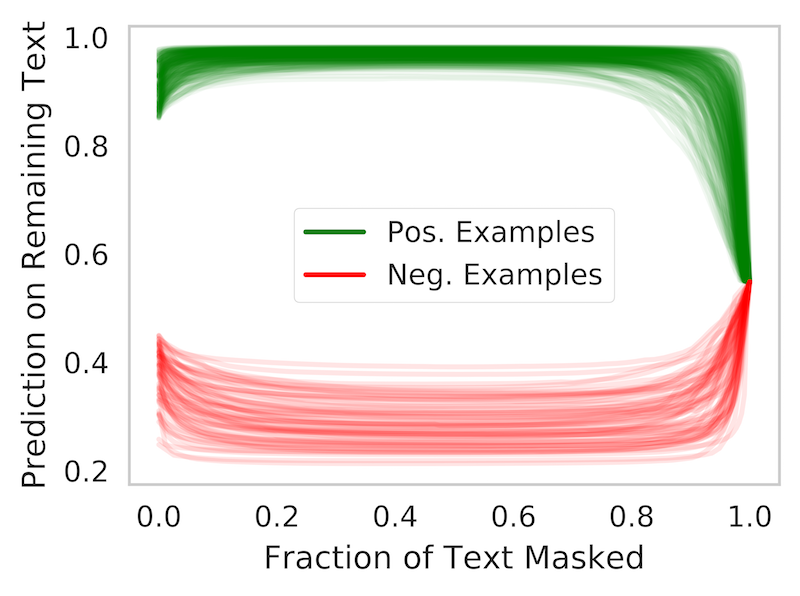}
    \caption[Prediction as function of remaining text for aroma prediction]{Prediction history on remaining (unmasked) text at each step of the \textbf{BackSelect} procedure, for examples where aroma sentiment is predicted.}
    \label{fig:beer-aroma-history-trends}
\end{figure}

We analyzed the predictor output following the elimination of each feature in the \textbf{BackSelect} procedure (Section~\ref{sec:methods}).
Figure~\ref{fig:beer-aroma-history-trends} shows the LSTM output on the remaining unmasked text $f(\mathbf{x}_{S \setminus \{i^*\} })$ at each iteration of \textbf{BackSelect}, for all examples.
This figure reveals that only a small number of features are needed by the model in order to make a strong prediction (most features can be removed without changing the prediction).
We see that as those final, critical features are removed, there is a rapid, monotonic decrease in output values.
Finally, we see that the first features to be removed by \textbf{BackSelect} are those which generally provide negative evidence against the decision.

\begin{table}
	\caption[SIS clusters for positive aroma prediction]{All clusters of sufficient input subsets extracted from reviews from the test set predicted to have positive aroma by the LSTM. Frequency indicates the number of occurrences of the SIS in the cluster.}
    \begingroup  
    \footnotesize
    \begin{tabularx}{\textwidth}{
        @{\hspace{-2.0em}}X
        @{\hspace{-1.5em}}X
        @{\hspace{-3.2em}}X
        @{\hspace{-2.0em}}X
        @{\hspace{-3.2em}}X
        @{\hspace{-2.0em}}X
        @{\hspace{-3.2em}}X
        @{\hspace{-2.0em}}X
        @{\hspace{-3.2em}}X
        @{\hspace{-2.5em}}
        }  
Cluster &                                            SIS \#1 &  Freq. &                                      SIS \#2 &  Freq. &                                 SIS \#3 &  Freq. &                                SIS \#4 &  Freq. \\
\midrule
$C_{1}$ &                          smell amazing wonderful &           2 &                        nice wonderful nose &           2 &                     wonderful amazing &           2 &                      amazing amazing &           2 \\[5pt]
$C_{2}$ &                       grapefruit mango pineapple &           2 &  pineapple grapefruit pineapple grapefruit &           1 &      hops grapefruit pineapple floyds &           1 &           mango pineapple incredible &           1 \\[5pt]
$C_{3}$ &          nice smell citrus nice grapefruit taste &           1 &             smell great complex ripe taste &           1 &  nice smell nice hop smell pine taste &           1 &     love nice nice smell bliss taste &           1 \\[5pt]
$C_{4}$ &                      fresh great fantastic taste &           1 &                 rich great fantastic hoped &           1 &          fantastic cherries fantastic &           1 &    everyone great snifters fantastic &           1 \\[5pt]
$C_{5}$ &                                   awesome bounds &           1 &                 awesome grapefruit awesome &           1 &              awesome awesome pleasing &           1 &                awesome nailed nailed &           1 \\[5pt]
$C_{6}$ &                              creme brulee brulee &           3 &                      creme brulee decadent &           1 &               incredible creme brulee &           1 &             creme brulee exceptional &           1 \\[5pt]
$C_{7}$ &  oak vanilla chocolate cinnamon vanilla oak love &           1 &          dose oak chocolate vanilla acidic &           1 &        vanilla figs oak thinner great &           1 &  chocolate aroma oak vanilla dessert &           1 \\
\end{tabularx}
    \endgroup
    \label{tab:beer-aroma-positives-clustering-supplement}
\end{table}

\begin{table}
    \caption[SIS clusters for negatives aroma prediction]{All clusters of sufficient input subsets extracted from reviews from the test set predicted to have negative aroma by the LSTM. Frequency indicates the number of occurrences of the SIS in the cluster. Dashes are used in clusters with under 4 unique SIS.}
    \begingroup  
    \footnotesize
    \begin{tabularx}{\textwidth}{
        @{\hspace{-2.0em}}X
        @{\hspace{-1.5em}}X
        @{\hspace{-3.2em}}X
        @{\hspace{-2.0em}}X
        @{\hspace{-3.2em}}X
        @{\hspace{-2.0em}}X
        @{\hspace{-3.2em}}X
        @{\hspace{-2.0em}}X
        @{\hspace{-3.2em}}X
        @{\hspace{-2.5em}}
        }  
Cluster &          SIS \#1 &  Freq. &          SIS \#2 & Freq. &           SIS \#3 & Freq. &              SIS \#4 & Freq. \\
\midrule
$C_{1}$ &          awful &          15 &  skunky skunky &          9 &        skunky t &          7 &       skunky taste &          6 \\[5pt]
$C_{2}$ &        garbage &           3 &  taste garbage &          1 &   garbage avoid &          1 &       garbage rice &          1 \\[5pt]
$C_{3}$ &          vomit &          16 &              - &          - &               - &          - &                  - &          - \\[5pt]
$C_{4}$ &   gross rotten &           1 &   rotten forte &          1 &  awkward rotten &          1 &  rotten offputting &          1 \\[5pt]
$C_{5}$ &  rancid horrid &           1 &       rancid t &          1 &          rancid &          1 &       rancid avoid &          1 \\[5pt]
$C_{6}$ &    rice t rice &           2 &      rice rice &          1 &  rice tasteless &          1 &     budweiser rice &          1 \\
\end{tabularx}

    \endgroup
    \label{tab:beer-aroma-negatives-clustering-supplement}
\end{table}



\subsection{Understanding Differences Between Sentiment Predictors}
\label{sec:beer-understanding-differences-dets}
We demonstrate how our SIS-clustering procedure can be used to understand differences in the types of concepts considered important by different neural network architectures.
In addition to the LSTM (see Section \ref{sec:beerdets-model-training}), we trained a convolutional neural network (CNN) on the same sentiment analysis task (on the aroma aspect).
The CNN architecture is as follows:
\begin{enumerate}
	\item \textbf{Input/Embeddings Layer}: Sequence with 500 timesteps, the word at each timestep is represented by a (learned) 100-dimensional embedding
    \item \textbf{Convolutional Layer 1}: Applies 128 filters of window size 3 over the sequence, with ReLU activation
    \item \textbf{Max Pooling Layer 1}: Max-over-time pooling, followed by flattening, to produce a $(128,)$ representation
    \item \textbf{Dense}: 1 neuron (sentiment output), sigmoid activation
\end{enumerate}
Note that a new set of embeddings was learned with the CNN.
As with the LSTM model, we use Adam \citepsi{adamsi} to minimize MSE on the training set.
For the aroma aspect, this CNN achieves 0.016 (0.850), 0.025 (0.748), 0.026 (0.741), 0.014 (0.662) MSE (and Pearson $\rho$) on the Train, Validation, Test, and Annotation sets, respectively.
We note that this performance is very similar to that from the LSTM (see Table~\ref{table:beerdets-dataset-stats}).

We apply our procedure to extract the SIS-collection from all applicable test examples using the CNN, as in Section~\ref{sec:sentanalysis}.
Figure~\ref{fig:beer-aroma-lstm-cnn-sis-preds} shows the predictions from one model (LSTM or CNN) when fed input examples that are SIS extracted with respect to the \textit{other} model (for reviews predicted to have positive sentiment toward the aroma aspect).
For example, in Figure~\ref{fig:beer-aroma-lstm-cnn-sis-preds}, ``CNN SIS Preds by LSTM'' refers to predictions made by the LSTM on the set of sufficient input subsets produced by applying our \textbf{SIScollection} procedure on all examples $\mathbf{x} \in \mathcal{X}_{\text{test}}$ for which $f_{\text{CNN}}(\mathbf{x}) \geq \tau_{+}$.\footnote{For experiments involving clustering and/or comparing different models, we use examples drawn from the Test fold (instead of Annotation fold, see Table~\ref{table:beerdets-dataset-stats}) to consider a larger number of examples.}
Since the word embeddings are model-specific, we embed each SIS using the embeddings of the model making the prediction (note that while the embeddings are different, the vocabulary is the same across the models).

In Table~\ref{tab:beer-aroma-model-differences-clustering}, we show five example clusters (and cluster composition) resulting from clustering the combined set of all sufficient input subsets extracted by the LSTM and CNN on reviews in the test set for which a model predicts positive sentiment toward the aroma aspect.
The complete clustering on reviews receiving positive sentiment predictions is shown in Table~\ref{tab:beer-aroma-full-joint-clustering-positives} and in Table~\ref{tab:beer-aroma-full-joint-clustering-negatives} for reviews receiving negative sentiment predictions.

\begin{table}[h!]
    \caption[Joint clustering of SIS from LSTM and CNN on beer reviews, positive aroma aspect]{Joint clustering of the SIS extracted from beer reviews predicted to have positive aroma by LSTM or CNN model. Frequency indicates the number of occurrences of the SIS in the cluster. Percentages quantify SIS per cluster from the LSTM. Dashes are used in clusters with under 4 unique SIS.}
    \begingroup  
    \footnotesize
    \begin{tabularx}{\textwidth}{
        X
        X
        @{\hspace{-3.2em}}X
        @{\hspace{-2.0em}}X
        @{\hspace{-3.2em}}X
        @{\hspace{-2.0em}}X
        @{\hspace{-3.2em}}X
        @{\hspace{-2.0em}}X
        @{\hspace{-3.2em}}X
        @{\hspace{-2.5em}}
        }  
Cluster &           SIS \#1 &  Freq. &                                  SIS \#2 & Freq. &                               SIS \#3 & Freq. &                              SIS \#4 & Freq. \\
\midrule
$C_{1}$ (LSTM: 20\%)  &  rich chocolate &          13 &                              very rich &          9 &                   chocolate complex &          5 &                        smells rich &          4 \\[5pt]
$C_{2}$ (LSTM: 21\%)  &           great &         248 &                                amazing &        119 &                           wonderful &        112 &                          fantastic &         75 \\[5pt]
$C_{3}$ (LSTM: 47\%)  &   best smelling &          23 &                        pineapple mango &          6 &                     mango pineapple &          6 &               pineapple grapefruit &          5 \\[5pt]
$C_{4}$ (LSTM: 5\%)   &       excellent &          42 &              excellent flemish flemish &          1 &      excellent excellent phenomenal &          1 &                                  - &          - \\[5pt]
$C_{5}$ (LSTM: 33\%)  &   oak chocolate &           2 &  chocolate raisins raisins oak bourbon &          1 &                       chocolate oak &          1 &                  raisins chocolate &          1 \\[5pt]
$C_{6}$ (LSTM: 5\%)   &        goodness &          19 &                      watering goodness &          1 &                                   - &          - &                                  - &          - \\[5pt]
$C_{7}$ (LSTM: 24\%)  &     pumpkin pie &          25 &         huge pumpkin aroma pumpkin pie &          1 &     aroma perfect pumpkin pie taste &          1 &  smell pumpkin nutmeg cinnamon pie &          1 \\[5pt]
$C_{8}$ (LSTM: 5\%)   &              jd &          13 &                             tremendous &          8 &                       tremendous jd &          1 &                                  - &          - \\[5pt]
$C_{9}$ (LSTM: 40\%)  &          brulee &          14 &                    creme brulee brulee &          3 &                         creme creme &          1 &               creme brulee amazing &          1 \\[5pt]
$C_{10}$ (LSTM: 0\%)  &           s wow &          20 &                                      - &          - &                                   - &          - &                                  - &          - \\[5pt]
$C_{11}$ (LSTM: 0\%)  &       delicious &          56 &                                      - &          - &                                   - &          - &                                  - &          - \\[5pt]
$C_{12}$ (LSTM: 0\%)  &       very nice &          23 &                                      - &          - &                                   - &          - &                                  - &          - \\[5pt]
$C_{13}$ (LSTM: 70\%) &   complex aroma &           5 &          aroma complex peaches complex &          1 &  aroma complex interesting cherries &          1 &                      aroma complex &          1 \\
\end{tabularx}

    \endgroup
    \label{tab:beer-aroma-full-joint-clustering-positives}
\end{table}

\begin{table}[h!]
    \caption[Joint clustering of SIS from LSTM and CNN on beer reviews, negative aroma aspect]{Joint clustering of the SIS extracted from beer reviews predicted to have negative aroma by LSTM or CNN model. Frequency indicates the number of occurrences of the SIS in the cluster. Percentages quantify SIS per cluster from the LSTM. Dashes are used in clusters with under 4 unique SIS.}
    \begingroup  
    \footnotesize
    \begin{tabularx}{\textwidth}{
        X
        X
        @{\hspace{-3.2em}}X
        @{\hspace{-2.0em}}X
        @{\hspace{-3.2em}}X
        @{\hspace{-2.0em}}X
        @{\hspace{-3.2em}}X
        @{\hspace{-2.0em}}X
        @{\hspace{-3.2em}}X
        @{\hspace{-2.5em}}
        }  
Cluster &           SIS \#1 &  Freq. &             SIS \#2 & Freq. &            SIS \#3 & Freq. &                          SIS \#4 & Freq. \\
\midrule
$C_{1}$ (LSTM: 29\%)  &             not &         247 &                no &        105 &              bad &        104 &                          macro &         94 \\[5pt]
$C_{2}$ (LSTM: 100\%) &    gross rotten &           1 &                 - &          - &                - &          - &                              - &          - \\[5pt]
$C_{3}$ (LSTM: 100\%) &  rotten garbage &           1 &                 - &          - &                - &          - &                              - &          - \\[5pt]
$C_{4}$ (LSTM: 62\%)  &           vomit &          26 &                 - &          - &                - &          - &                              - &          - \\[5pt]
$C_{5}$ (LSTM: 21\%)  &       budweiser &          22 &  sewage budweiser &          1 &  metal budweiser &          1 &  budweiser budweiser budweiser &          1 \\[5pt]
$C_{6}$ (LSTM: 100\%) &    garbage rice &           1 &                 - &          - &                - &          - &                              - &          - \\[5pt]
$C_{7}$ (LSTM: 3\%)   &             n't &          19 &          adjuncts &         14 &     n't adjuncts &          1 &                              - &          - \\[5pt]
$C_{8}$ (LSTM: 0\%)   &           faint &          82 &                 - &          - &                - &          - &                              - &          - \\[5pt]
$C_{9}$ (LSTM: 0\%)   &         adjunct &          42 &                 - &          - &                - &          - &                              - &          - \\
\end{tabularx}

    \endgroup
    \label{tab:beer-aroma-full-joint-clustering-negatives}
\end{table}

\clearpage
\subsection{Results for Appearance and Palate aspects}
\label{sec:beer-other-aspects}
For posterity, we include results here from repeating the analysis in our paper for the two other non-aroma aspects measured in the beer reviews data: appearance and palate.

\begin{figure}[h]
    \centering
    \includegraphics[width=0.65\textwidth]{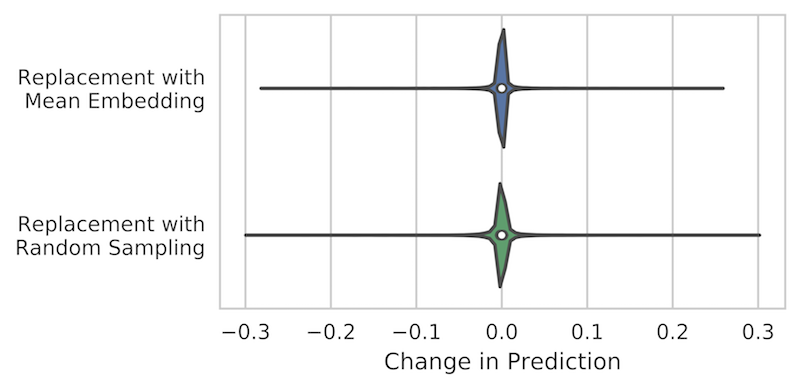}
    \caption[Mean vs. hot-deck imputation in appearance prediction]{Change in appearance prediction ($f(\mathbf{x} \setminus \{i\}) - f(\mathbf{x})$) after masking a randomly chosen word with mean imputation or hot-deck imputation. 10,000 replacements were sampled from the appearance beer reviews training set.}
    \label{fig:beer-appearance-mean-embedding}
\end{figure}

\begin{figure}[h]
    \centering
    \begin{minipage}[t]{0.48\textwidth}
        \centering
        \includegraphics[width=1.0\textwidth]{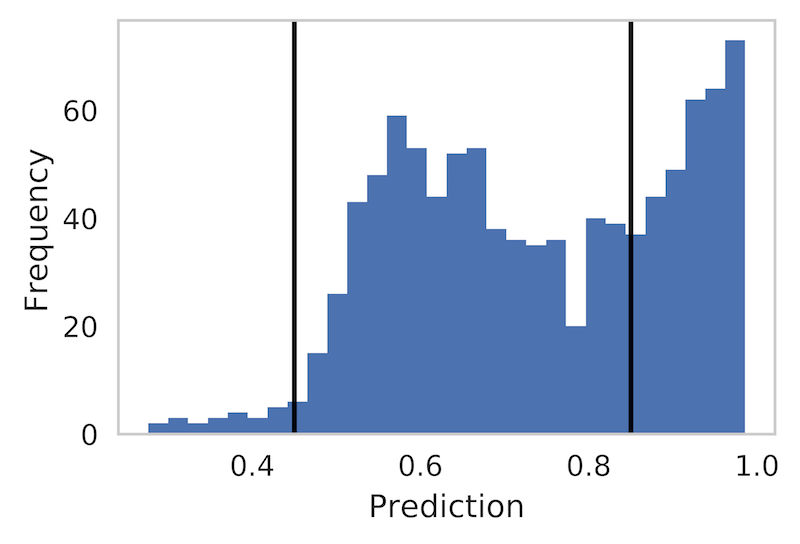}
        \caption[Prediction of appearance sentiment on the annotation set]{Predictive distribution on the annotation set (held-out) using the LSTM model for appearance. Vertical lines indicate decision thresholds ($\tau_{+} = 0.85$, $\tau_{-} = 0.45$) selected for \textbf{SIScollection}.}
        \label{fig:beer-appearance-predictive-dist}
    \end{minipage}\hfill
    \begin{minipage}[t]{0.48\textwidth}
        \centering
        \includegraphics[width=1.0\textwidth]{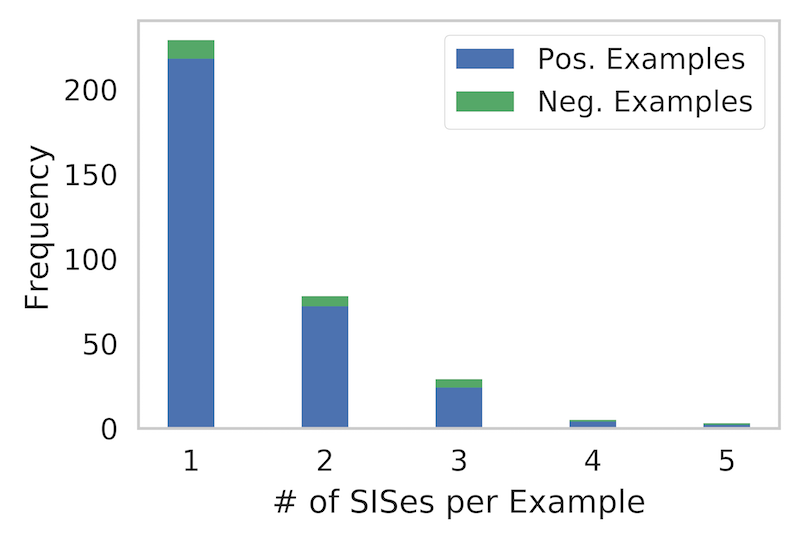}
        \caption[Number of SIS per for appearance beer reviews]{Number of sufficient input subsets for appearance identified by \textbf{SIScollection} per example.}
        \label{fig:beer-appearance-num-sis}
    \end{minipage}
\end{figure}

\begin{figure}[h]
    \centering
    \begin{minipage}{0.48\textwidth}
        \centering
        \includegraphics[width=1.0\textwidth]{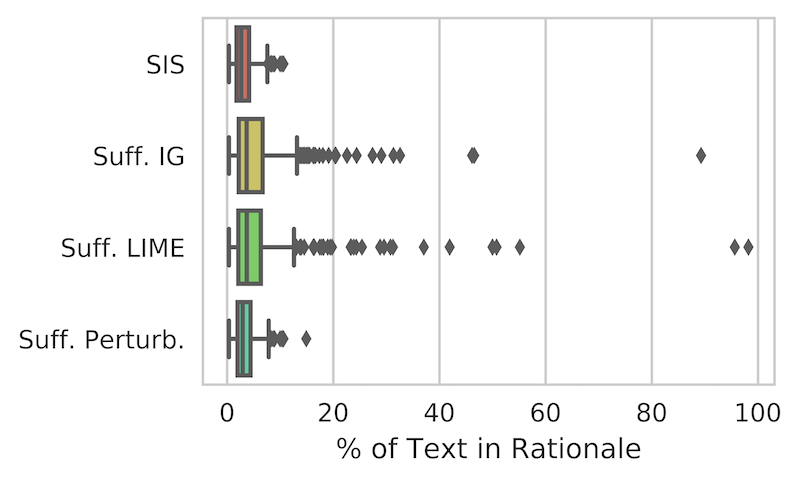}
        \caption[Length of rationales for appearance prediction]{Length of rationales for appearance prediction}
        \label{fig:beer-appearance-rationale-lengths}
    \end{minipage}\hfill
    \begin{minipage}{0.48\textwidth}
        \centering
        \includegraphics[width=1.0\textwidth]{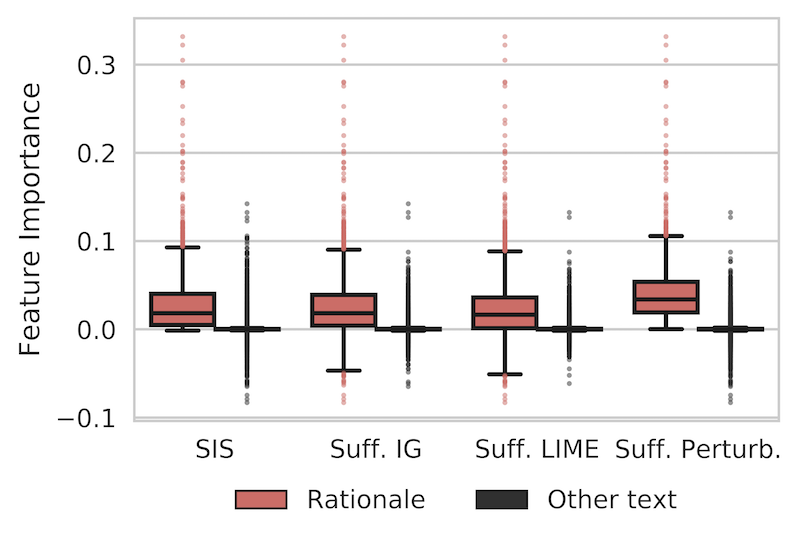}
        \caption[Feature importance for appearance prediction]{Importance of individual features for appearance prediction in beer review}
        \label{fig:beer-appearance-perturbation}
    \end{minipage}
\end{figure}

\begin{table}[h]
  \caption[Statistics for rationale length and feature importance in appearance prediction]{Statistics for rationale length and feature importance in appearance prediction. For rationale length, median and max indicate percentage of input text in the rationale. For marginal perturbed feature importance, we indicate the median importance of features in rationales and features from the other (non-rationale) text. $p$-values are computed using a Wilcoxon rank-sum test.}
  \label{table:beer-stats-appearance}
  \centering
  \begin{tabular}{lcccccc}
    \toprule
    \multirow{2}{*}{Method} & \multicolumn{3}{c}{Rationale Length (\% of text)} & \multicolumn{3}{c}{Marginal Perturbed Feature Importance}  \\
    	       & Med. & Max & $p$ (vs. SIS) & Med. (Rationale) & Med. (Other) & $p$ (vs. SIS)  \\
    \midrule
    SIS & \textbf{2.6\%} & \textbf{10.6\%} & -- & 0.0183 & 1.72e-05 & --  \\
    Suff. IG & 3.7\% & 89.3\% & 2e-09 & 0.0184 & 2.41e-05 & 1e-02  \\
    Suff. LIME & 3.7\% & 98.2\% & 8e-09 & 0.0167 & 2.38e-05 & 6e-09  \\
    Suff. Perturb. & 3.0\% & 14.9\% & 9e-03 & 0.0339 & 2.51e-05 & 5e-44  \\
    \bottomrule
  \end{tabular}
\end{table}

\begin{figure}[h]
    \centering
    \begin{minipage}[t]{0.48\textwidth}
        \centering
        \includegraphics[width=0.97\linewidth]{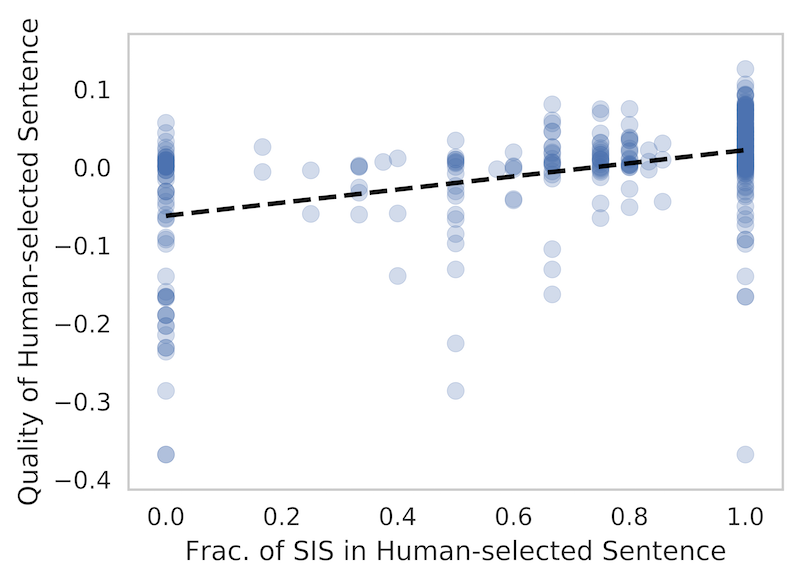}
        \caption[QHS vs. fraction of SIS in human rationale for appearance prediction]{QHS vs. fraction of SIS in human rationale for appearance prediction}
        \label{fig:beer-appearance-quality-annotations}
    \end{minipage}\hfill
    \begin{minipage}[t]{0.48\textwidth}
        \centering
        \includegraphics[width=1.0\textwidth]{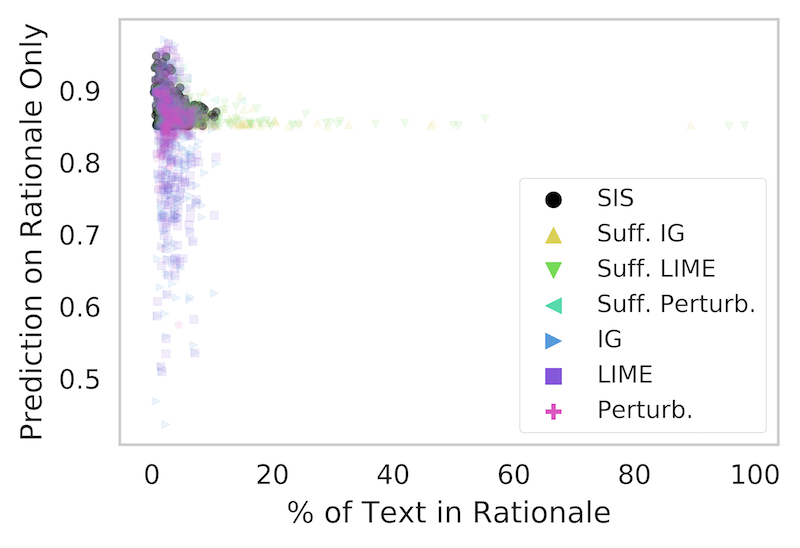}
        \caption[Prediction on only rationale vs. rationale length]{Prediction on rationales only vs. rationale length for various methods in positive sentiment examples for appearance. The threshold for sufficiency was $\tau_{+} = 0.85$.}
        \label{fig:beer-appearance-prediction-vs-length}
    \end{minipage}
\end{figure}

\begin{figure}[h]
    \centering
    \includegraphics[width=0.5\textwidth]{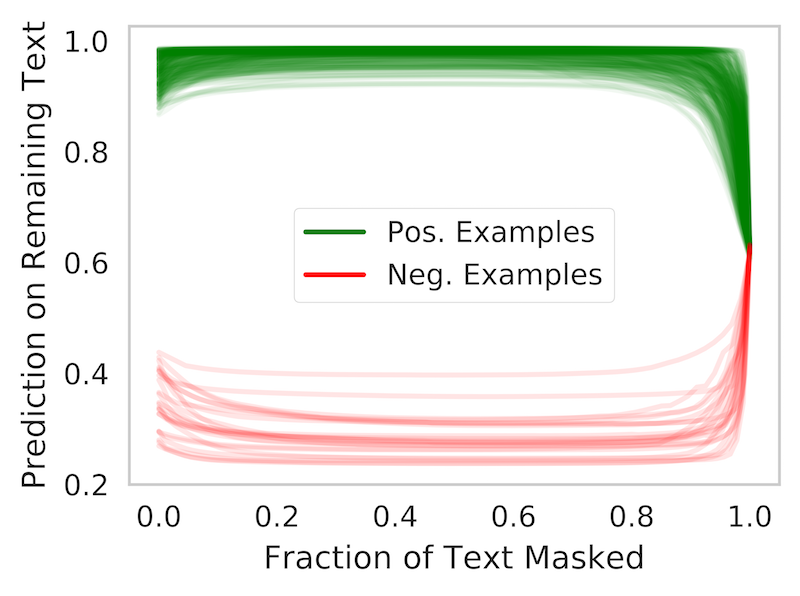}
    \caption[Prediction as function of remaining text for appearance aspect]{Prediction history on remaining (unmasked) text at each step of the \textbf{BackSelect} procedure, for examples where appearance sentiment is predicted.}
    \label{fig:beer-appearance-history-trends}
\end{figure}

\begin{table}[h]
    \caption[SIS clusters for positive appearance prediction]{All clusters of sufficient input subsets extracted from reviews from the test set predicted to have positive appearance by the LSTM. Frequency indicates the number of occurrences of the SIS in the cluster. Dashes are used in clusters with under 4 unique SIS.}
    \begingroup  
    \footnotesize
    \begin{tabularx}{\textwidth}{
        @{\hspace{-2.0em}}X
        @{\hspace{-1.5em}}X
        @{\hspace{-3.2em}}X
        @{\hspace{-2.0em}}X
        @{\hspace{-3.2em}}X
        @{\hspace{-2.0em}}X
        @{\hspace{-3.2em}}X
        @{\hspace{-2.0em}}X
        @{\hspace{-3.2em}}X
        @{\hspace{-2.5em}}
        }  
Cluster &           SIS \#1 &  Freq. &                   SIS \#2 & Freq. &                  SIS \#3 & Freq. &                 SIS \#4 & Freq. \\
\midrule
$C_{1}$ &       beautiful &         376 &                   nitro &         51 &            looks great &         38 &         great looking &         32 \\[5pt]
$C_{2}$ &        gorgeous &          83 &                       - &          - &                      - &          - &                     - &          - \\[5pt]
$C_{3}$ &     beautifully &           7 &  absolutely beautifully &          2 &    beautifully pillowy &          1 &     beautifully bands &          1 \\[5pt]
$C_{4}$ &       brilliant &           5 &        brilliant slowly &          1 &  wonderfully brilliant &          1 &  appearance brilliant &          1 \\[5pt]
$C_{5}$ &  lovely looking &           3 &            black lovely &          3 &      impressive lovely &          1 &        lovely crystal &          1 \\
\end{tabularx}

    \endgroup
    \label{tab:beer-appearance-positives-clustering-supplement}
\end{table}

\begin{table}[h]
    \caption[SIS clusters for negative appearance prediction]{All clusters of sufficient input subsets extracted from reviews from the test set predicted to have negative appearance by the LSTM. Frequency indicates the number of occurrences of the SIS in the cluster. Dashes are used in clusters with under 4 unique SIS.}
    \begingroup  
    \footnotesize
    \begin{tabularx}{\textwidth}{
        @{\hspace{-2.0em}}X
        @{\hspace{-1.5em}}X
        @{\hspace{-3.2em}}X
        @{\hspace{-2.0em}}X
        @{\hspace{-3.2em}}X
        @{\hspace{-2.0em}}X
        @{\hspace{-3.2em}}X
        @{\hspace{-2.0em}}X
        @{\hspace{-3.2em}}X
        @{\hspace{-2.5em}}
        }  
Cluster &        SIS \#1 &  Freq. &     SIS \#2 & Freq. &                 SIS \#3 & Freq. &  SIS \#4 & Freq. \\
\midrule
$C_{1}$ &         piss &          46 &      zero &         38 &           water water &         37 &  water &         27 \\[5pt]
$C_{2}$ &  unappealing &          12 &  floaties &         12 &  floaties unappealing &          1 &      - &          - \\[5pt]
$C_{3}$ &         ugly &          12 &         - &          - &                     - &          - &      - &          - \\
\end{tabularx}

    \endgroup
    \label{tab:beer-appearance-negatives-clustering-supplement}
\end{table}


\begin{figure}[h]
    \centering
    \includegraphics[width=0.65\textwidth]{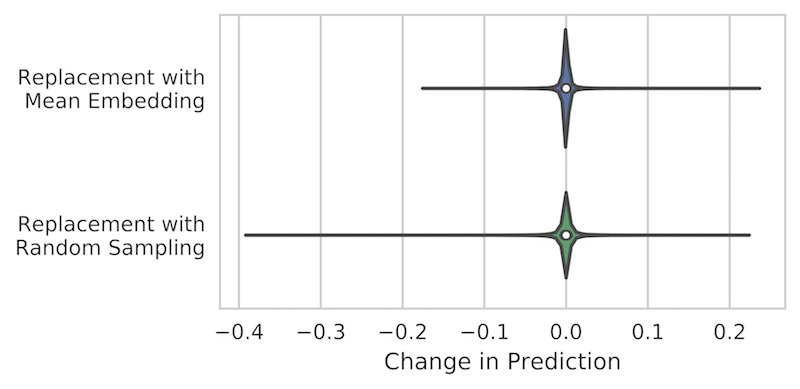}
    \caption[Mean vs. hot-deck imputation for palate prediction]{Change in palate prediction ($f(\mathbf{x} \setminus \{i\}) - f(\mathbf{x})$) after masking a randomly chosen word with mean imputation or hot-deck imputation. 10,000 replacements were sampled from the palate beer reviews training set.}
    \label{fig:beer-palate-mean-embedding}
\end{figure}

\begin{figure}[h]
    \centering
    \begin{minipage}[t]{0.48\textwidth}
        \centering
        \includegraphics[width=1.0\textwidth]{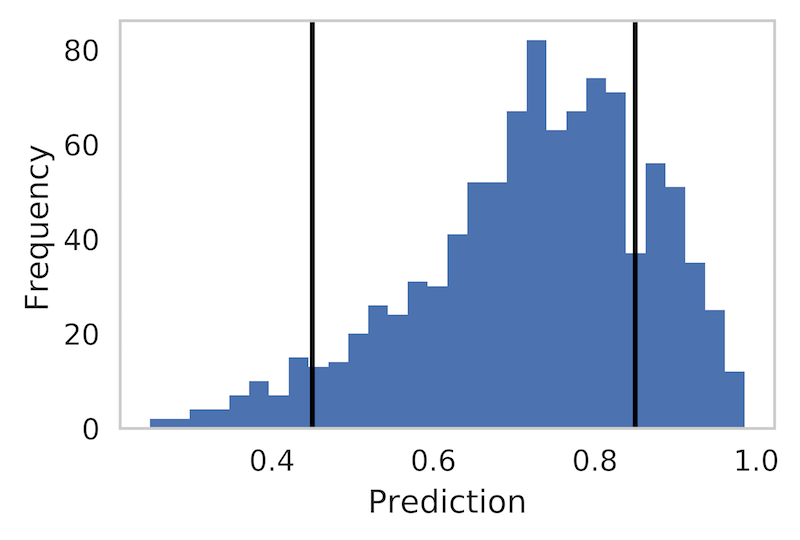}
        \caption[Prediction of palate sentiment on the annotation set]{Predictive distribution on the annotation set (held-out) using the LSTM model for palate. Vertical lines indicate decision thresholds ($\tau_{+} = 0.85$, $\tau_{-} = 0.45$) selected for \textbf{SIScollection}.}
        \label{fig:beer-palate-predictive-dist}
    \end{minipage}\hfill
    \begin{minipage}[t]{0.48\textwidth}
        \centering
        \includegraphics[width=1.0\textwidth]{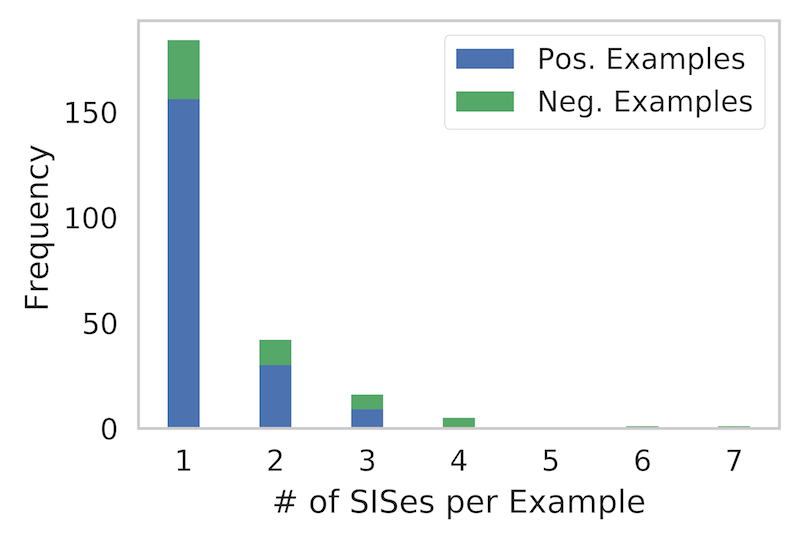}
        \caption[Number of SIS per for palate beer reviews]{Number of sufficient input subsets for palate identified by \textbf{SIScollection} per example.}
        \label{fig:beer-palate-num-sis}
    \end{minipage}
\end{figure}

\begin{figure}[h]
    \centering
    \begin{minipage}{0.48\textwidth}
        \centering
        \includegraphics[width=1.0\textwidth]{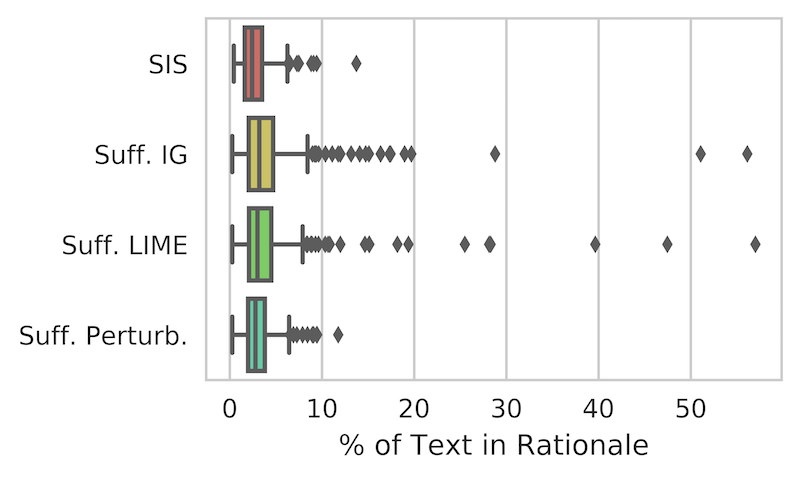}
        \caption[Length of rationales for palate prediction]{Length of rationales for palate prediction}
        \label{fig:beer-palate-rationale-lengths}
    \end{minipage}\hfill
    \begin{minipage}{0.48\textwidth}
        \centering
        \includegraphics[width=1.0\textwidth]{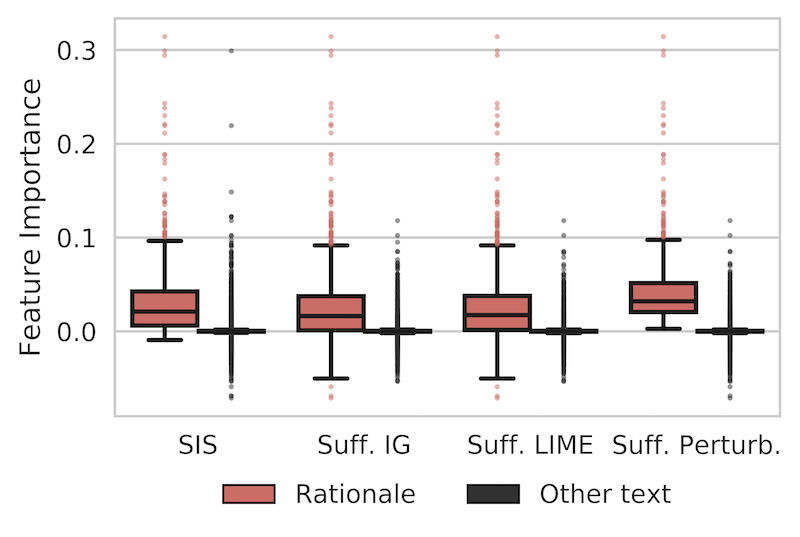}
        \caption[Feature importance for palate prediction]{Importance of individual features in beer review palate rationales}
        \label{fig:beer-palate-perturbation}
    \end{minipage}
\end{figure}

\begin{table}[h]
  \caption[Statistics for rationale length and feature importance in palate prediction]{Statistics for rationale length and feature importance in palate prediction. For rationale length, median and max indicate percentage of input text in the rationale. For marginal perturbed feature importance, we indicate the median importance of features in rationales and features from the other (non-rationale) text. $p$-values are computed using a Wilcoxon rank-sum test.}
  \label{table:beer-stats-palate}
  \centering
  \begin{tabular}{lcccccc}
    \toprule
    \multirow{2}{*}{Method} & \multicolumn{3}{c}{Rationale Length (\% of text)} & \multicolumn{3}{c}{Marginal Perturbed Feature Importance}  \\
    	       & Med. & Max & $p$ (vs. SIS) & Med. (Rationale) & Med. (Other) & $p$ (vs. SIS)  \\
    \midrule
    SIS & \textbf{2.4\%} & 13.7\% & -- & 0.0210 & -8.94e-07 & --  \\
    Suff. IG & 3.2\% & 56.1\% & 2e-06 & 0.0163 & -9.54e-07 & 6e-10  \\
    Suff. LIME & 3.0\% & 57.0\% & 7e-06 & 0.0173 & -1.19e-06 & 2e-07  \\
    Suff. Perturb. & 2.8\% & \textbf{11.8\%} & 3e-03 & 0.0319 & -1.25e-06 & 5e-26  \\
    \bottomrule
  \end{tabular}
\end{table}

\begin{figure}[h]
    \centering
    \begin{minipage}[t]{0.48\textwidth}
        \centering
        \includegraphics[width=0.97\linewidth]{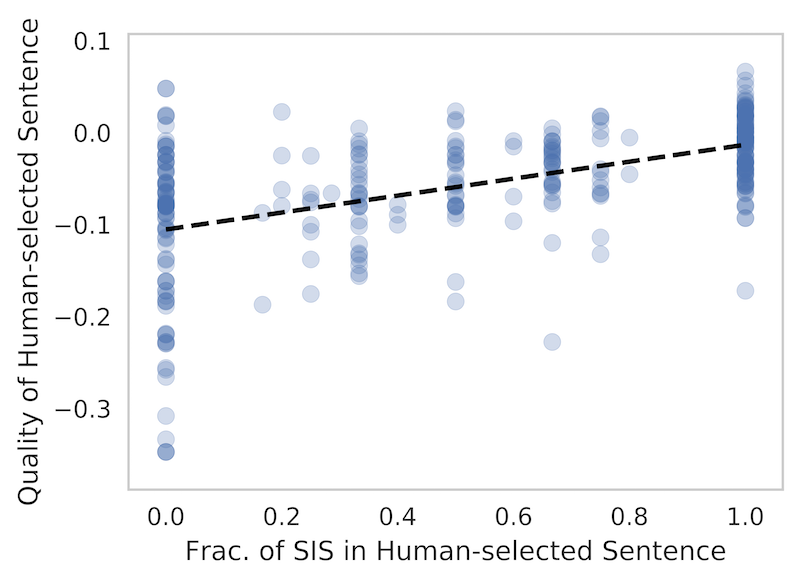}
        \caption[QHS vs. fraction of SIS in human rationale for palate prediction]{QHS vs. fraction of SIS in human rationale for palate prediction}
        \label{fig:beer-palate-quality-annotations}
    \end{minipage}\hfill
    \begin{minipage}[t]{0.48\textwidth}
        \centering
        \includegraphics[width=1.0\textwidth]{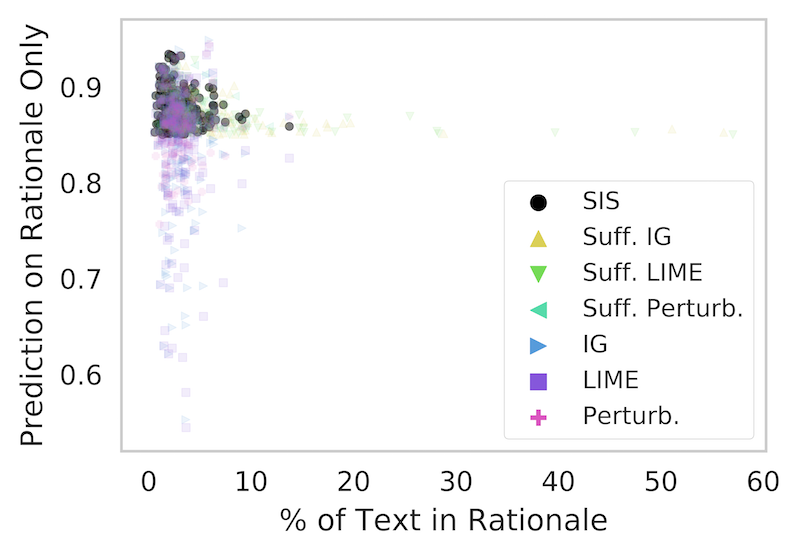}
        \caption[Prediction on only rationale vs. rationale length for palate prediction]{Prediction on rationales only vs. rationale length for various methods in positive sentiment examples for palate. The threshold for sufficiency was $\tau_{+} = 0.85$.}
        \label{fig:beer-palate-prediction-vs-length}
    \end{minipage}
\end{figure}

\begin{figure}[h]
    \centering
    \includegraphics[width=0.51\textwidth]{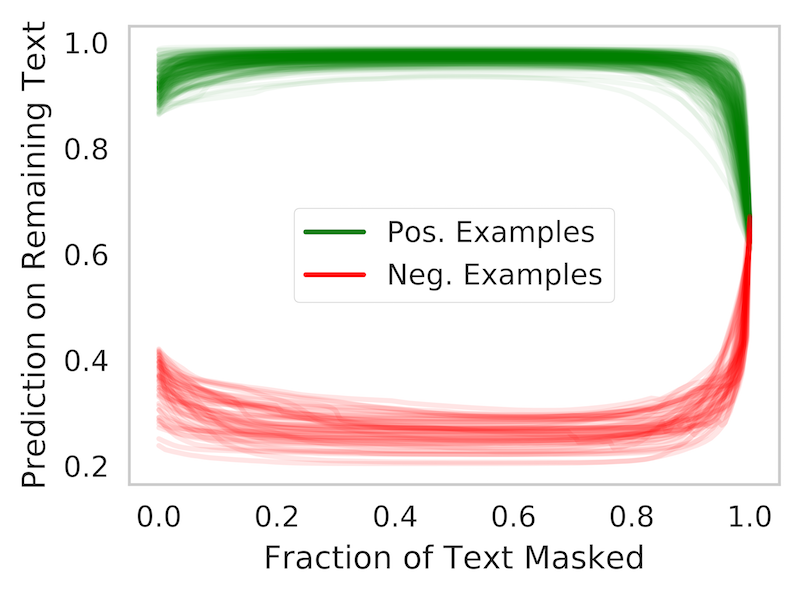}
    \caption[Prediction as function of remaining text for palate aspect]{Prediction history on remaining (unmasked) text at each step of the \textbf{BackSelect} procedure, for examples where palate sentiment is predicted.}
    \label{fig:beer-palate-history-trends}
\end{figure}

\begin{table}[h]
    \caption[SIS clusters for positive palate prediction]{All clusters of sufficient input subsets extracted from reviews from the test set predicted to have positive palate by the LSTM. Frequency indicates the number of occurrences of the SIS in the cluster. Dashes are used in clusters with under 4 unique SIS.}
    \begingroup  
    \footnotesize
    \begin{tabularx}{\textwidth}{
        @{\hspace{-2.0em}}X
        @{\hspace{-1.5em}}X
        @{\hspace{-3.2em}}X
        @{\hspace{-2.0em}}X
        @{\hspace{-3.2em}}X
        @{\hspace{-2.0em}}X
        @{\hspace{-3.2em}}X
        @{\hspace{-2.0em}}X
        @{\hspace{-3.2em}}X
        @{\hspace{-2.5em}}
        }  
Cluster &                  SIS \#1 &  Freq. &                  SIS \#2 & Freq. &              SIS \#3 & Freq. &                SIS \#4 & Freq. \\
\midrule
$C_{1}$ &          smooth creamy &          27 &           silky smooth &         20 &  mouthfeel perfect &         16 &       creamy perfect &         12 \\[5pt]
$C_{2}$ &  mouthfeel exceptional &           6 &  exceptional mouthfeel &          4 &                  - &          - &                    - &          - \\[5pt]
$C_{3}$ &                perfect &          50 &        perfect perfect &          6 &                  - &          - &                    - &          - \\[5pt]
$C_{4}$ &         smooth velvety &           6 &         velvety smooth &          6 &                  - &          - &                    - &          - \\[5pt]
$C_{5}$ &                   silk &          11 &                      - &          - &                  - &          - &                    - &          - \\[5pt]
$C_{6}$ &         smooth perfect &           8 &   mouth smooth perfect &          1 &     perfect smooth &          1 &                    - &          - \\[5pt]
$C_{7}$ &          perfect great &           5 &          great perfect &          2 &      feels perfect &          2 &  perfect feels great &          1 \\
\end{tabularx}

    \endgroup
    \label{tab:beer-palate-positives-clustering-supplement}
\end{table}

\begin{table}
    \caption[SIS clusters for negative palate prediction]{All clusters of sufficient input subsets extracted from reviews from the test set predicted to have negative palate by the LSTM. Frequency indicates the number of occurrences of the SIS in the cluster.}
    \begingroup  
    \footnotesize
    \begin{tabularx}{\textwidth}{
        @{\hspace{-2.0em}}X
        @{\hspace{-1.5em}}X
        @{\hspace{-3.2em}}X
        @{\hspace{-2.0em}}X
        @{\hspace{-3.2em}}X
        @{\hspace{-2.0em}}X
        @{\hspace{-3.2em}}X
        @{\hspace{-2.0em}}X
        @{\hspace{-3.2em}}X
        @{\hspace{-2.5em}}
        }  
Cluster &                  SIS \#1 &  Freq. &                     SIS \#2 &  Freq. &                  SIS \#3 &  Freq. &                     SIS \#4 &  Freq. \\
\midrule
$C_{1}$ &         overcarbonated &          12 &  mouthfeel overcarbonated &           3 &     way overcarbonated &           1 &  overcarbonated mouthfeel &           1 \\[5pt]
$C_{2}$ &                 watery &         302 &                      thin &         238 &                   flat &         118 &            mouthfeel thin &          33 \\[5pt]
$C_{3}$ &  too carbonation masks &           1 &         too carbonation d &           1 &  mouthfeel odd too too &           1 &     too carbonated admire &           1 \\[5pt]
$C_{4}$ &       lack carbonation &           4 &          carbonation lack &           4 &      carbonation hurts &           2 &          issue lack hurts &           1 \\
\end{tabularx}

    \endgroup
    \label{tab:beer-palate-negatives-clustering-supplement}
\end{table}


\clearpage
\bibliographystylesi{apa}
{ 
\bibliographysi{interpretability}
}

\end{document}